\documentclass{article} 
\usepackage{amssymb}
\usepackage{subcaption}
\usepackage{amsmath,amsfonts,amsthm}
\usepackage{array}
\usepackage{multirow}
\usepackage[toc,page,header]{appendix}
\usepackage[linesnumbered,ruled]{algorithm2e}
\usepackage{algpseudocode}
\usepackage{mathtools}
\usepackage{amssymb,mathtools}
\usepackage{cuted}
\usepackage{afterpage}
\usepackage{bbm}
\usepackage{color}
\usepackage{algorithm2e}
\usepackage{mathtools}

\usepackage{iclr2019_conference,times}

\usepackage{amsmath,amsfonts,bm}

\def\eqref#1{equation~\ref{#1}}

\def\1{\bm{1}}

\DeclareMathAlphabet{\mathsfit}{\encodingdefault}{\sfdefault}{m}{sl}
\SetMathAlphabet{\mathsfit}{bold}{\encodingdefault}{\sfdefault}{bx}{n}

\usepackage{hyperref}
\usepackage{url}

\title{Data Poisoning attack against Unsupervised Node Embedding Methods}

\author{
Mingjie Sun$^{1}{}$\thanks{Work done while visiting UC Berkeley. } , Jian Tang$^2$, Huichen Li$^3$, Bo Li$^3$, Chaowei Xiao$^4$, Yao Chen$^5$, Dawn Song$^6$ \vspace{0.25ex} \\
$^1$Tsinghua University~~$^2$Mila \& HEC Montreal, Canada~~$^3$UIUC~~$^4$University of Michigan\\$^5$Tecent~~$^6$University of California, Berkeley \\
}

\iclrfinalcopy 
\begin{document}
\maketitle
\begin{abstract}

Unsupervised node embedding methods (e.g., DeepWalk, LINE, and node2vec) have attracted growing interests given their simplicity and effectiveness. However, although these methods have been proved effective in a variety of applications, none of the existing work has analyzed the robustness of them. This could be very risky if these methods are attacked by an adversarial party. In this paper, we take the task of link prediction as an example, which is one of the most fundamental problems for graph analysis, and introduce a data poisoning
attack to node embedding methods. We give a complete characterization of attacker's utilities and present efficient solutions to adversarial attacks for two popular node embedding methods: DeepWalk and LINE. We evaluate our proposed attack model on multiple real-world graphs. Experimental results show that our proposed model can significantly affect the results of link prediction by slightly changing the graph structures (e.g., adding or removing a few edges). We also show that our proposed model is very general and can be transferable across different embedding methods. 
Finally, we conduct a case study on a coauthor network to better understand our attack method.

\end{abstract}

\section{Introduction}

Node representations, which represent each node with a low-dimensional vector, have been proved effective in a variety of applications such as node classification~\citep{perozzi2014deepwalk}, link prediction~\citep{grover2015node2vec}, and visualization~\citep{tang2016visualizing}. Some popular node embedding methods include DeepWalk~\citep{perozzi2014deepwalk}, LINE~\citep{tang2015LINE}, and node2vec~\citep{grover2015node2vec}. These methods learn the node embeddings by preserving graph structures, which do not depend on specific tasks. As a result, the learned node embeddings are very general and can be potentially useful to multiple downstream tasks. 

However, although these methods are very effective and have been used for a variety of tasks, none of the existing work has studied the robustness of these methods. As a result, these methods are susceptible to a risk of being maliciously attacked. Take the task of link prediction in a social network (e.g., Twitter) as an example, which is one of the most important applications of node embedding methods.  A malicious party may create malicious users in a social network and attack the graph structures (e.g., adding and removing edges) so that the effectiveness of node embedding methods is maximally degraded. For example, the attacker may slightly change the graph structures (e.g., following more users) so that the probability of a specific user to be recommended/linked can be significantly increased or decreased. Such a kind of attack is known as \emph{data poisoning}. In this paper we are interested in the robustness of the node embedding methods w.r.t. data poisoning, and their vulnerability to the adversarial attacks in the worst case.  

We are inspired by existing literature on adversarial attack, which has been extensively studied for different machine learning systems~\citep{szegedy2013intriguing,goodfellow2014explaining,moosavi2016deepfool,carlini2017towards,xiao2018spatially,xiao2018generating,xiao2018characterizing,xie2017adversarial,cisse2017houdini,yang2018realistic}. Specifically, it has been shown that deep neural networks are very sensitive to adversarial attacks, which can significantly change the prediction results by slightly perturbing the input data. However, most of existing work on adversarial attack focus on image~\citep{szegedy2013intriguing,goodfellow2014explaining,moosavi2016deepfool,carlini2017towards,xiao2018spatially} and text data~\citep{cheng2018seq2sick,jia2017adversarial}, which are independently distributed while this work focuses on graph data. There are some very recent work which studied adversarial attack for graph data~\citep{ICML2018graph, KDD2018graph}. However, these work mainly studied graph neural networks, which are supervised methods, and the gradients for changing the output label can be leveraged. Therefore, in this paper we are looking for an approach that is able to attack the unsupervised node embedding methods for graphs.  

In this paper, we introduce a systematic approach to adversarial attacks against unsupervised node embedding methods. We assume that the attacker can poison the graph structures by either removing or adding edges. Two types of adversarial goals are studied including \emph{integrity attack}, which aims to attack the probabilities of specific links, and \emph{availability attack}, which aims to increase overall prediction errors. We propose a unified optimization framework based on projected gradient descent to optimally attack both goals. In addition, we conduct a case study on a coauthor network to better understand our attack method. To summarize, we make the following contributions:
\begin{itemize}
    \item We formulate the problem of attacking unsupervised node embeddings for the task of link prediction and introduce a complete characterization of attacker utilities.
    \item We propose an efficient algorithm based on projected gradient descent to attack unsupervised node embedding algorithms, specifically DeepWalk and LINE, based on the first order Karush Kuhn Tucker (KKT) conditions.
    \item We conduct extensive experiments on real-world graphs to show the efficacy of our proposed attack model on the task of link prediction. 
    Moreover, results show that our proposed attack model is transferable across different node embedding methods.
    \item Finally, we conduct a case study on a coauthor network and give an intuitive understanding of our attack method.
\end{itemize}

\section{Related Work}
Adversarial attack against image classification has been extensively studied in recent years~\citep{szegedy2013intriguing,goodfellow2014explaining,moosavi2016deepfool,carlini2017towards,xiao2018spatially,xiao2018generating}. However, adversarial attacks against graph have rarely been investigated before. Existing work~\citep{ICML2018graph,KDD2018graph} on adversarial attacks on graph are limited to graph neural networks~\citep{thomas2017gcn}, a supervised learning method. Our work, instead, shows the vulnerabilities of unsupervised methods on graph. Here we briefly summarize previous work on graph embedding methods and then we will give an overview of adversarial attacks on graph data. Last, we will show the related work on the connection to matrix  factorization.
\vspace{-2ex}
\paragraph{Unsupervised Learning on Graph} Previous work on knowledge mining in graph has mainly focused on embedding methods, where the goal is to learn a latent embedding for each node in the graph. DeepWalk~\citep{perozzi2014deepwalk}, LINE~\citep{tang2015LINE} and Node2vec~\citep{grover2015node2vec} are the three most representative unsupervised methods on graph. 
\vspace{-2ex}
\paragraph{Adversarial Attack on Graph} There are a few work on adversarial attack on graph before. Test time attack on graph convolutional network has been investigated~ \citep{ICML2018graph}. Also, poisoning attack against graph is also studied~ \citep{KDD2018graph}. However, they only consider the attack against graph convolutional network.
\vspace{-2ex}
\paragraph{Matrix Factorization} Skip-gram model from the NLP community has been shown to be doing implicit matrix factorization~\citep{mf2014nips}. Recently, based on the previous work, it has been shown that most of the popular unsupervised methods for graph is doing implicit matrix factorization~\citep{WSDM2018factorization}. Moreover, poisoning attack has been demonstrated for matrix factorization problem~\citep{bo2016nips}.

\section{Preliminaries}
We first introduce the graph embedding problem and link prediction problem. Then we will give an overview of the two existing algorithms for computing the embedding. Given a graph $G=(V,E)$, where $V$ is the node set and $|E|$ is the edge set, the goal of graph embedding methods is to learn a mapping from $V$ to $R^{d}$ which maps each node in the graph to a d-dimensional vector. We use $A$ to denote the adjacency matrix of the graph $G$. For each node $i$, we use $\Omega_{i}$ to denote node $i$'s neighbor. $X$ represents the learnt node embedding matrix where $X_{i}$ is the embedding of node $i$.

In link prediction, the goal is to predict the missing edges or the edges that are most likely to emerge in the future. Formally, given a set of node pair $\mathcal{T}\in V\times V$, the task is to predict a score for each node pair. In this paper, we compute the score of each edge from the cosine similarity matrix $XX^{T}$.

Now we briefly review two popular graph embedding methods: DeepWalk~\citep{perozzi2014deepwalk} and LINE~\citep{tang2015LINE}. DeepWalk extends the idea of Word2vec~\citep{word2vec} to graph, where it views each node as a word and use the generated random walks on graph as sentences. Then it uses Word2vec to get the node embeddings. LINE learns the node embeddings by keeping both the first-order proximity (LINE$_{1st}$), which describes local pairwise proximity, and the second-order proximity (LINE$_{2nd}$) for sampled node pairs.
For DeepWalk and LINE$_{2nd}$, there is a context embedding matrix computed together with the node embedding matrix. We use $Y$ to denote the context embedding matrix.

Previous work~\citep{WSDM2018factorization} has shown that DeepWalk and LINE$_{2nd}$ is implicitly doing matrix factorization.
\vspace{-1ex}
\begin{itemize}
    \item DeepWalk is solving the following matrix factorization problem:
    \vspace{-1ex}
    \begin{equation}
    \log \big(vol(G)\big(\frac{1}{ T}\sum_{i=1}^{T}(D^{-1}A)^{r}\big)D^{-1}\big)-\log b=XY^{T}
    \label{deepwalk}
    \end{equation}
    \vspace{-2ex}
    \item LINE$_{2nd}$ is solving the following  matrix factorization problem: 
    \begin{equation}
    \log \big(vol(G)D^{-1}AD^{-1}\big)-\log b=XY^{T}
    \end{equation}
\end{itemize}
\vspace{-0.5ex}
where $vol(G)=\sum A_{ij}$ is the volume of graph $G$, $D$ is the diagonal matrix where each element represents the degree of the corresponding node, $T$ is the context window size and $b$ is the number of negative samples. We use $Z$ to denote the matrix that DeepWalk and LINE$_{2nd}$ is factorizing. We denote $\Omega=\{(i,j):Z_{ij}\neq 0\}$ as the observable elements in $Z$ when solving matrix factorization and $\Omega_{i}=\{j: Z_{ij}\neq 0\}$ as the nonzero elements in row $i$ of $Z$. With these notations defined, we now give a unified formulation for DeepWalk and LINE$_{2nd}$:

\begin{equation}
    \min_{X,Y} \|R_{\Omega}(Z - XY^{T}) \|_{F}^{2}
    \label{MF}
\end{equation}
where $[\mathcal R_\Omega(\mathbf A)]_{ij}$ is $\mathbf A_{ij}$ if $(i,j)\in\Omega$ and $0$ otherwise, $\|A\|_{F}^{2}$ denotes the squared Frobenious norm of matrix $A$.

\section{Problem Statement}
In this section we introduce the attack model, including attacker's action, attacker's utilities and the constraints on the attacker. 
We assume that the attacker can manipulate the poisoned graph $G$ by adding or deleting  edges. 
In this paper, we consider these two type of manipulation: adding edges and deleting edges respectively. We use $G^{adv}$ to denote the poisoned graph.

We characterize two kinds of adversarial goals: 

\textbf{Integrity attack}: Here the attacker's goal is either to increase or decrease the probability (similarity score) of a target node pair.
For example, in social network, the attacker may be interested in increasing (or decreasing) the probability that a friendship occurs between two people. Also, in recommendation system, an attacker associated with the producer of a product may be interested to increase the probability of recommending the users with that specific product. Specifically, the attacker aims to change the probability of the edge connected with a pair of nodes whose embedding is learnt from the poisoned graph  $G^{adv}$.

For integrity attack, we consider two kinds of constraints on the attacker: 1. \emph{Direct Attack}: the attacker can only manipulate  edges adjacent to the target node pair; 2. \emph{Indirect Attack}: the attacker can only manipulate edges without connecting to the target node pair.

\textbf{Availability attack}
Here the adversarial goal of availability attack is to reduce the prediction performance over a test set consisting of a set of node pairs $\mathcal{T} \in V\times V$. 
(Here $\mathcal{T}$ consists of both positive examples $\mathcal{T}_{pos}$ indicating the existence of edges, and negative examples $\mathcal{T}_{neg}$ for the absence of edges). In this paper, we choose average precision score (AP score) to evaluate the attack performance. Specifically, we consider the attacker whose goal is to decrease the AP score over $\mathcal{T}$ by adding small perturbation to a given graph.

\section{Attacking Unsupervised Graph Embedding}
In this section we show our algorithm for computing the adversarial strategy. Given that DeepWalk and LINE is implicitly doing matrix factorization, we can directly derive the  back-propogated gradient based on the first order KKT condition. Our algorithm has two steps: 1. Projected Gradient Descent (PGD) step: gradient descent on the weighted adjacency matrix. 2. Projection step: projection of weighted adjacency matrix onto $\{0,1\}^{|V|\times |V|}$. We first describe the projected gradient descent step and then we describe the projection method we use to choose which edge to add or delete.

\subsection{Projected Gradient Descent (PGD)}\label{pgd}
Based on the matrix factorization formulation above, we describe the algorithm we use to generate the adversarial graph. The core part of our method is \textit{projected gradient descent} (PGD) step. In this step, the adjacency matrix is continuous since we view the graph as a weighted graph, which allows us to use gradient descent.

First we describe the loss function we use. We use $L(X)$ to denote the loss function. For integrity attack, $L$ is $\pm [XX^{T}]_{ij}$ where $(i,j)$ is the target node pair. (Here the $+$ or $-$ sign depends on whether the attacker wants to increase or decrease the score of the target edge.) For availability attack, the loss function is $\sum_{(i,j)\in T_{pos}} [XX^{T}]_{ij} - \sum_{(i,j) \in T_{neg}}[XX^{T}]_{ij}$. The update of the weighted adjacency matrix $A$ in iteration $t$ is as follows:
\begin{equation}
A^{t+1} = \mathrm{Proj}_{\mathbb{A}}(A^{t}-s_{t}\cdot \nabla_{A}L)
\end{equation}
Here $\mathrm{Proj}$ is the projection function which projects the matrix to $[0,1]$ space and $s_{t}$ is the step size in iteration $t$.  The non-trivial part is to compute $\nabla_{A}L$. We note that 
\begin{equation} 
\nabla_{A}L = \nabla_{X}L\cdot \nabla_{A}X
\end{equation}
The computation of $\nabla_{X}L$ is trivial. Now to compute $\nabla_{A}X$, using the chain rule,  we have: $\nabla_{A}X=\nabla_{Z}X\cdot \nabla_{A}Z$. First we show how to compute $\nabla_{Z}X$.

For the matrix factorization problem defined in Eq.~\ref{MF}, using the KKT condition, we have:
\begin{equation}
    \sum_{j\in \Omega_{i}} (Z_{ij}-X_{i}Y_{j}^{T})Y_{j}=0
\end{equation}
\begin{equation}
    \frac{\partial X_{i}}{\partial Z_{ij}} = (\sum_{j'\in \Omega_{i}} Y_{j'} Y_{j'}^{T})^{-1}Y_{j}
    \label{grad_Z}
\end{equation}

Next we show how to compute $\nabla_{A}Z$ for DeepWalk and LINE separately. In the derivation of $\nabla_{A}Z$, we view $vol(G)$ and $D$ as constant. 

\textbf{DeepWalk} We show how to compute $\nabla_{A}Z$  for DeepWalk. For DeepWalk, From Eq.~\ref{deepwalk}:
   \begin{equation}
    Z = \log \big(vol(G)\big(\frac{1}{ T}\sum_{i=1}^{T}(D^{-1}A)^{r}\big)D^{-1}\big)-\log b
    \end{equation}
Now to compute $\nabla_{A}Z$, let $P=D^{-1}A$, then we only need to derive $\nabla_{P}P^{r}$ for each $r\in [1,T]$. Note that $(P^{r})'=\sum_{k=1}^{r}P^{k-1}P'P^{r-k}$. Then since computing $\nabla_{A}P$ and $\nabla_{P}{Z}$ is easy, once we have $\nabla_{P}P^{r}$, we can compute $\nabla_{A}Z$.

\textbf{LINE} We show the derivation of $\nabla_{A}Z$ for LINE$_{2nd}$:
\begin{equation}
    Z = \log(vol(G)D^{-1}AD^{-1}) - \log b
\end{equation}
since $Z_{ij} = \log(vol(G)d_{i}^{-1}A_{ij}d_{j}^{-1})-\log b$ where $d_{i}=D_{ii}$. We have:
\begin{equation}
    \frac{\nabla Z_{ij}}{\nabla A_{ij}} = \frac{1}{A_{ij}}
    \label{grad_A}
\end{equation}
Once we have $\nabla_{Z}L$ and $\nabla_{L}Z$, we can compute $\nabla_{A}X$. 

\subsection{Projection}
 In the Projected Gradient Descent step, we compute a weighted adjacency matrix $A^{opt}$. We use $A^{org}$ to denote the adjacency matrix of the clean graph. Therefore we need to project it back to $\{0,1\}^{|V|\times|V|}$. Now we describe the projection method we use, which is  straightforward. First, we show our projection method for an attacker that can add edges. To add edges, the attacker needs to choose some cells in $S=\{(i,j)\mid A^{org}_{ij}=0\}$ and turn it into 1. Our projection strategy is that the attacker  chooses the cells $(i,j)$ in $S$ where $A^{opt}_{ij}$ is closest to 1 as the candidate set of edges to add. For deleting edges, it works in a similar way. The only difference is that we start from the cells that are originally 1 in $A^{org}$ and choose the cells that are closest to 0 in $A^{opt}$ as the candidate set of edges to delete. 

Now we briefly discuss when to use the projection step. A natural choice is to project once after the projected gradient descent step. Another choice is to incorporate the projection step into the gradient descent computation where we project every $k$ iterations. The second projection strategy induces less loss in the projection step and  is more accurate for computation but can take more iterations to converge than the first projection strategy. In our experiments, we choose the first projection strategy for its ease of computation.
\section{Experiments}
In this section, we show the results of poisoning attack against  DeepWalk and LINE on real-world graph datasets. 
In the experiments, we denote our poisoning attack method as `Opt-attack'. 
We evaluate our attack method on three real-world graph datasets: 1). Facebook~\citep{facebook}: a social networks with 4039 nodes and 88234 edges. 2). Cora~\citep{cora}: a citation network with 2708 nodes and 2708 edges. 3). Citeseer~\citep{citeseer}: a citation network with 2110 nodes and 7336 edges. 

\noindent \textbf{Baselines} We compare with several baselines: (1) \textit{random attack}: this baseline is general and used for both integrity attack and availability attack. We randomly add or remove edges; (2) \textit{personalized PageRank}~\citep{bahmani2010fast}: this baseline is only used for integrity attack. Given a target edge $(A,B)$, we use personalized PageRank to calculate the importance of the nodes. Given a list of nodes ranked by their importance, e.g., $(x_1, x_2, x_3, x_4, ...)$, we select the edges which connect the top ranked nodes to A or B, i.e., $(A,x_1), (B,x_1), (A,x_2), (B, x_2), ...$; (3) \textit{degree sum}: this baseline is used for availability attack. We rank the node pair by the sum of degree of its two nodes. Then we add or delete the node pair with the largest degree sum.
(4) \textit{shortest path}: this baseline is used for availability attack. We rank the edge by the number of times that it is on the shortest paths between two nodes in graph. Then we delete the important edges measured by the number of shortest paths in graph that go through this edge.

\vspace{-0.5ex}

In our experiments, we choose 128 as the latent embedding dimension. We use the default parameter settings for DeepWalk and LINE. 
We generate the test set and validation set with a proportion of 2:1 where positive examples are sampled by removing 15\% of the edges from the graph and negative examples are got by sampling an equal number of node pairs from the graph which has no edge connecting them.
For both our attack and random attack, we guarantee that the attacker can't directly modify any node pairs in the target set.

\subsection{Integrity Attack}\label{integrity}

For each attack scenario, we choose 32 different target node pairs and use our algorithm to generate the adversarial graph. For increasing the score of the target edge, the target node pair is randomly sampled from the negative examples in the test set. For decreasing the score of the target edge, the target node pair is randomly sampled from the positive examples in the test set.  We report the average score increase and compare it with the random attack baseline. We consider two kinds of attacker's actions: adding or deleting edges and two constraints: direct attack and indirect attack. Results for Citeseer dataset are deferred to the appendix.
\vspace{-2ex}

\begin{figure*}[!htbp]
\centering
\begin{minipage}{.245\textwidth}
 \begin{subfigure}{\textwidth}
 \centering
 \includegraphics[width=\textwidth]{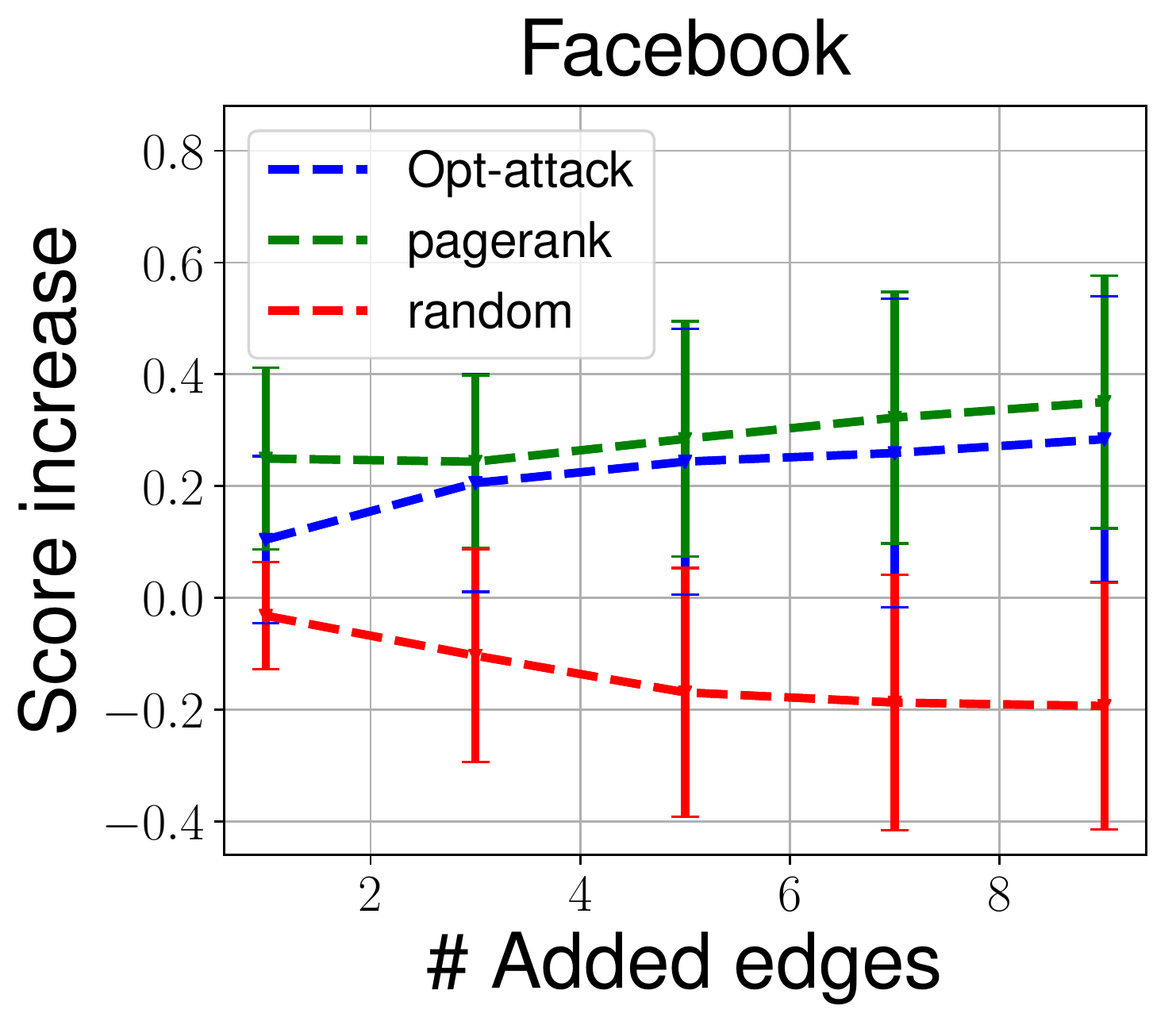}
 \caption{DeepWalk-Add-Up}
 \label{1:a}
 \end{subfigure}
\end{minipage}
\begin{minipage}{.245\textwidth}
 \begin{subfigure}{\textwidth}
 \centering
 \includegraphics[width=\textwidth]{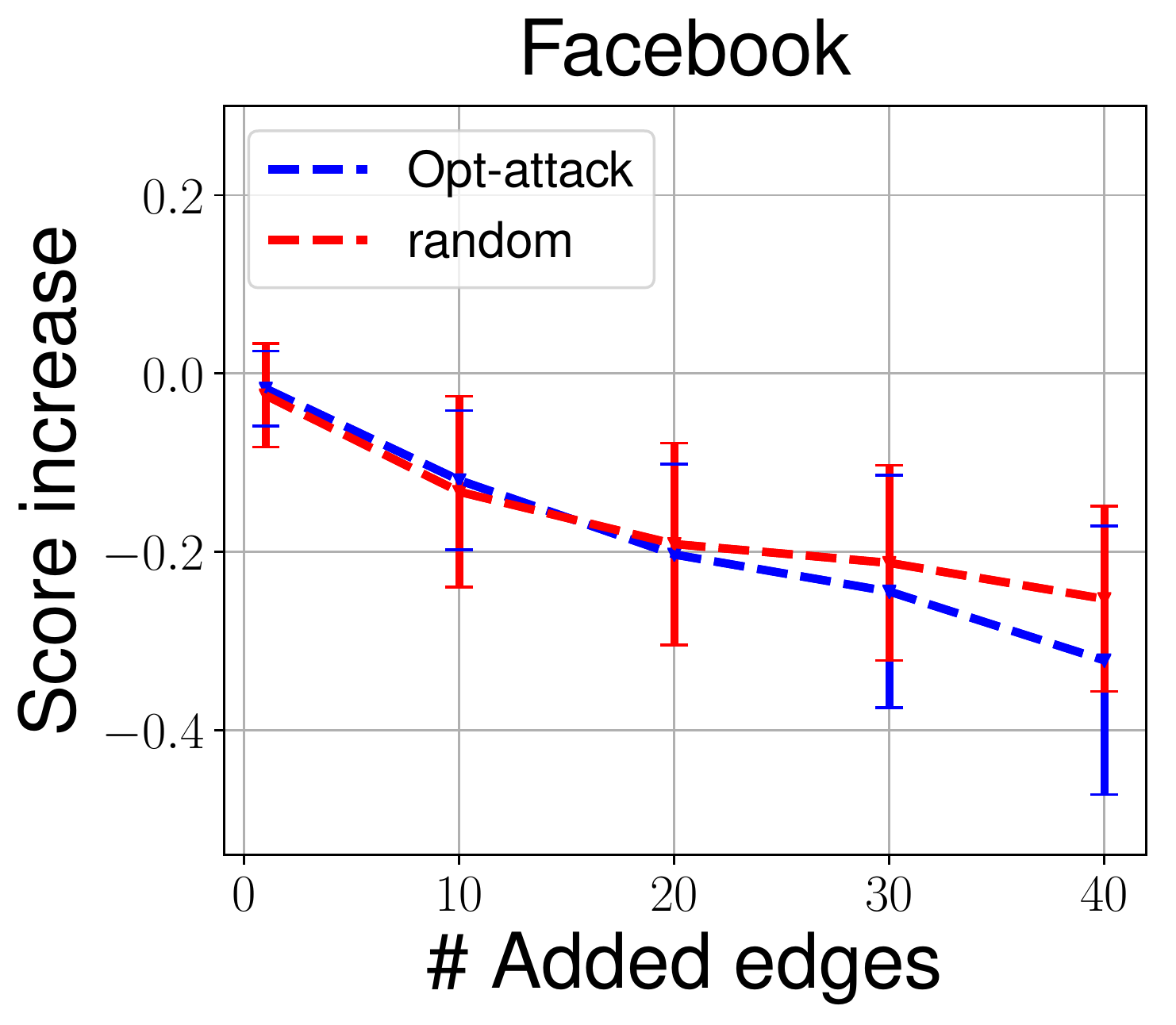}
 \caption{DeepWalk-Add-Down}
 \label{1:b}
 \end{subfigure}
\end{minipage}
\begin{minipage}{.245\textwidth}
 \begin{subfigure}{\textwidth}
 \centering
 \includegraphics[width=\textwidth]{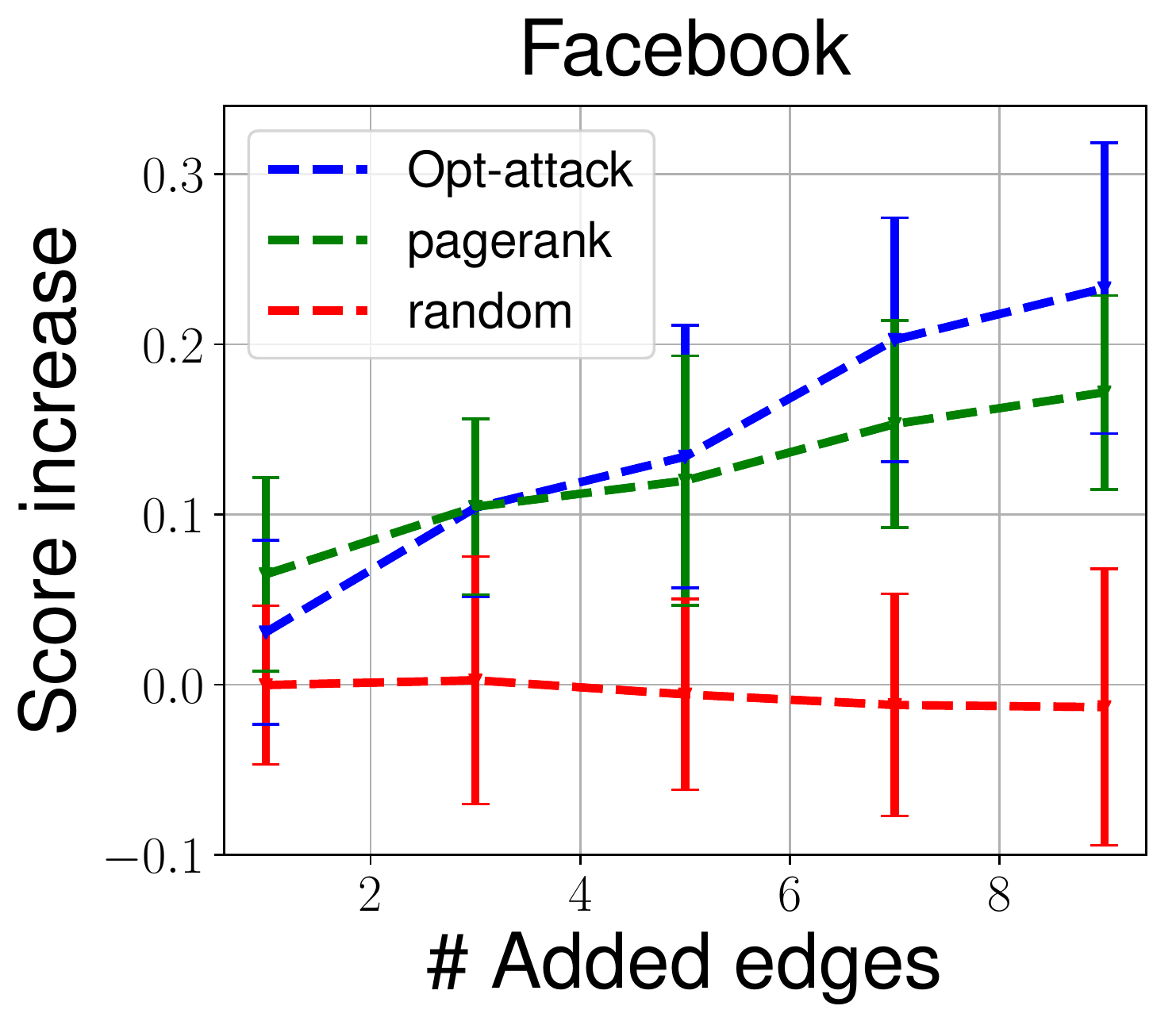}
 \caption{LINE-Add-Up}
 \label{1:c}
 \end{subfigure}
\end{minipage}
\begin{minipage}{.245\textwidth}
 \begin{subfigure}{\textwidth}
 \centering
 \includegraphics[width=\textwidth]{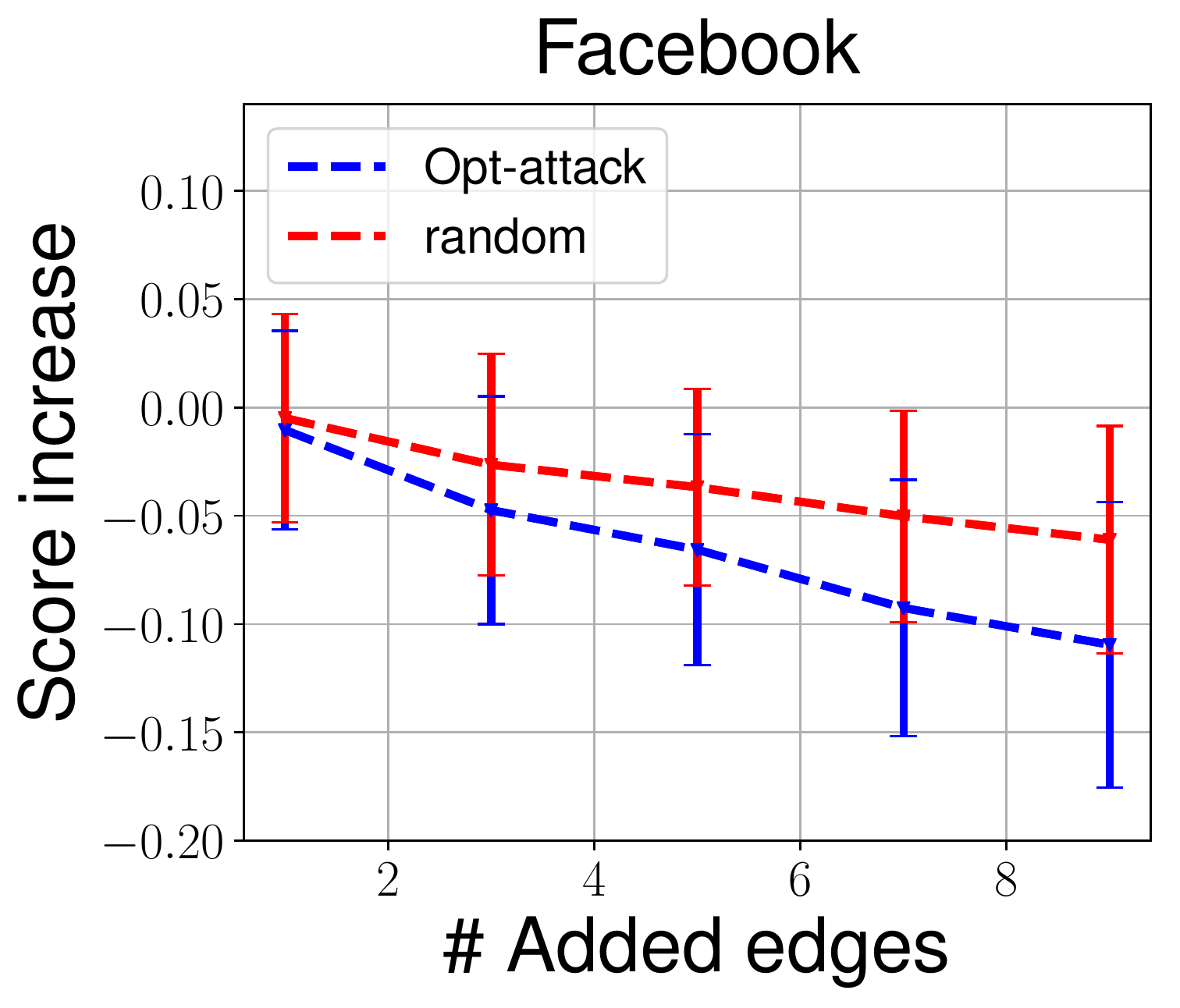}
 \caption{LINE-Add-Down}
 \label{1:d}
 \end{subfigure}
\end{minipage}
\begin{minipage}{.245\textwidth}
 \begin{subfigure}{\textwidth}
 \centering
 \includegraphics[width=\textwidth]{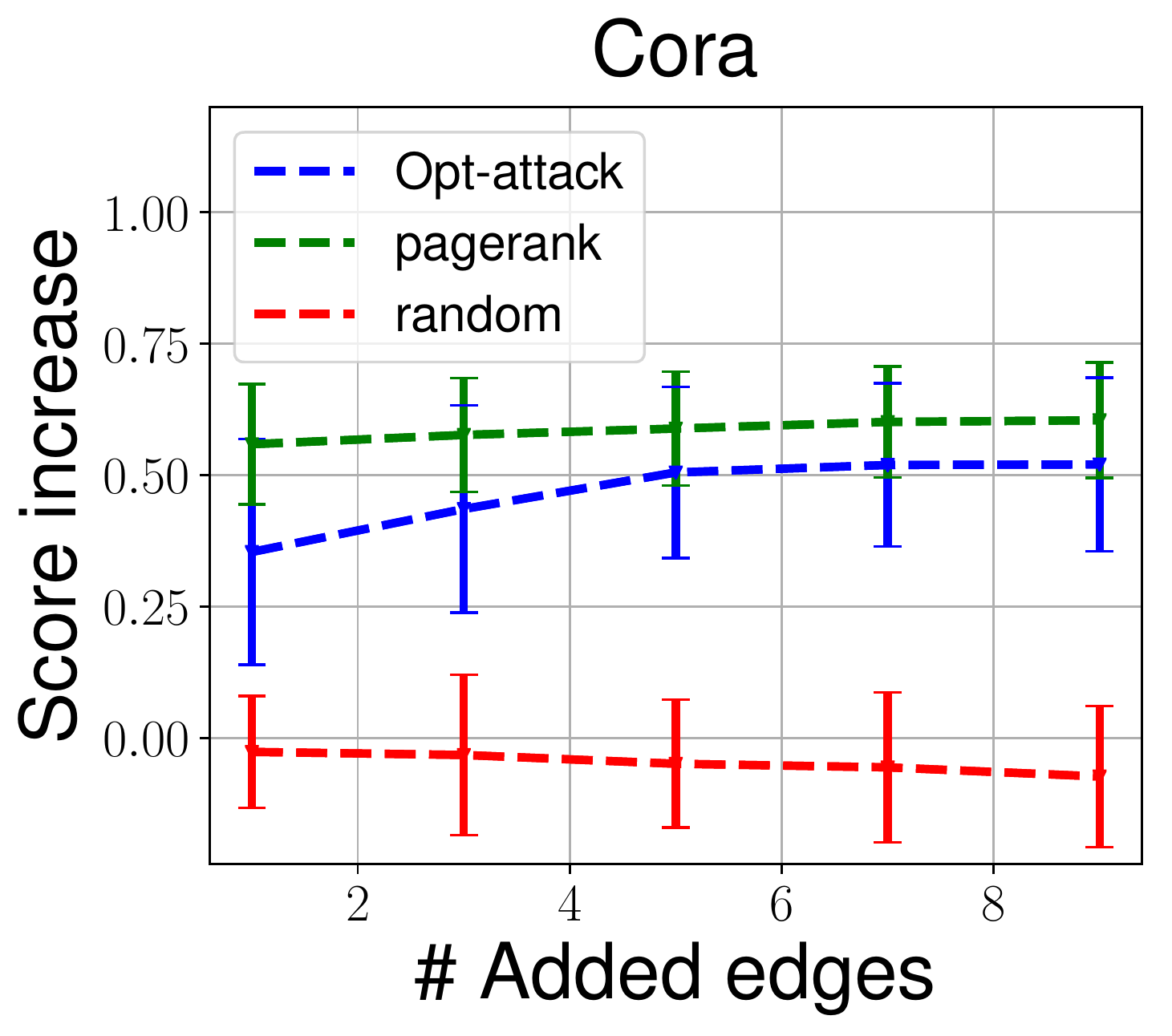}
 \caption{DeepWalk-Add-Up}
 \label{1:e}
 \end{subfigure}
\end{minipage}
\begin{minipage}{.245\textwidth}
 \begin{subfigure}{\textwidth}
 \centering
 \includegraphics[width=\textwidth]{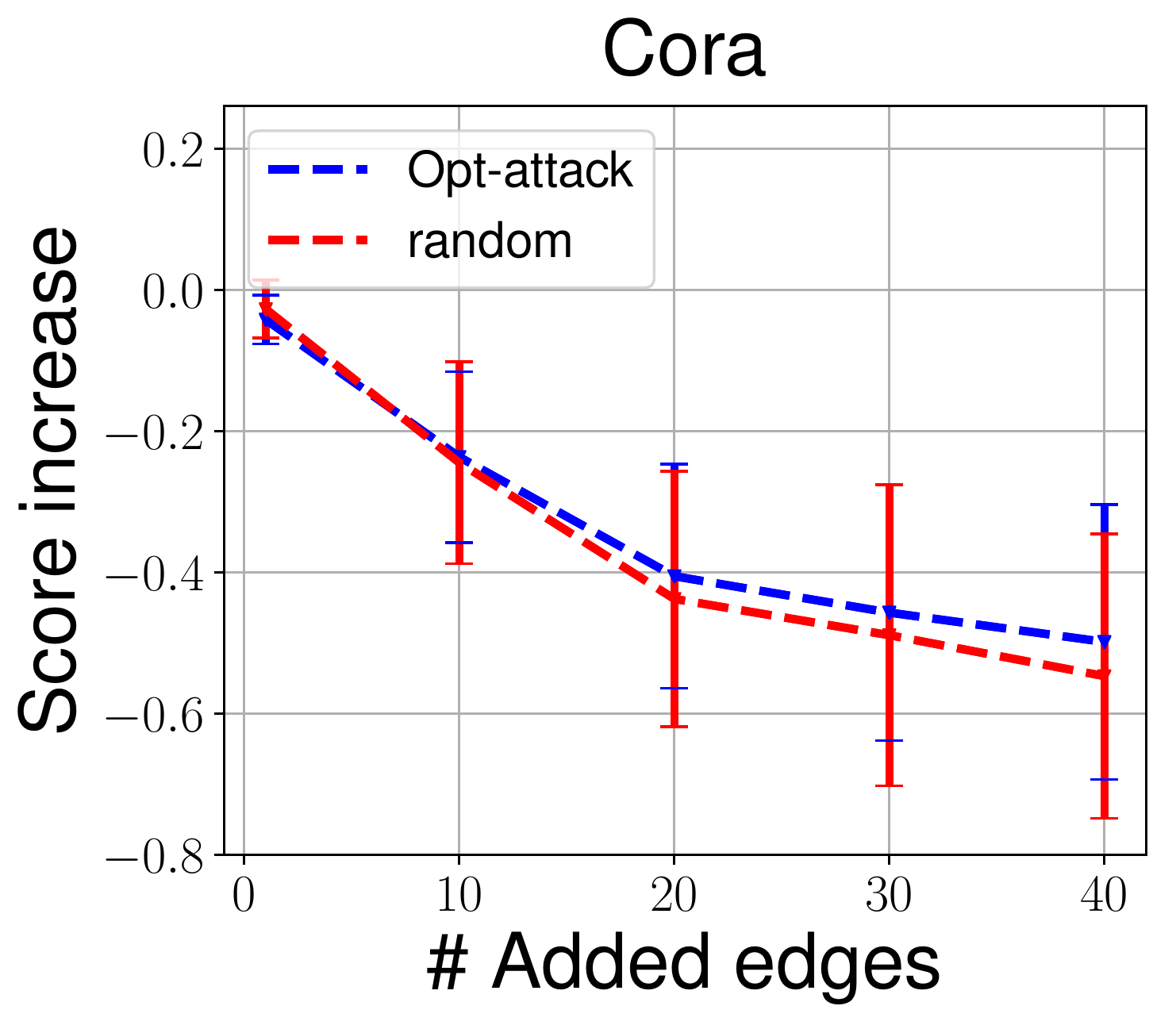}
 \caption{DeepWalk-Add-Down}
 \label{1:f}
 \end{subfigure}
\end{minipage}
\begin{minipage}{.245\textwidth}
 \begin{subfigure}{\textwidth}
 \centering
 \includegraphics[width=\textwidth]{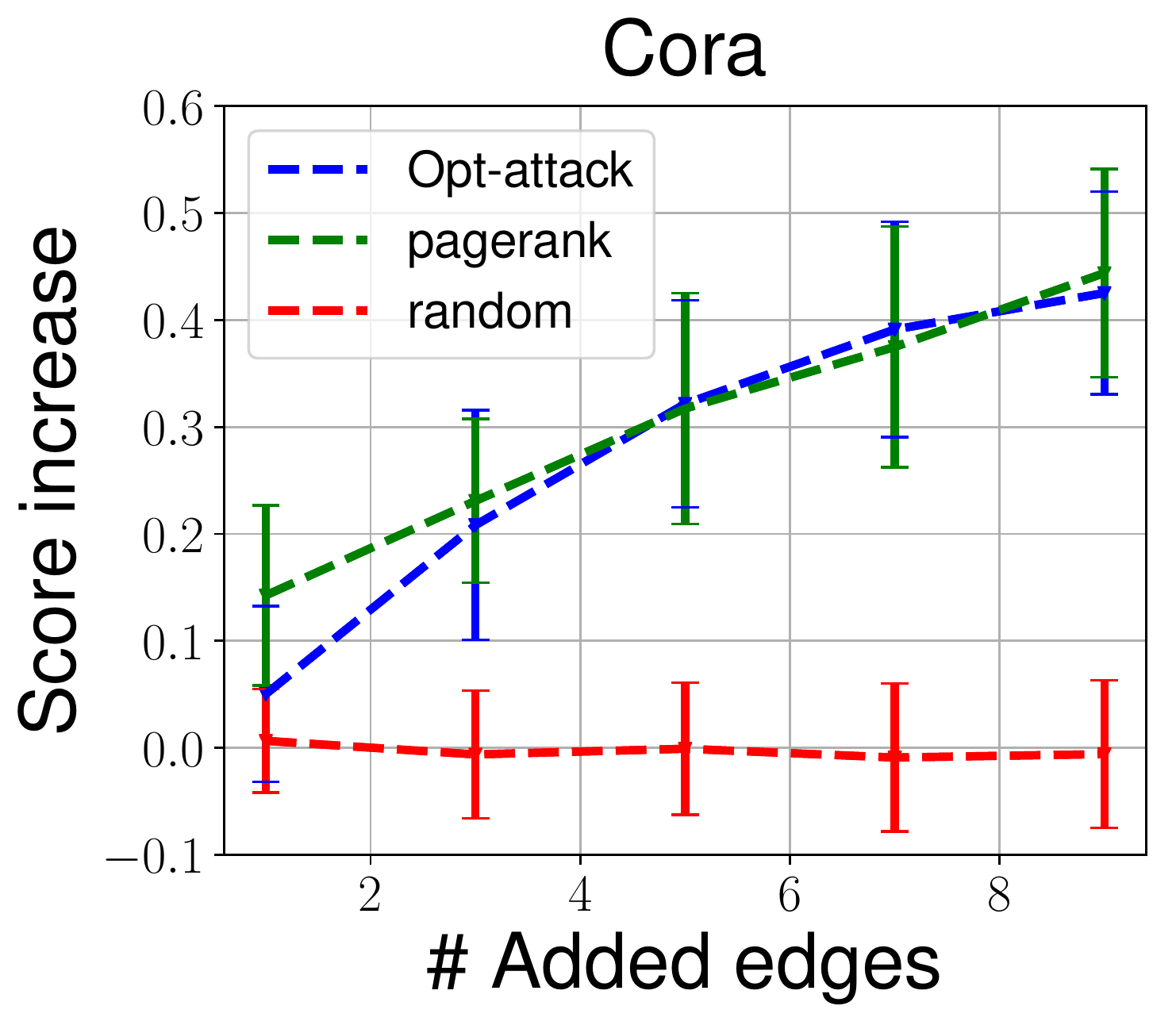}
 \caption{LINE-Add-Up}
 \label{1:g}
 \end{subfigure}
\end{minipage}
\begin{minipage}{.245\textwidth}
 \begin{subfigure}{\textwidth}
 \centering
 \includegraphics[width=\textwidth]{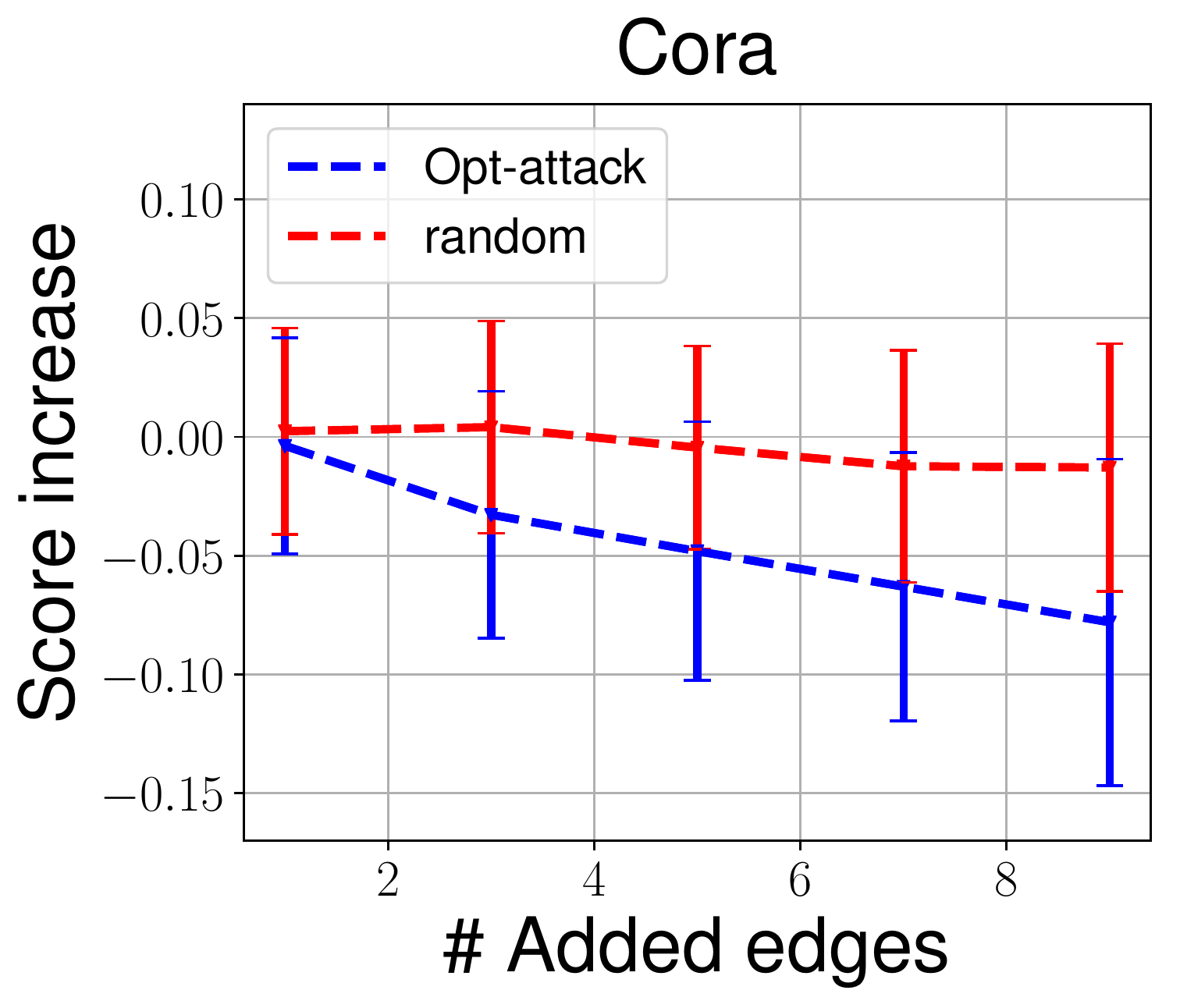}
 \caption{LINE-Add-Down}
  \label{1:h}
 \end{subfigure}
\end{minipage}
\caption{Result for direct integrity attack on two datasets. The first line contains the results for Facebook dataset. The second line contains the results for Cora dataset. The format ``Method | Type |Direction '' is used to label the each sub-caption. ``Method'' refers to embedding method while ``Type'' refers to adding or deleting edges to poison the graph. ``Direction'' refers to increasing or decreasing the similarity score of the target node pair. This notation is also used in Figure~\ref{influence-add},~\ref{target-delete},~\ref{influence}.}
\label{integrity-add}
\vspace{-3ex}
\end{figure*}

\paragraph{Adding edges} We consider the adversary which can only add edges.  Figure~\ref{integrity-add} shows the results under direct attack setting. In Figure~\ref{1:a}~\ref{1:c}~\ref{1:e}~\ref{1:g}, when the adversarial goal is to increase the score of the target node pair, we find that our attack method outperforms the random attack baseline by a significant margin. We also plot the personalized pagerank baseline. We can see this second baseline we propose is a very strong baseline, with attack performance the same level as our proposed method. To further understand it, we analyze the edges we add in this attack scenario. We find the following pattern: if the target node pair is $(i,j)$, then our attack tends to add edges from node $i$ to the neighbors of node $j$ and also from node $j$ to the neighbors of node $i$. This is intuitive because connecting to other node's neighbors can increase the similarity score of two nodes. In Figure~\ref{1:d}~\ref{1:h}, when the adversarial goal is to decrease the score of target node pair, our method is better than the random attack baseline for attacking LINE. For attacking DeepWalk (figure~\ref{1:b}~\ref{1:f}), the algorithm is able to outperform, on Facebook dataset (figure~\ref{1:b}), our attack is better than the random baseline when the number of added edges is large. Although for Cora (figure~\ref{1:f}) our attack is close to the random attack baseline, we note that in this attack case, random attack is already powerful and can lead to large drop (e.g. 0.8) in similarity score. 

\begin{figure*}[!htbp]
\centering
\begin{minipage}{.245\textwidth}
 \begin{subfigure}{\textwidth}
 \centering
 \includegraphics[width=\textwidth]{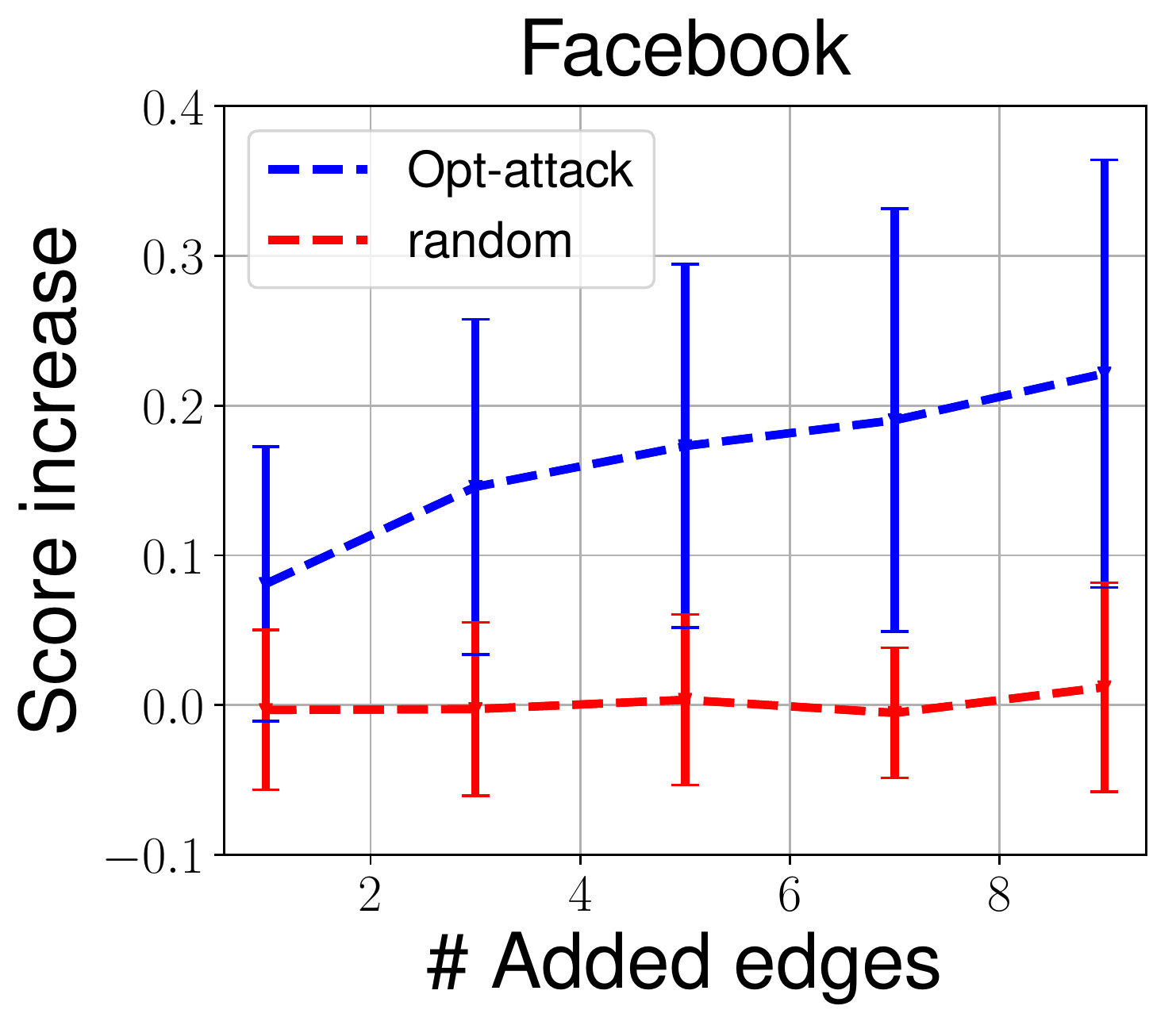}
 \caption{DeepWalk-Add-Up}
 \label{fig:fig3a}
 \end{subfigure}
\end{minipage}
\begin{minipage}{.245\textwidth}
 \begin{subfigure}{\textwidth}
 \centering
 \includegraphics[width=\textwidth]{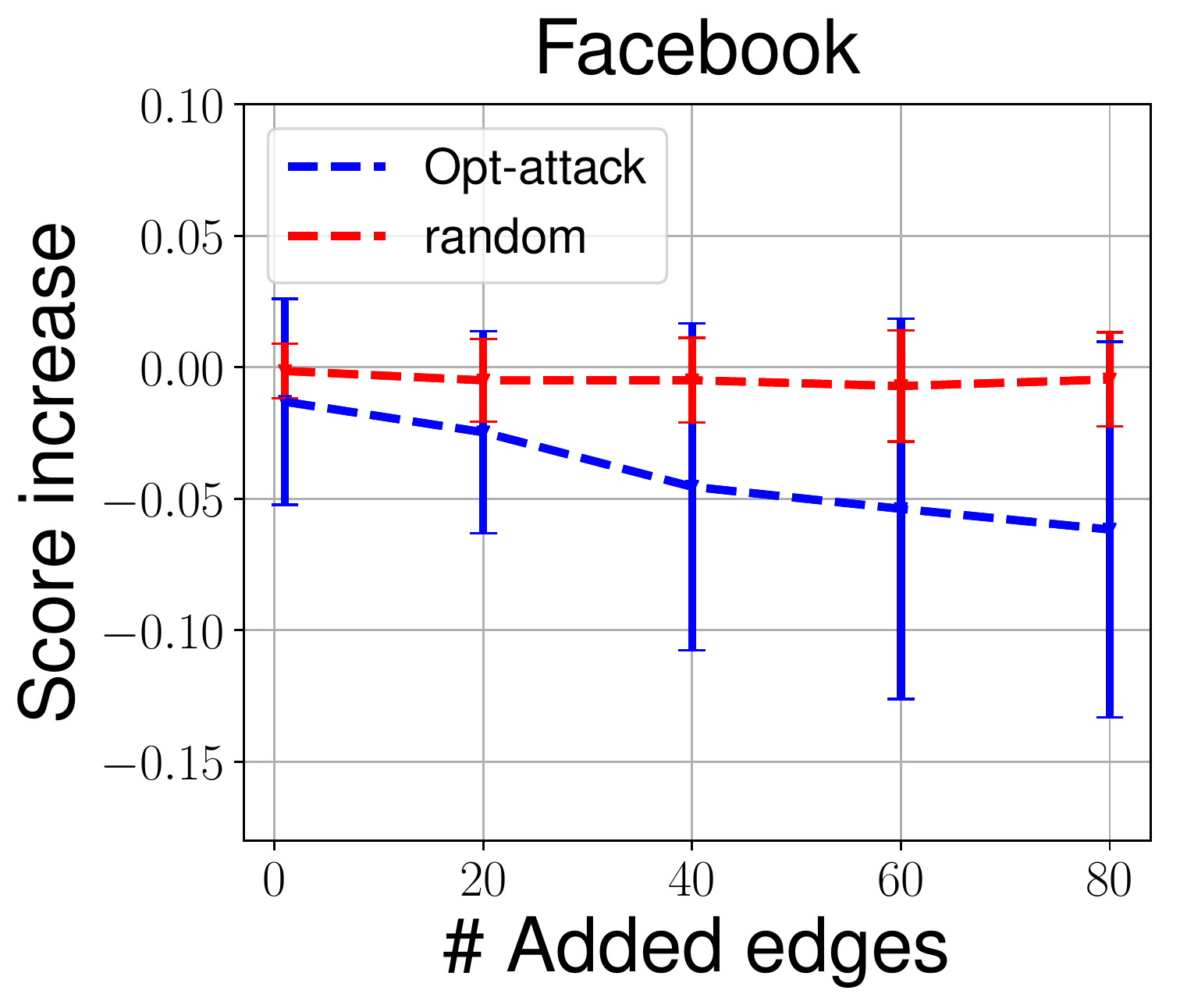}
 \caption{DeepWalk-Add-Down}
 \label{fig:fig3b}
 \end{subfigure}
\end{minipage}
\begin{minipage}{.245\textwidth}
 \begin{subfigure}{\textwidth}
 \centering
 \includegraphics[width=\textwidth]{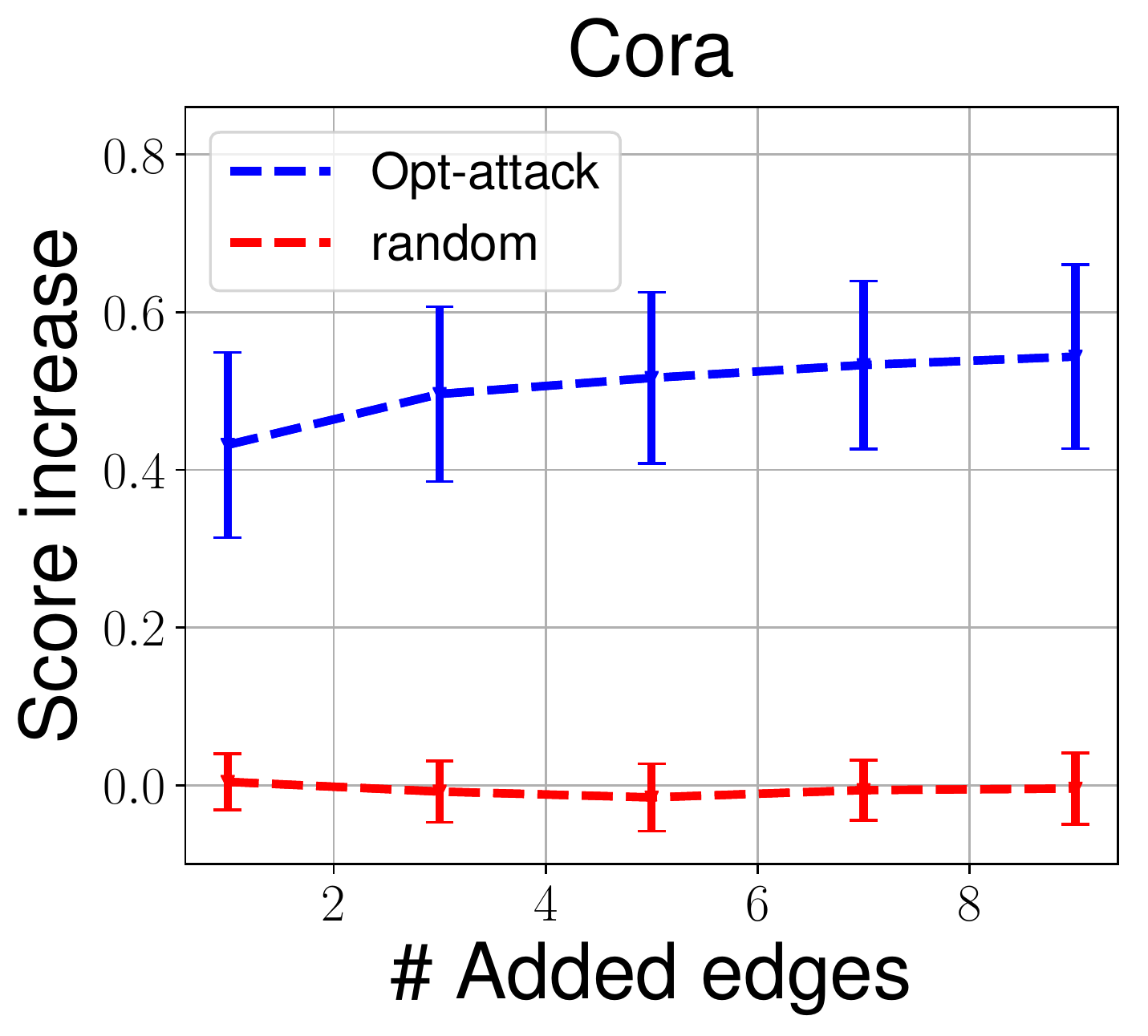}
 \caption{DeepWalk-Add-Up}
 \label{fig:fig3c}
 \end{subfigure}
\end{minipage}
\begin{minipage}{.245\textwidth}
 \begin{subfigure}{\textwidth}
 \centering
 \includegraphics[width=\textwidth]{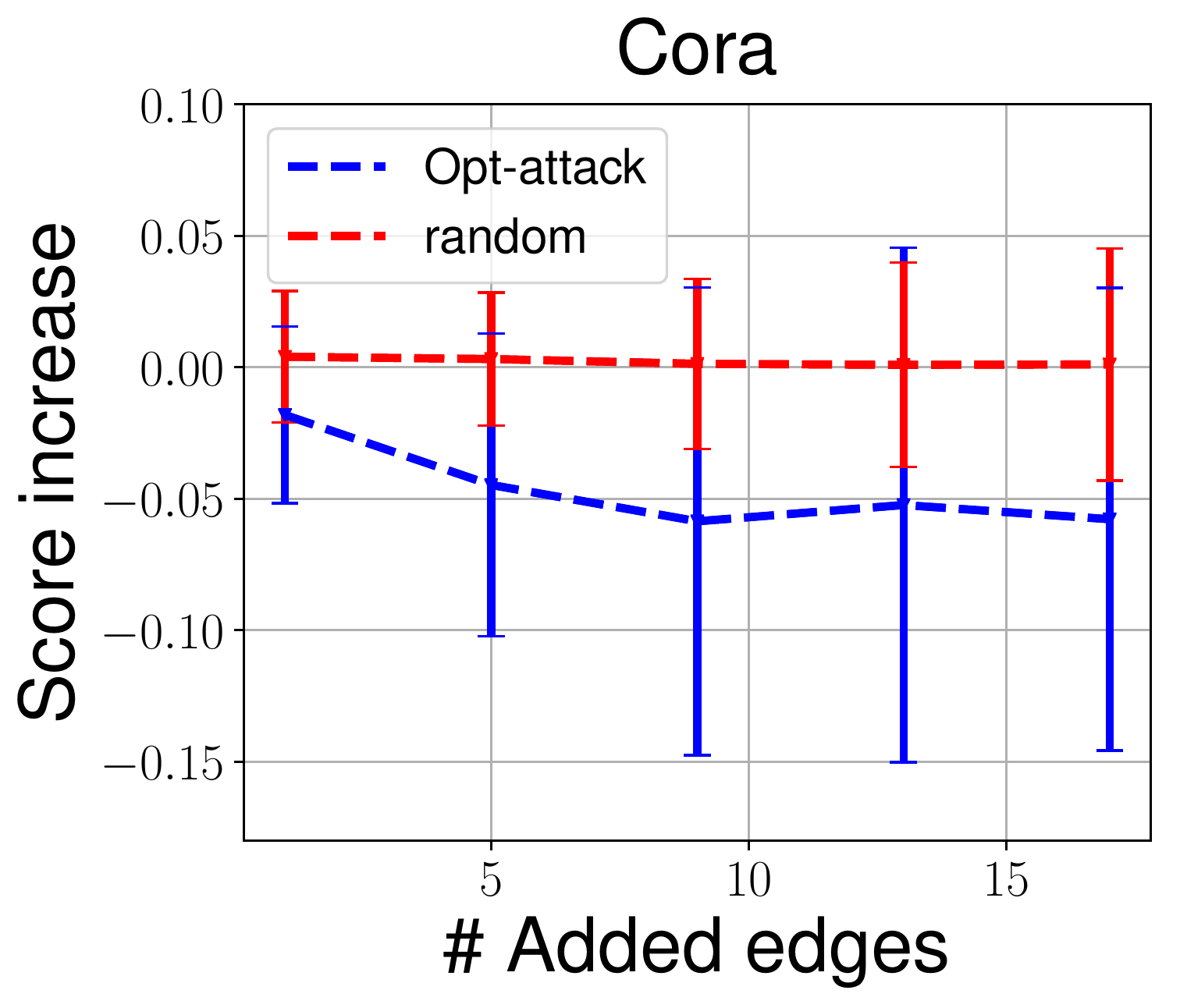}
 \caption{DeepWalk-Add-Down}
 \label{fig:fig3d}
 \end{subfigure}
\end{minipage}
\caption{Result for indirect integrity attack against DeepWalk on two datasets where the action of attacker is adding edges.}
\label{influence-add}
\end{figure*}

Figure~\ref{influence-add} shows our result of indirect attack. We can see that for DeepWalk, even if the attacker can't modify the edges adjacent to the target node pair, it can still manipulate the score of the target edge with a few edges added. We also analyze the edges our algorithm chooses when the adversarial goal is to increase the score of the target node pair $(i,j)$ (the case in figure~\ref{fig:fig3a}~\ref{fig:fig3c}), we find that our attack tends to add edges between the neighbors of node $i$ and the neighbors of node $j$. It also follows the intuition that connecting the neighbor of two nodes can increase the similarity of two nodes. When the goal is to decrease the score (figure~\ref{fig:fig3b}~\ref{fig:fig3d}), our attack is still better than random baseline by a noticeable margin.

\begin{figure*}[!htbp]
\centering
\begin{minipage}{.245\textwidth}
 \begin{subfigure}{\textwidth}
 \centering
 \includegraphics[width=\textwidth]{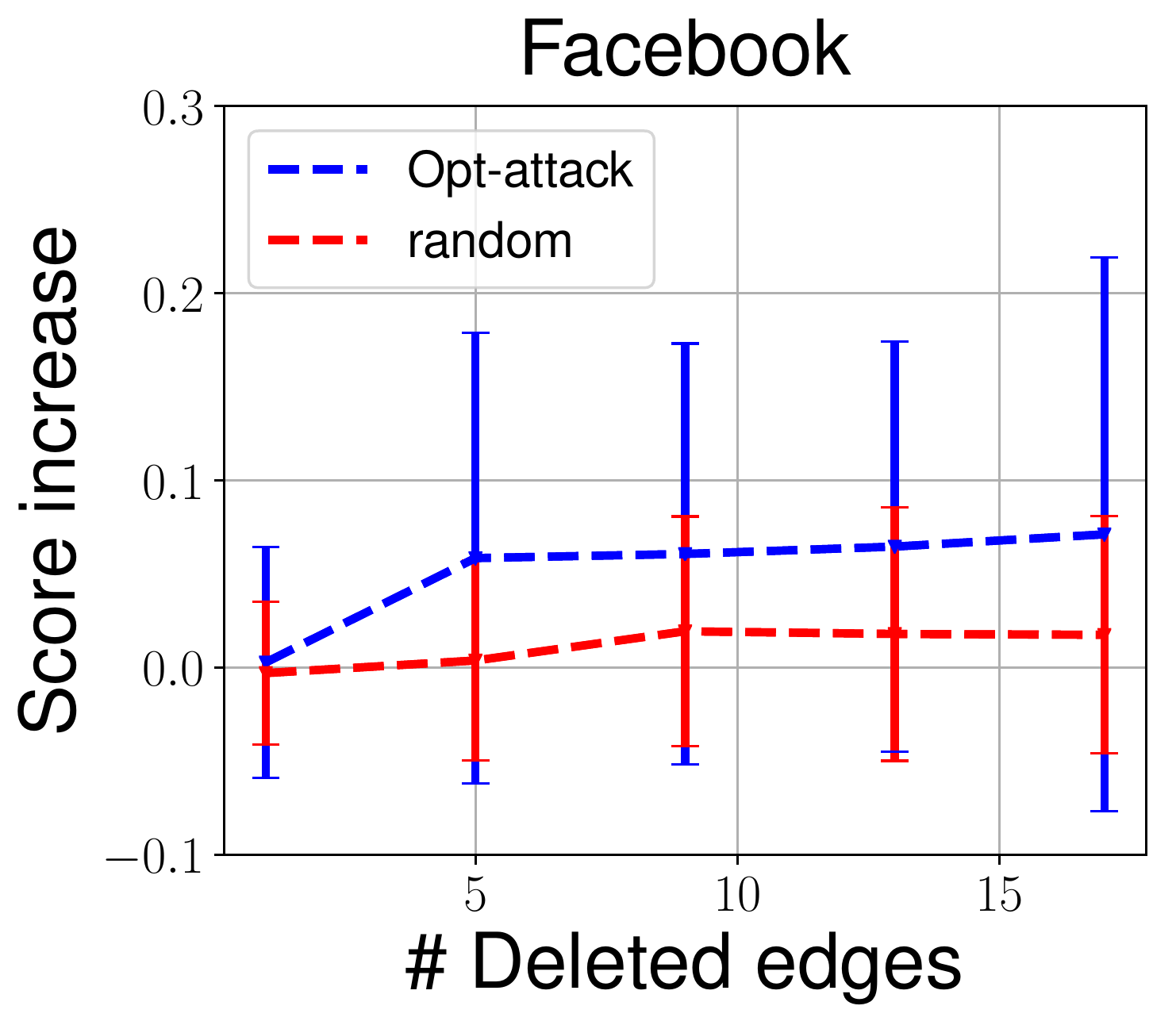}
 \caption{DeepWalk-Del-Up}
 \label{fig:4a}
 \end{subfigure}
\end{minipage}
\begin{minipage}{.245\textwidth}
 \begin{subfigure}{\textwidth}
 \centering
 \includegraphics[width=\textwidth]{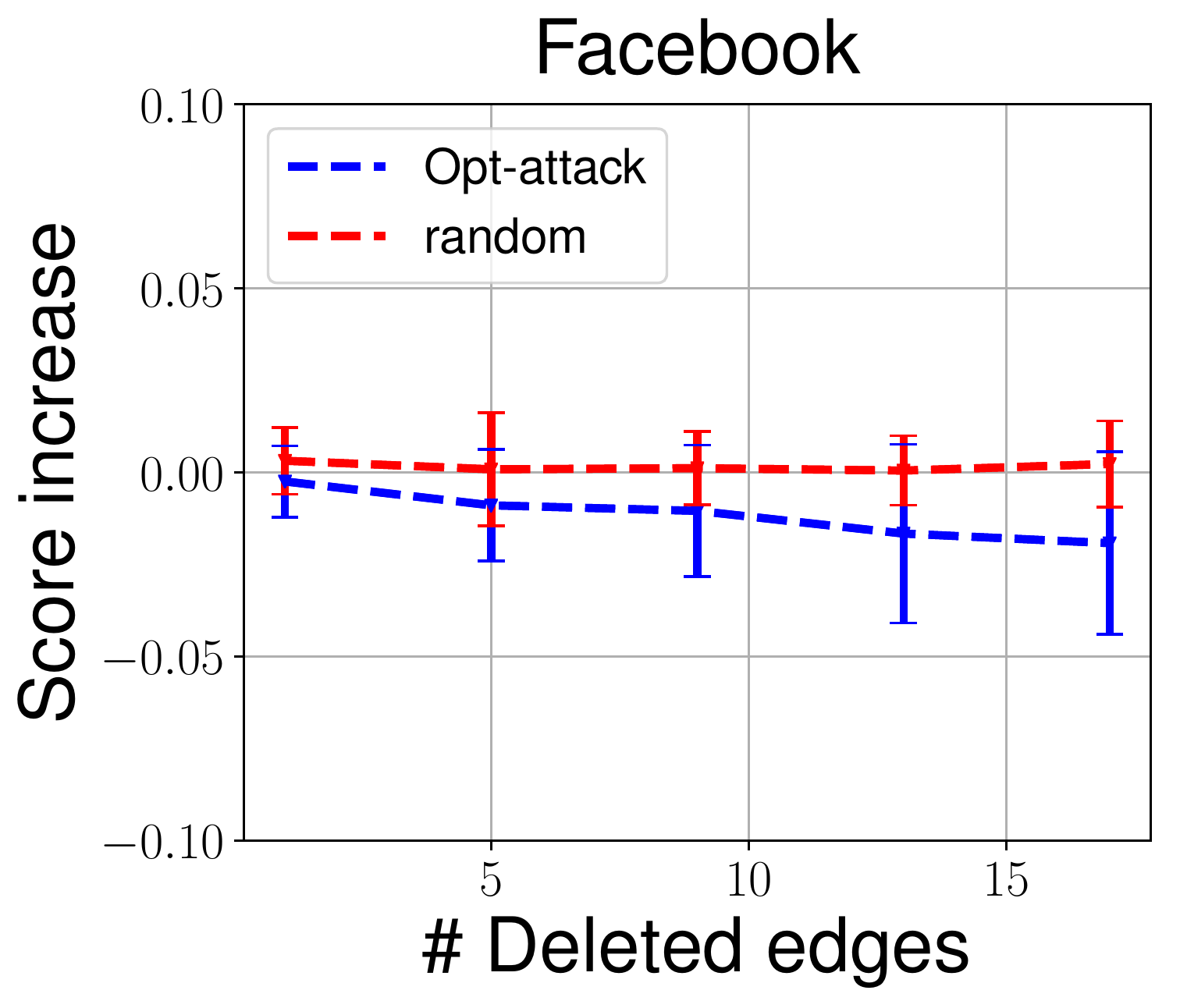}
 \caption{DeepWalk-Del-Down}
 \label{fig:4b}
 \end{subfigure}
\end{minipage}
\begin{minipage}{.245\textwidth}
 \begin{subfigure}{\textwidth}
 \centering
 \includegraphics[width=\textwidth]{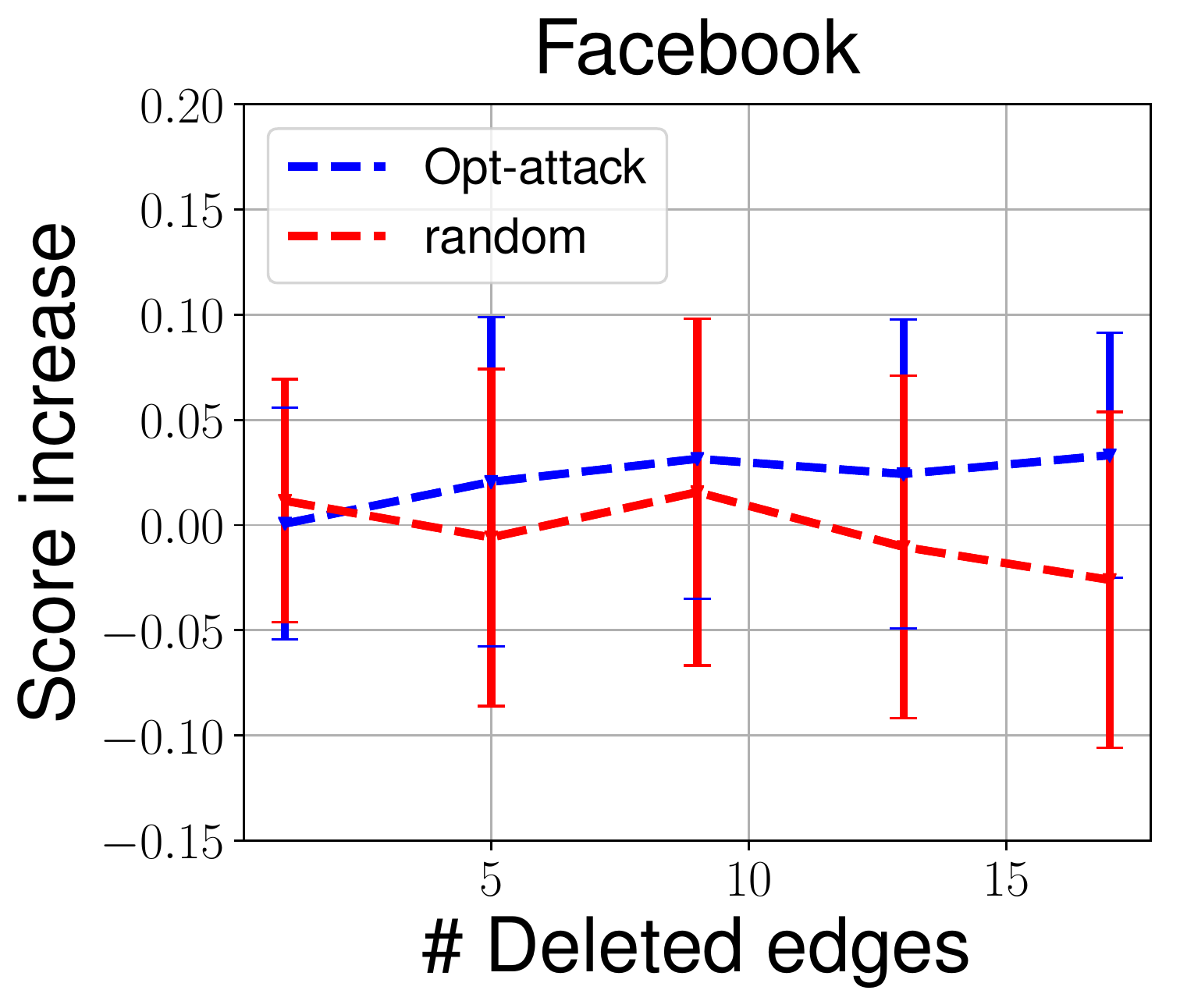}
 \caption{LINE-Del-Up}
 \label{fig:4c}
 \end{subfigure}
\end{minipage}
\begin{minipage}{.245\textwidth}
 \begin{subfigure}{\textwidth} 
 \centering
 \includegraphics[width=\textwidth]{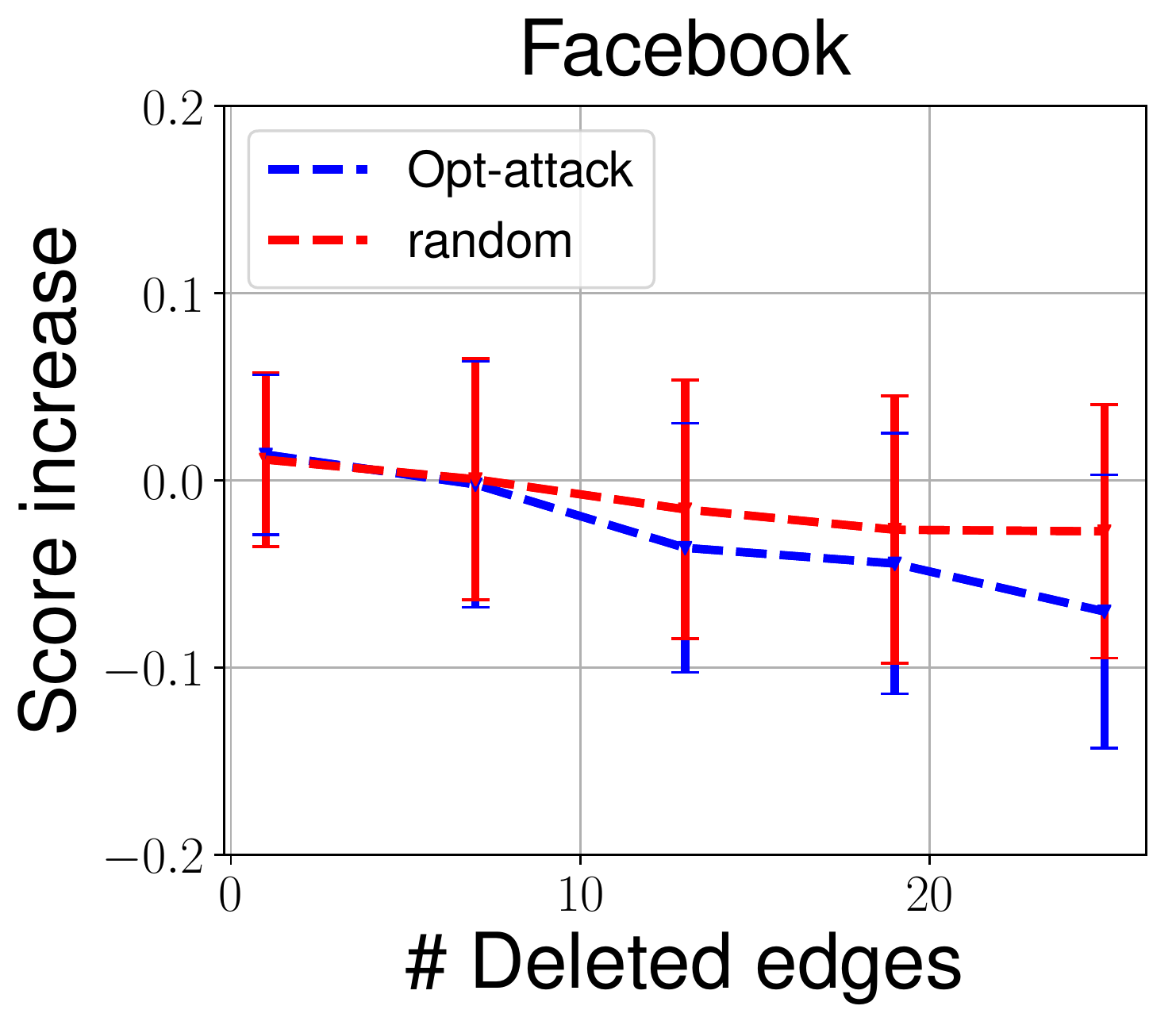}
 \caption{LINE-Del-Down}
 \label{fig:4d}
 \end{subfigure}
\end{minipage}
\begin{minipage}{.245\textwidth}
 \begin{subfigure}{\textwidth}
 \centering
 \includegraphics[width=\textwidth]{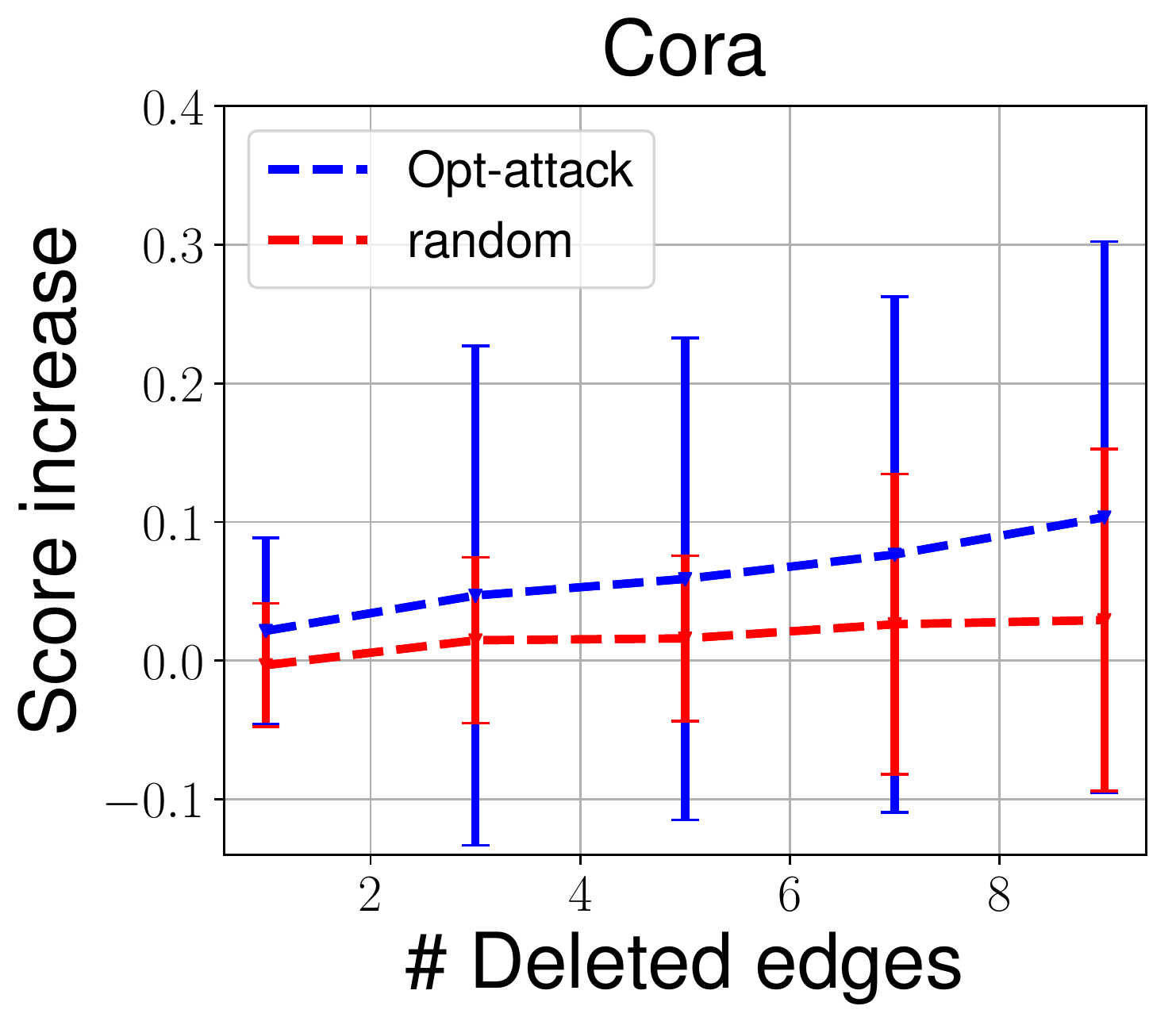}
 \caption{DeepWalk-Del-Up}
 \label{fig:4e}
 \end{subfigure}
\end{minipage}
\begin{minipage}{.245\textwidth}
 \begin{subfigure}{\textwidth}
 \centering
 \includegraphics[width=\textwidth]{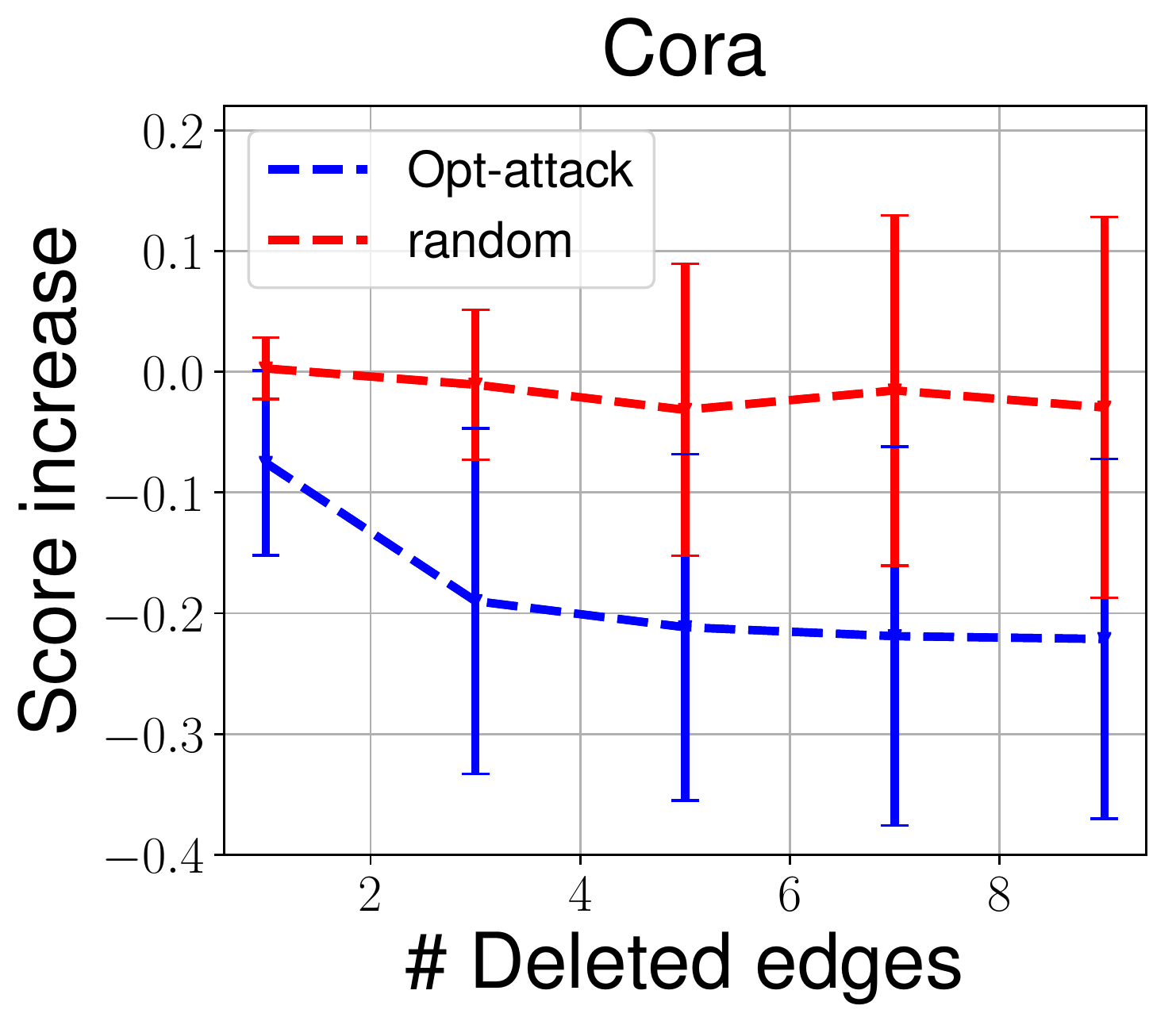}
 \caption{DeepWalk-Del-Down}
 \label{fig:4f}
 \end{subfigure}
\end{minipage}
\begin{minipage}{.245\textwidth}
 \begin{subfigure}{\textwidth}
 \centering
 \includegraphics[width=\textwidth]{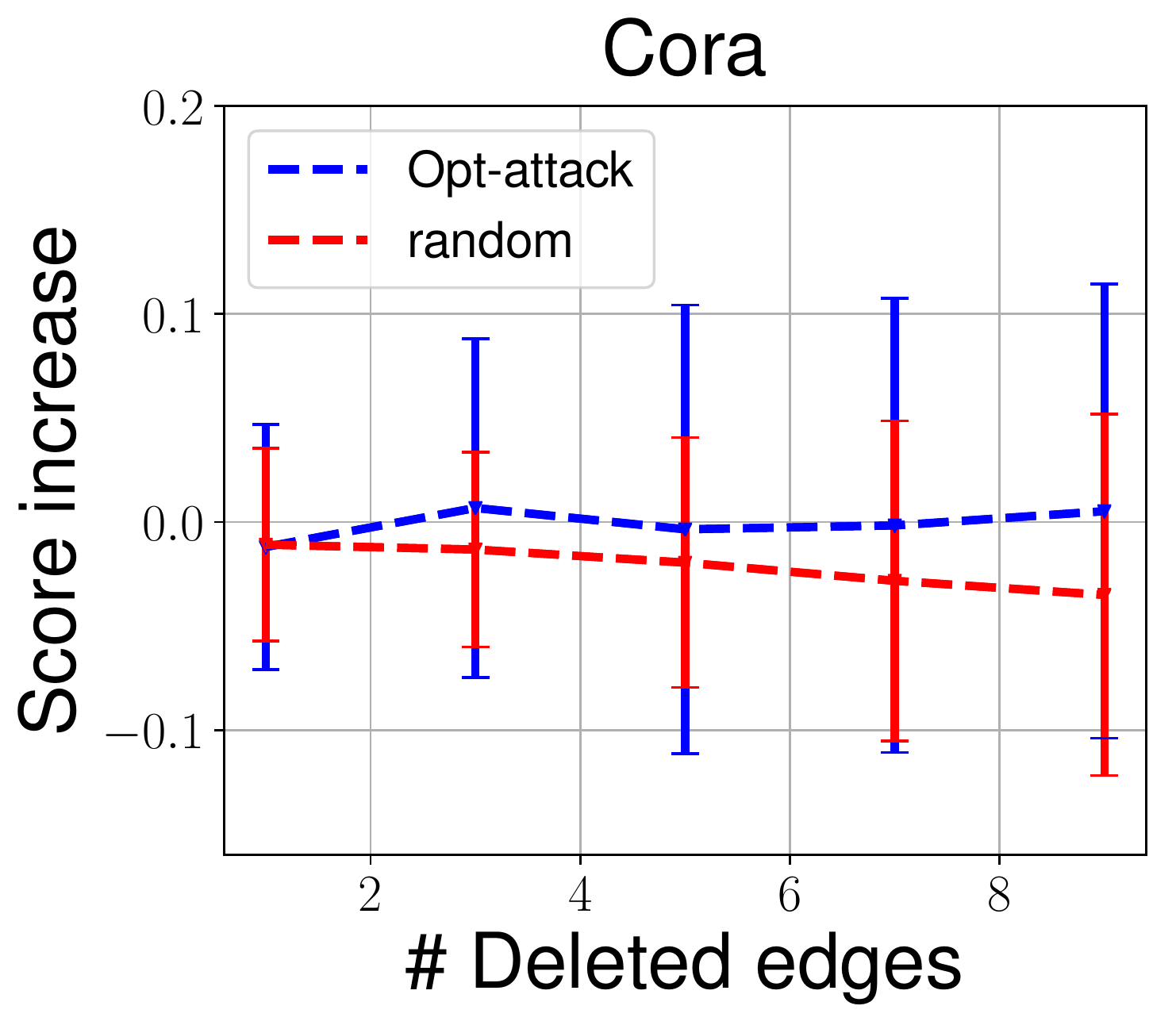}
 \caption{LINE-Del-Up}
 \label{fig:4g}
 \end{subfigure}
\end{minipage}
\begin{minipage}{.245\textwidth}
 \begin{subfigure}{\textwidth}
 \centering
 \includegraphics[width=\textwidth]{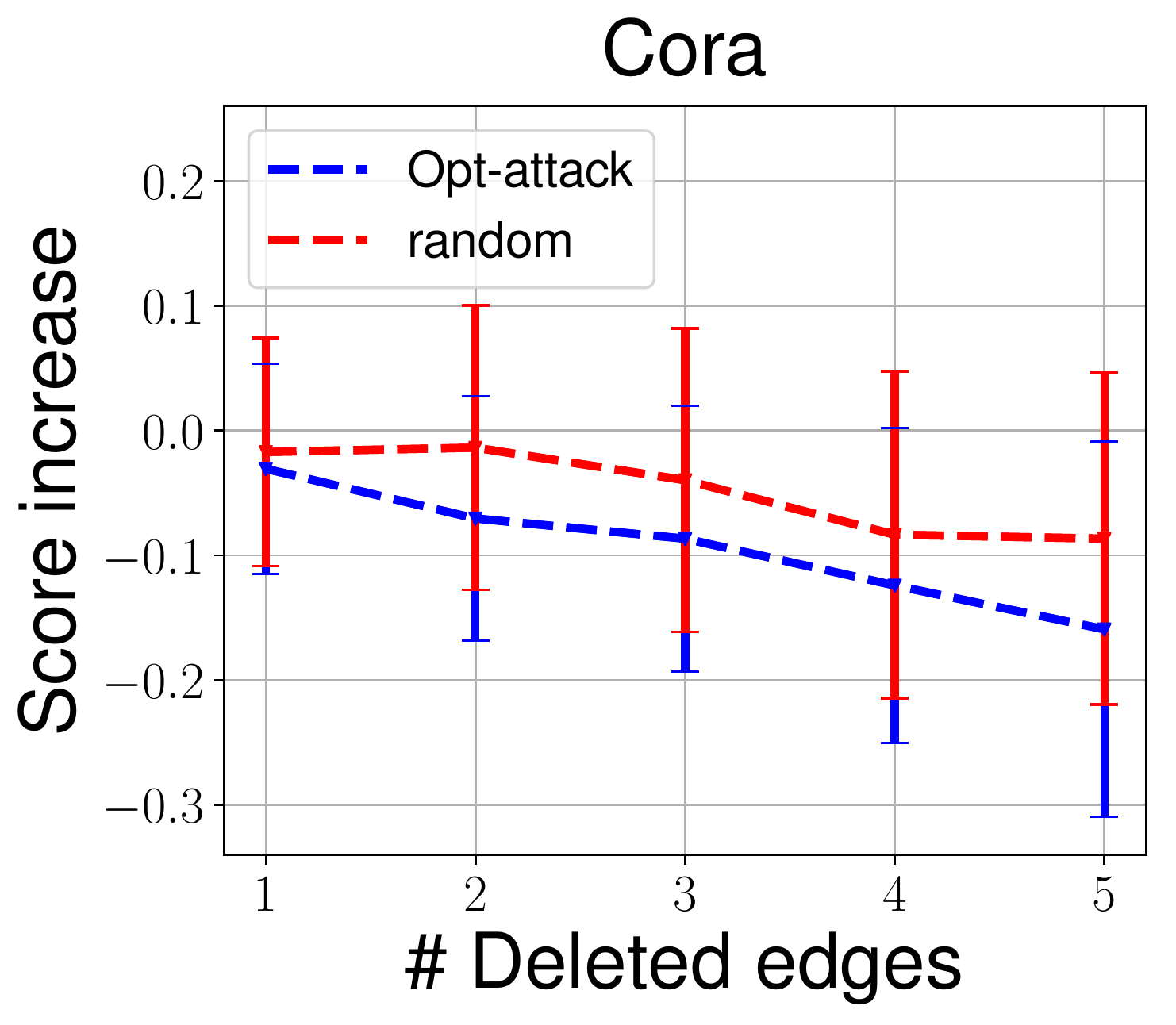}
 \caption{LINE-Del-Down}
 \label{fig:4h}
 \end{subfigure}
\end{minipage}
\caption{Result for direct integrity attack against two methods on two datasets where the action of the attack is deleting edges.}
\label{target-delete}
\vspace{-2ex}
\end{figure*}

\begin{figure*}[!htbp]
\centering
\begin{minipage}{.245\textwidth}
 \begin{subfigure}{\textwidth}
 \centering
 \includegraphics[width=\textwidth]{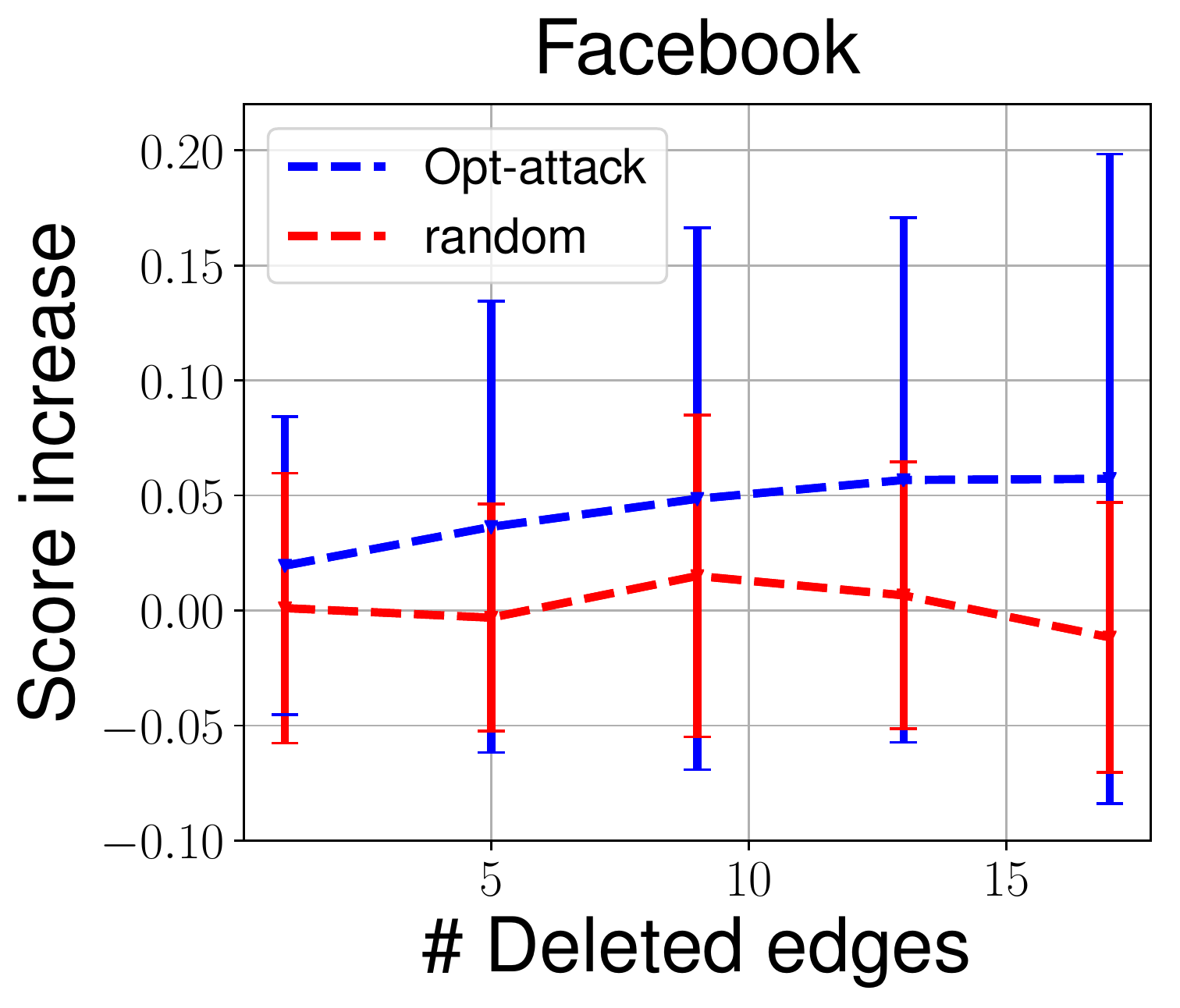}
 \caption{DeepWalk-Del-Up}
 \label{fig:fig3x}
 \end{subfigure}
\end{minipage}
\begin{minipage}{.245\textwidth}
 \begin{subfigure}{\textwidth}
 \centering
 \includegraphics[width=\textwidth]{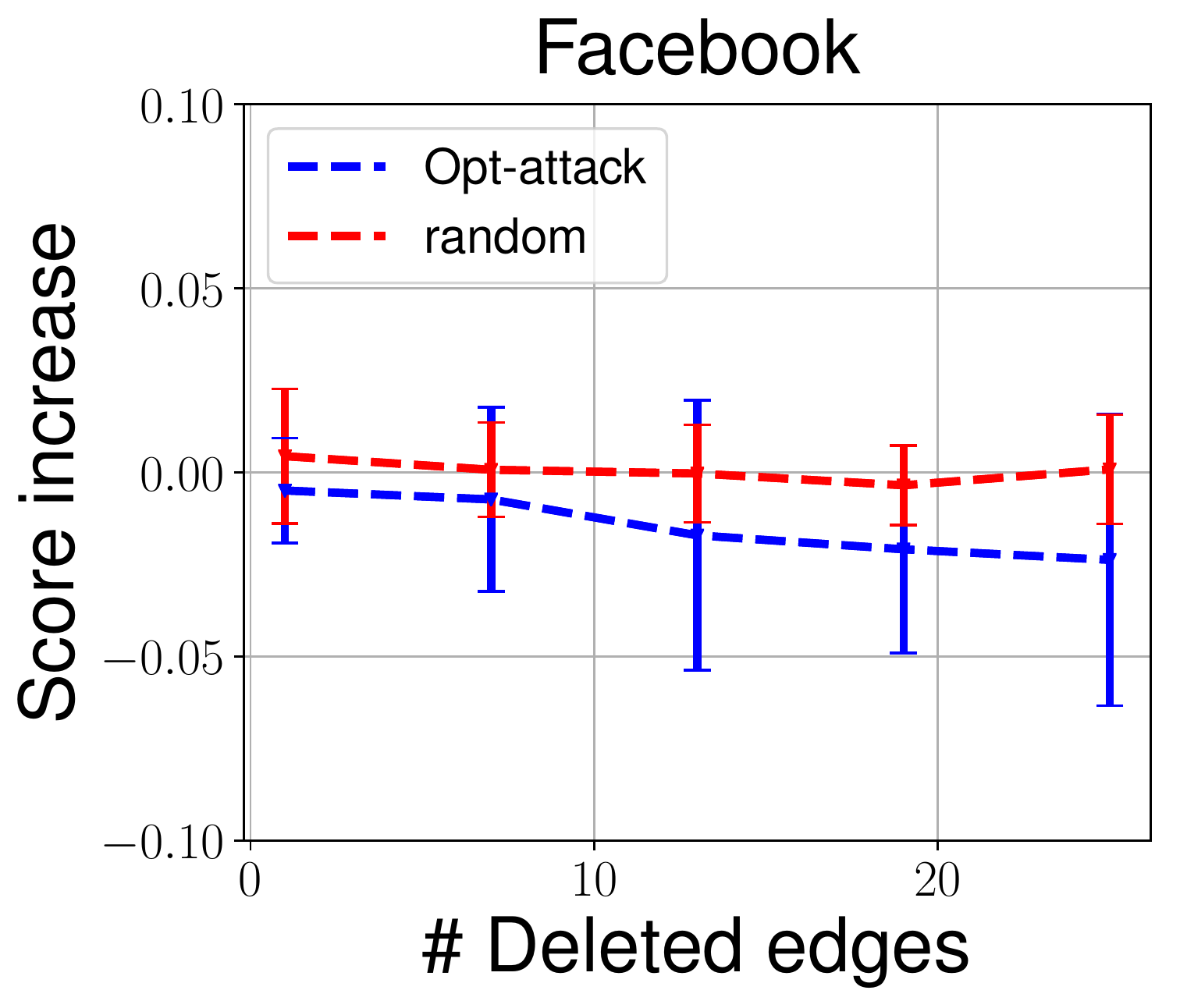}
 \caption{DeepWalk-Del-Down}
 \label{fig:fig3y}
 \end{subfigure}
\end{minipage}
\begin{minipage}{.245\textwidth}
 \begin{subfigure}{\textwidth}
 \centering
 \includegraphics[width=\textwidth]{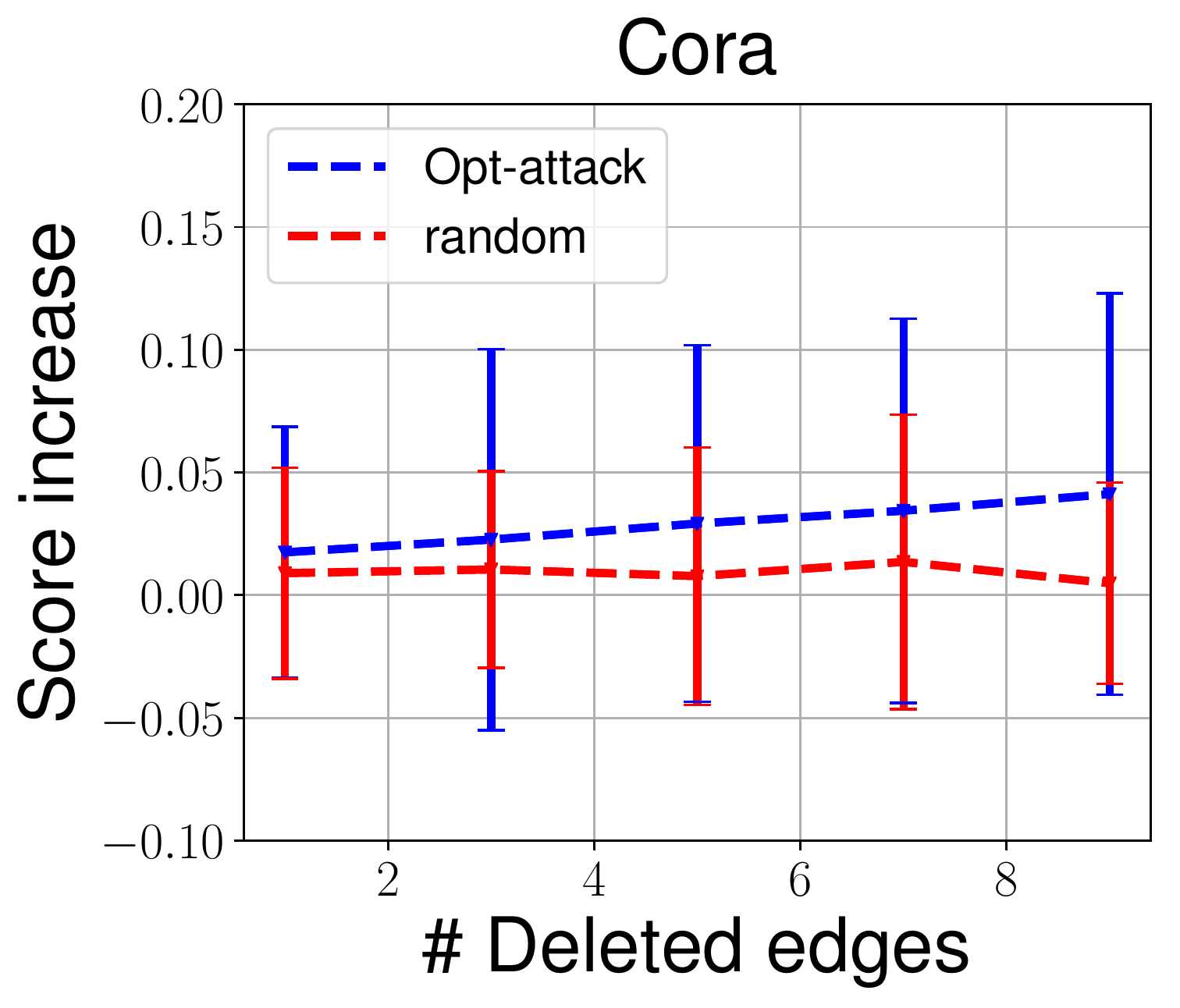}
 \caption{DeepWalk-Del-Up}
 \end{subfigure}
\end{minipage}
\begin{minipage}{.245\textwidth}
 \begin{subfigure}{\textwidth}
 \centering
 \includegraphics[width=\textwidth]{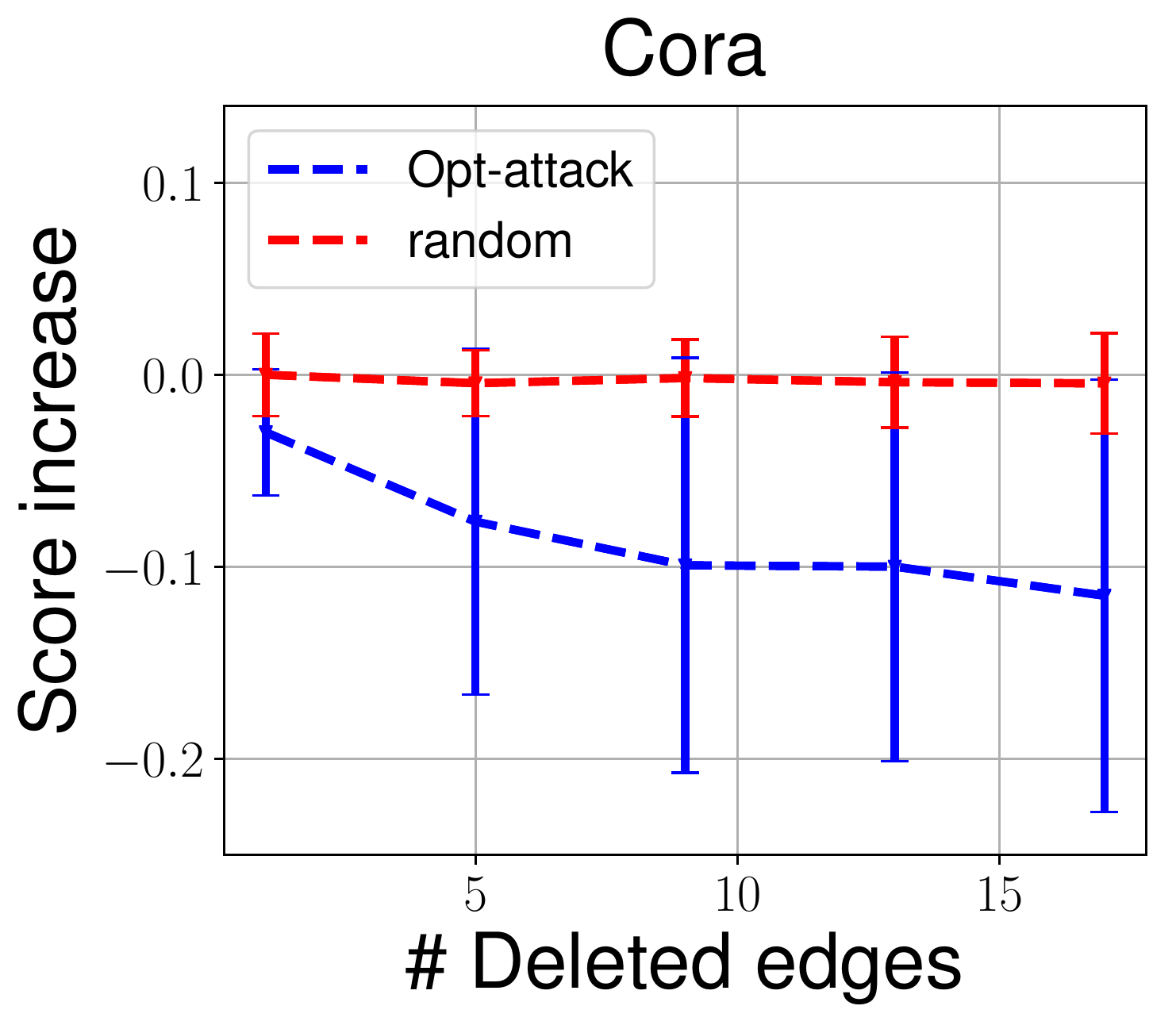}
 \caption{DeepWalk-Del-Down}
 \end{subfigure}
\end{minipage}
\caption{Result for indirect integrity attack against DeepWalk on two datasets where the action of the attacker is deleting edges.}
\label{influence}
\vspace{-3ex}
\end{figure*}

\paragraph{Deleting edges} Now we consider the adversary which can delete existing edges.  Figure~\ref{target-delete} summarizes our result for direct attack. We can see that our attack method works well for attacking DeepWalk (figure ~\ref{fig:4a}~\ref{fig:4b}~\ref{fig:4e}~\ref{fig:4f}). The large variance may be because that different edges have different sensitivity to deleting edges. Also, we notice that LINE is more robust to deleting edges, with average magnitude of score increase(decrease) lower than  DeepWalk.
Figure~\ref{influence} summarizes our results for indirect attack. Still, on average, our attack is able to outperform the random attack baseline. 

\subsection{Availability Attack}\label{avail-attack}
In this part, we show the results for availability attack. We report our results on two dataset: Cora and Citeseer. Results for Citeseer are deferred to the appendix. For both datasets, we choose the test set to be  attack. Table~\ref{avail-cora} summarizes our result. We can see that our attack almost always outperforms the baselines. When adding edges,  our optimization attack outperforms all other baselines by a significant margin. When deleting edges, we can see that LINE is more robust than DeepWalk. In general, we notice the that adding edges is more powerful than deleting edges in our attack.

\renewcommand{\arraystretch}{1.2}
\begin{table}[!htbp]
\resizebox{1.0\textwidth}{!}{
\begin{tabular}{l|ccclllllll}
\hline
\multicolumn{1}{c|}{\multirow{2}{*}{Action}} & \multirow{2}{*}{Model}                        & \multirow{2}{*}{Baseline} & \multirow{2}{*}{Attack Method} & \multicolumn{7}{c}{\# added/deleted edges}                                                                                                                                        \\
\multicolumn{1}{c|}{}                        &                                               &                           &                                & \multicolumn{1}{c}{25} & \multicolumn{1}{c}{50} & \multicolumn{1}{c}{100} & \multicolumn{1}{c}{150} & \multicolumn{1}{c}{200} & \multicolumn{1}{c}{250} & \multicolumn{1}{c}{300} \\ \hline
\multicolumn{1}{c|}{\multirow{6}{*}{Add}}    & \multicolumn{1}{l}{\multirow{3}{*}{DeepWalk}} & \multirow{3}{*}{0.917}    & random                         & 0.922                  & 0.923                  & 0.920                   & 0.919                   & 0.923                   & 0.922                   & 0.922                   \\
\multicolumn{1}{c|}{}                        & \multicolumn{1}{l}{}                          &                           & degree sum                     & 0.915                  & 0.913                  & 0.909                   & 0.904                   & 0.906                   & 0.906                   & 0.904                   \\
\multicolumn{1}{c|}{}                        & \multicolumn{1}{l}{}                          &                           & Opt-attack                           & \textbf{0.832}                  & \textbf{0.773}                  & \textbf{0.666}                   & \textbf{0.597}                   & \textbf{0.559}                   & \textbf{0.532}                   & \textbf{0.503}                   \\ \cline{2-11} 
\multicolumn{1}{c|}{}                        & \multirow{3}{*}{LINE}                         & \multirow{3}{*}{0.909}    & random                         & 0.908                  & 0.908                  & 0.913                   & 0.900                   & 0.901                   & 0.903                   & 0.905                   \\
\multicolumn{1}{c|}{}                        &                                               &                           & degree sum                     & 0.903                  & 0.904                  & 0.899                   & 0.890                   & 0.886                   & 0.886                   & 0.888                   \\
\multicolumn{1}{c|}{}                        &                                               &                           & Opt-attack                           & \textbf{0.898}                  & \textbf{0.892}                  & \textbf{0.886}                   & \textbf{0.846}                   & \textbf{0.826}                   & \textbf{0.803}                   & \textbf{0.766}                   \\ \hline
\multirow{8}{*}{Delete}                      & \multirow{4}{*}{DeepWalk}                     & \multirow{4}{*}{0.917}    & random                         & 0.916                  & 0.917                  & 0.921                   & 0.918                   & 0.920                   & 0.914                   & 0.913                   \\
                                             &                                               &                           & degree sum       & 0.917                  & 0.918                  & 0.919                   & 0.920                   & 0.922                   & 0.918                   & 0.916                   \\
                                             &                                               &                           & shortest path                  & 0.916                  & 0.919                  & 0.917                   & 0.918                   & 0.922                   & 0.921                   & 0.924                   \\
                                             &                                               &                           & Opt-attack                           & \textbf{0.897}                  & \textbf{0.892}                  & \textbf{0.876}                   & \textbf{0.866}                   & \textbf{0.853}                   & \textbf{0.838}                   & \textbf{0.835}                   \\ \cline{2-11} 
                                             & \multirow{4}{*}{LINE}                         & \multirow{4}{*}{0.909}    & random                         & \textbf{0.901}                  & \textbf{0.899}                  & \textbf{0.892}                   & 0.901                   & 0.898                   & 0.894                   & 0.886                   \\
                                             &                                               &                           & degree sum               & 0.903                  & 0.904                  & 0.908                   & 0.919                   & 0.911                   & 0.890                   & 0.888                   \\
                                             &                                               &                           & shortest path                  & 0.903                  & 0.904                  & 0.908                   & 0.919                   & 0.911                   & 0.890                   & 0.889                   \\
                                             &                                               &                           & Opt-attack                           & 0.915                  & 0.909                  & 0.898                   & \textbf{0.890}                   & \textbf{0.876}                   & \textbf{0.859}                   & \textbf{0.861}                  \\ \hline
\end{tabular}
}
\caption{Results for availability attack on Cora dataset. Here we report the AP score.
}
\label{avail-cora}
\end{table}

\subsection{Transferability}
In this part, we show that our attack can be transferred across different embedding methods. Besides DeepWalk and LINE, we choose another three embedding methods to test the transferability of our approach: 1. Variational Graph Autoencoder(GAE)~\citep{gae}; 2. Spectral Clustering~\citep{SC}; 3. Node2Vec~\citep{grover2015node2vec}. For GAE, we use the default setting as in the original paper. For Node2Vec, we first tune the parameters $p$, $q$ on a validation set and use the best $p$, $q$ for Node2Vec. 

\begin{figure*}[!htbp]
\centering
\begin{minipage}{.49\textwidth}
 \begin{subfigure}{\textwidth}
 \centering
 \captionsetup{singlelinecheck=off, margin={3.0cm, 0cm}, format=hang}
 \includegraphics[width=\textwidth]{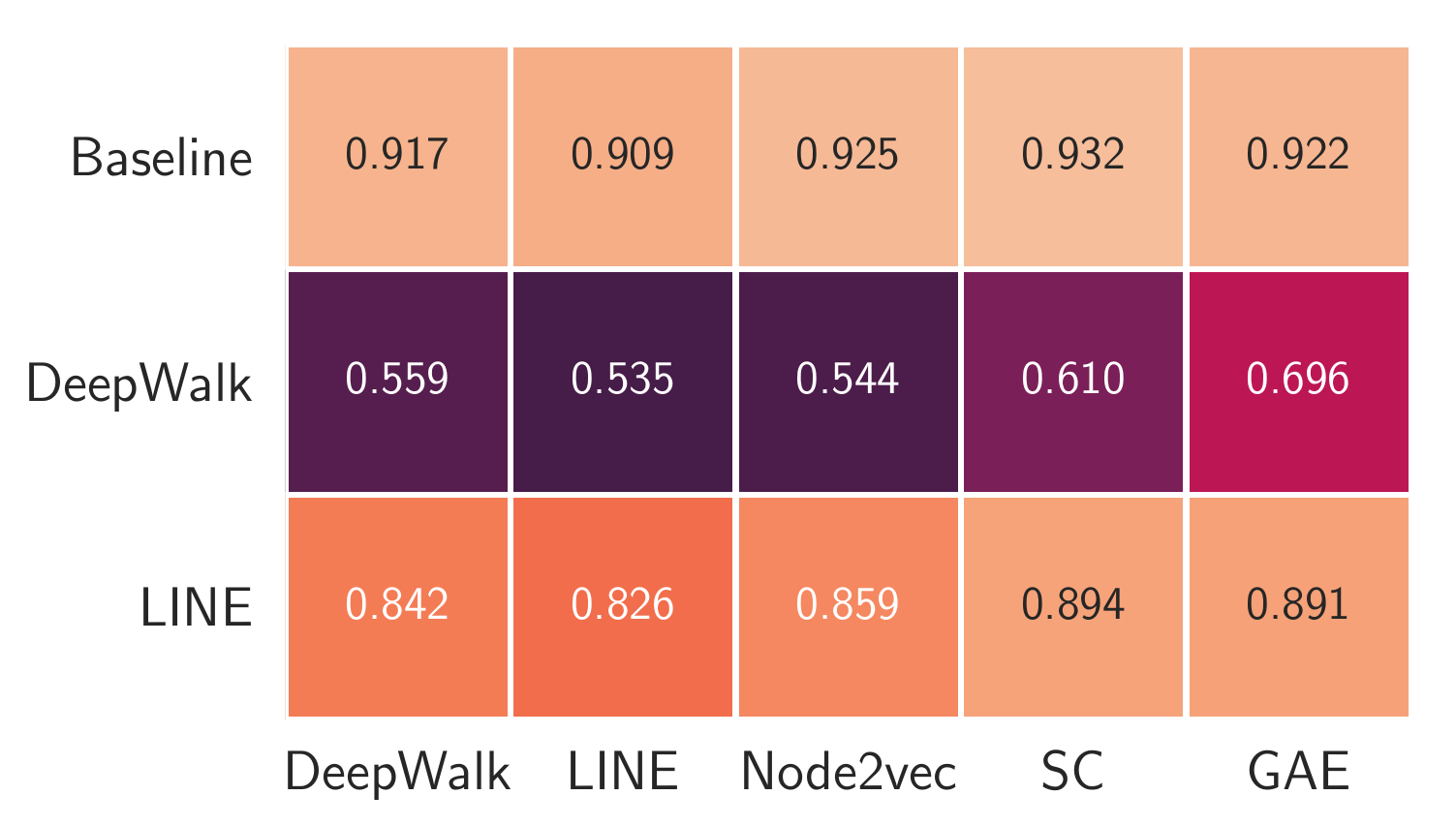}
 \caption{Cora-Add}
 \label{fig7:1}
 \end{subfigure}
\end{minipage}
\begin{minipage}{.49\textwidth}
 \begin{subfigure}{\textwidth}
 \centering
  \captionsetup{singlelinecheck=off, margin={3.0cm, 0cm}, format=hang}
 \includegraphics[width=\textwidth]{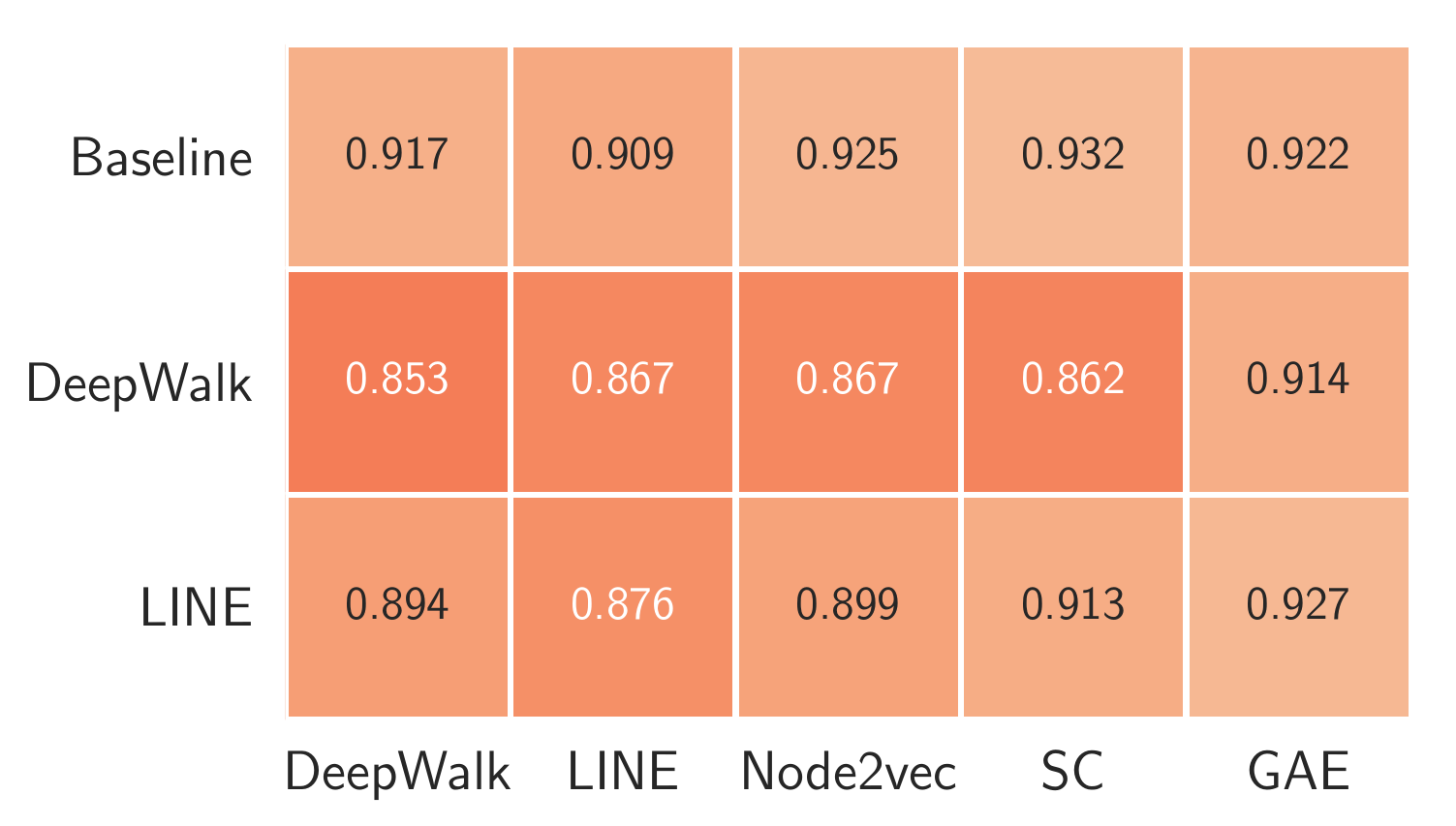}
 \caption{Cora-Del}
 \label{fig7:2}
 \end{subfigure}
\end{minipage}
\begin{minipage}{.49\textwidth}
 \begin{subfigure}{\textwidth}
 \centering
  \captionsetup{singlelinecheck=off, margin={2.8cm, 0cm}, format=hang}
 \includegraphics[width=\textwidth]{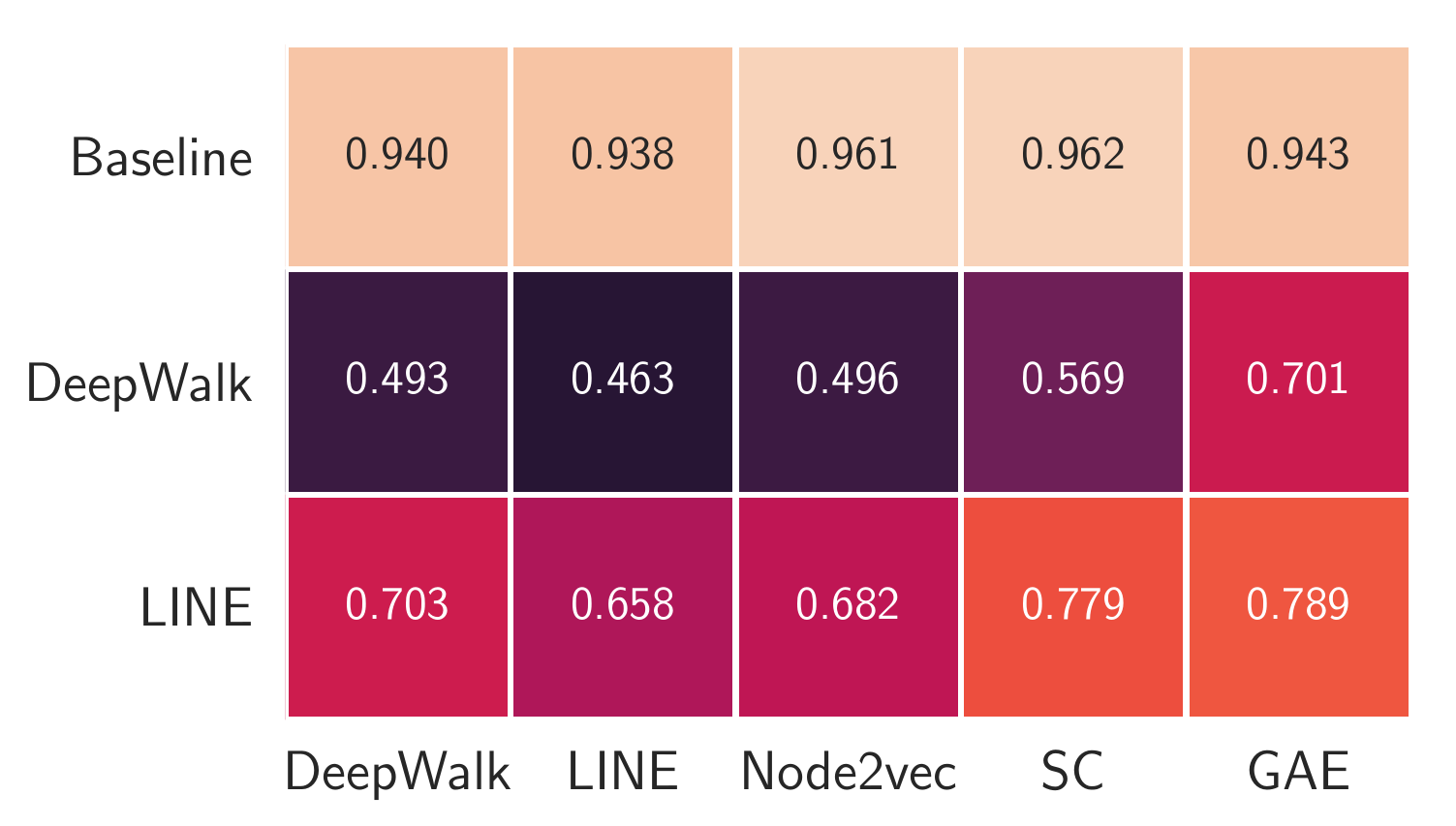}
 \caption{Citeseer-Add}
 \label{fig7:3}
 \end{subfigure}
\end{minipage}
\begin{minipage}{.49\textwidth}
 \begin{subfigure}{\textwidth}
 \centering
  \captionsetup{singlelinecheck=off, margin={2.8cm, 0cm}, format=hang}
 \includegraphics[width=\textwidth]{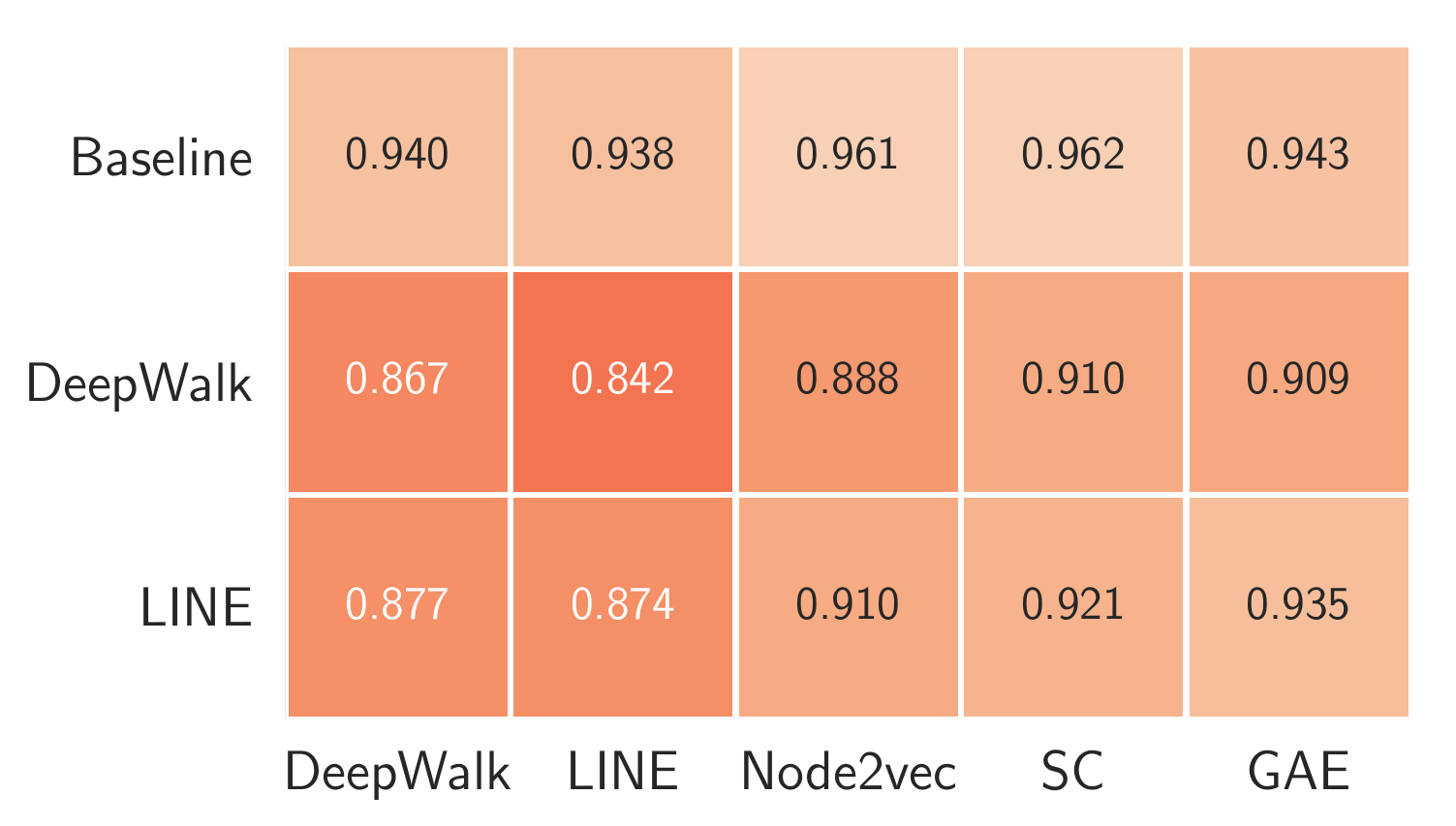}
 \caption{Citeseer-Add}
 \label{fig7:4}
 \end{subfigure}
\end{minipage}
\caption{Result for transferability analysis of our attack on two datasets, where the number of added/deleted edges is 200. X-axis indicates the method the attack is evaluated on. Y-axis includes the methods to generate the attack and also the baseline. The format ``Dataset | Type '' here is used to label the each sub-caption. ``Dataset'' refers to dataset while ``Type'' refers to attacker's action. }
\label{trans}
\end{figure*}

Figure~\ref{trans} shows our result for transferability test of our attack, where the number of added(deleted) edges is 200. Results when the number of added(deleted) edges is 100 and 300 are deferred to the appendix. Comparing Figure~\ref{fig7:1}~\ref{fig7:3} and Figure~\ref{fig7:2}~\ref{fig7:4}, we can see that adding edges is more effective than deleting edges in our attack. 
The attack on DeepWalk has higher transferability compared with other four methods (including LINE$_{2nd}$).  Comparing the decrease of AP score for all five methods, we can see that GAE is more robust against transferability based attacks.

\vspace{-1ex}
 \section{Case Study: Attack Deepwalk on Coauthor Network}

In this section, we conduct a case study on a real-world 
coauthor network extracted from DBLP~\citep{tang2008arnetminer}.
We construct a coauthor network from two different research communities: machine learning \& data mining (ML\&DM), and security (Security). For each research community, we select some conferences in each field: ICML, ICLR, NIPS, KDD, WWW, ICWD, ICDM from ML\&DM and IEEE S\&P, CCS, Usenix and NDSS from Security. We sort the authors according to the number of published papers and keep the top-500 authors that publish most papers in each community, which eventually yields a coauthor network with 1,000 nodes in total. The constructed coauthor graph contains 27260 in-field edges and 1014 cross-field edges. We analyze both integrity attack and availability attack on this coauthor network.

\begin{figure*}[!htbp]
\centering
\begin{minipage}{.32\textwidth}
 \begin{subfigure}{\textwidth}
 \centering
 \includegraphics[width=\textwidth]{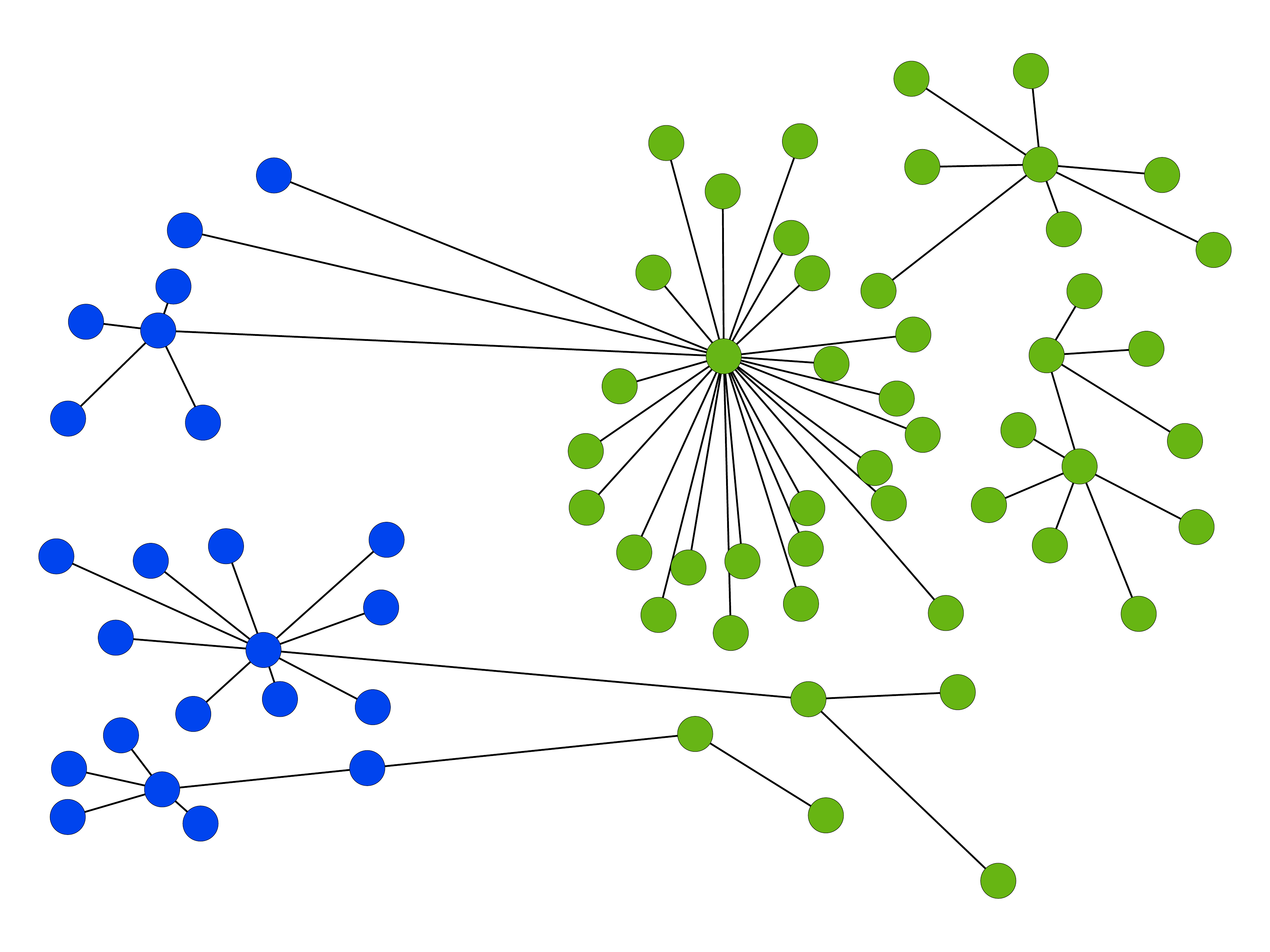}
 \caption{Original Network}
 \label{fig:casestudy1}
 \end{subfigure}
\end{minipage}
\begin{minipage}{.32\textwidth}
 \begin{subfigure}{\textwidth}
 \centering
 \includegraphics[width=\textwidth]{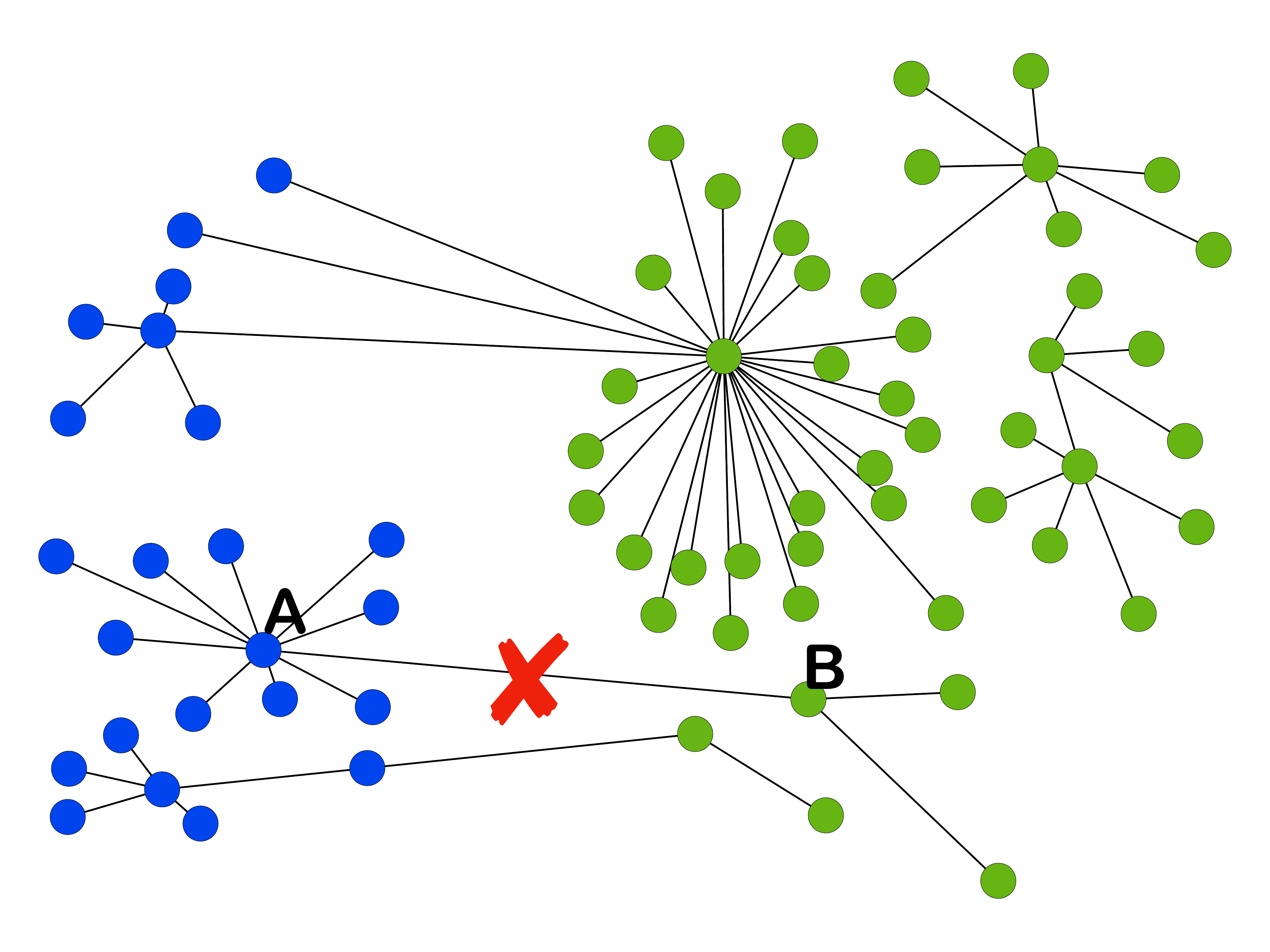}
 \caption{Adversarial Goal}
 \label{fig:casestudy2}
 \end{subfigure}
\end{minipage}
\begin{minipage}{.32\textwidth}
 \begin{subfigure}{\textwidth}
 \centering
 \includegraphics[width=\textwidth]{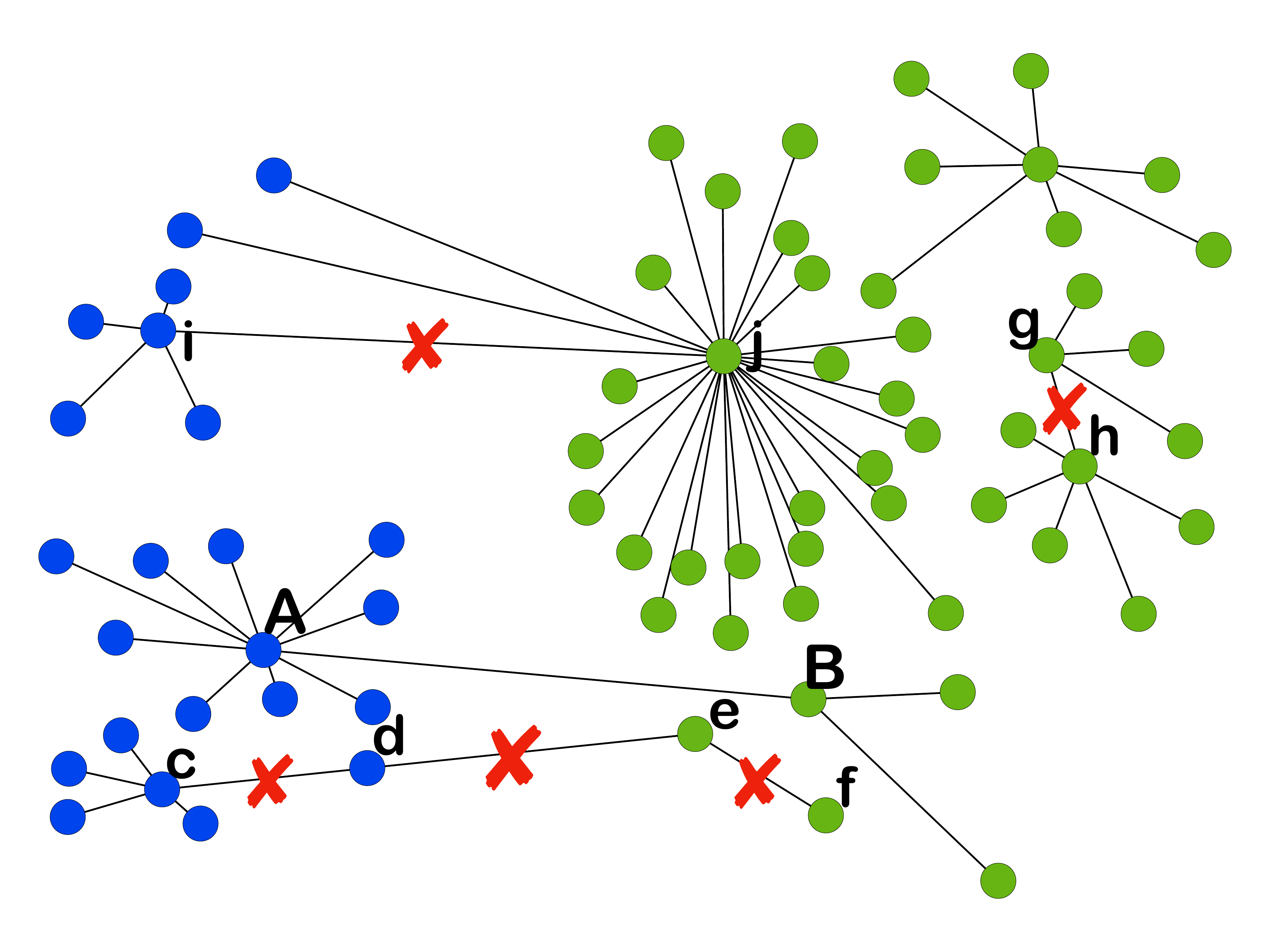}
 \caption{Attack Example: Victims}
 \label{fig:casestudy3}
 \end{subfigure}
\end{minipage}
\caption{Visualization for integrity attack. Green nodes denote authors from ML\&DM community and blue nodes denote authors from Security community. Nodes and their corresponding researchers: A: John C. Mitchell; B: Trevor Hastie; c: Susan Landau; d: Michael Lesk; e: Hector Garcia-Molina; f: Jure Leskovec; g: David D. Jensen; h: Thomas G. Dietterich; i: Matthew Fredrikson; j: Jiawei Han.}
\label{fig:casestudy}
\vspace{-2ex}
\end{figure*}

\textbf{Integrity Attack} We consider an indirect attack setting where the adversarial goal is to decrease the score of a target node pair and the attacker's action is deleting existing edges. 
We show the sub-graph that contains the nodes in the target edge and 5 edges chosen by our algorithm, as well as the nodes that coauthored more than 3 papers with them, e.g. frequent collaborators. We visualize the original graph in Figure~\ref{fig:casestudy1}. 
Green nodes represent ML\&DM authors and blue nodes denote authors from Security community. 

In Figure~\ref{fig:casestudy2}, we show the target node pair A and B. Node A denotes John C. Mitchell, a professor in Stanford University from the security community and node B denotes Trevor Hastie, also a Stanford professor, from the ML \& DM community. After the attack, the similarity score of the target node pair is reduced from 0.67 to 0.37. We make the following observations: 1) We find that the top 2 edges (d, e) and (e,f) chosen by our attack  lie on the shortest path between A and B, which corresponds to the intuitive understanding that cutting the paths connecting A with B makes it less likely to predict that an edge exists between A and B; 2) We find that many of the edges chosen by our algorithm are  cross-field edges (figure~\ref{fig:casestudy3}): edge (i, j) and edge (d, e). Considering how small a proportion the cross-field edges consist of($3.6\%$ of all edges), we hypothesize that cutting the cross-field edges could impede the information flow between two communities and therefore making the similarity score of cross-field link lower.

\textbf{Availability Attack} For availability attack, we analyze the adversary that add edges. It turns out that our algorithm tends to add cross-field edges: for the top 40 added edges chosen by our attack method, 39 are cross-field edges. We hypothesize that this is because adding more edges between two communities can disrupt the existing information flow and therefore lead the learnt embedding to carry incorrect information about the network structure.
\section{Conclusion}
In this paper, we investigate data poisoning attack against unsupervised node embedding methods and take the task of link prediction as an example. We study two types of data poisoning attacks including integrity attack and availability attack. We propose a unified optimization framework to optimally attack the node embedding methods for both types of attacks. Experimental results on several real-world graphs show that our proposed approach can effectively attack the results of link prediction by adding or removing a few edges. Results also show that the adversarial examples discovered by our proposed approach are transferable across different node embedding methods. Finally, we conduct a case study analysis to better understand our attack method. In the future, we plan to study how to design effective defense strategies for node embedding methods.

\bibliographystyle{iclr2019_conference}
\bibliography{iclr2019_conference}
\clearpage
\begin{appendices}
 \section{Implementation Detail}
In this part, we discuss the initialization of weighted adjacency matrix in the projected gradient descent step. From the formulation in section~\ref{pgd}, if we initialize all cells which are initially 0 to 0. Then there won't be back-propagated gradient on these cells. (This is because $\Omega$ won't contain these cells.) To handle this issue, we initialize these cells with a small value, for example 0.001, which allows the gradient on these cells to be efficiently computed.

\section{Additional Experimental Results}
\subsection{Integrity attack}
\begin{figure*}[!htbp]
\centering
\begin{minipage}{.24\textwidth}
 \begin{subfigure}{\textwidth}
 \centering
 \includegraphics[width=\textwidth]{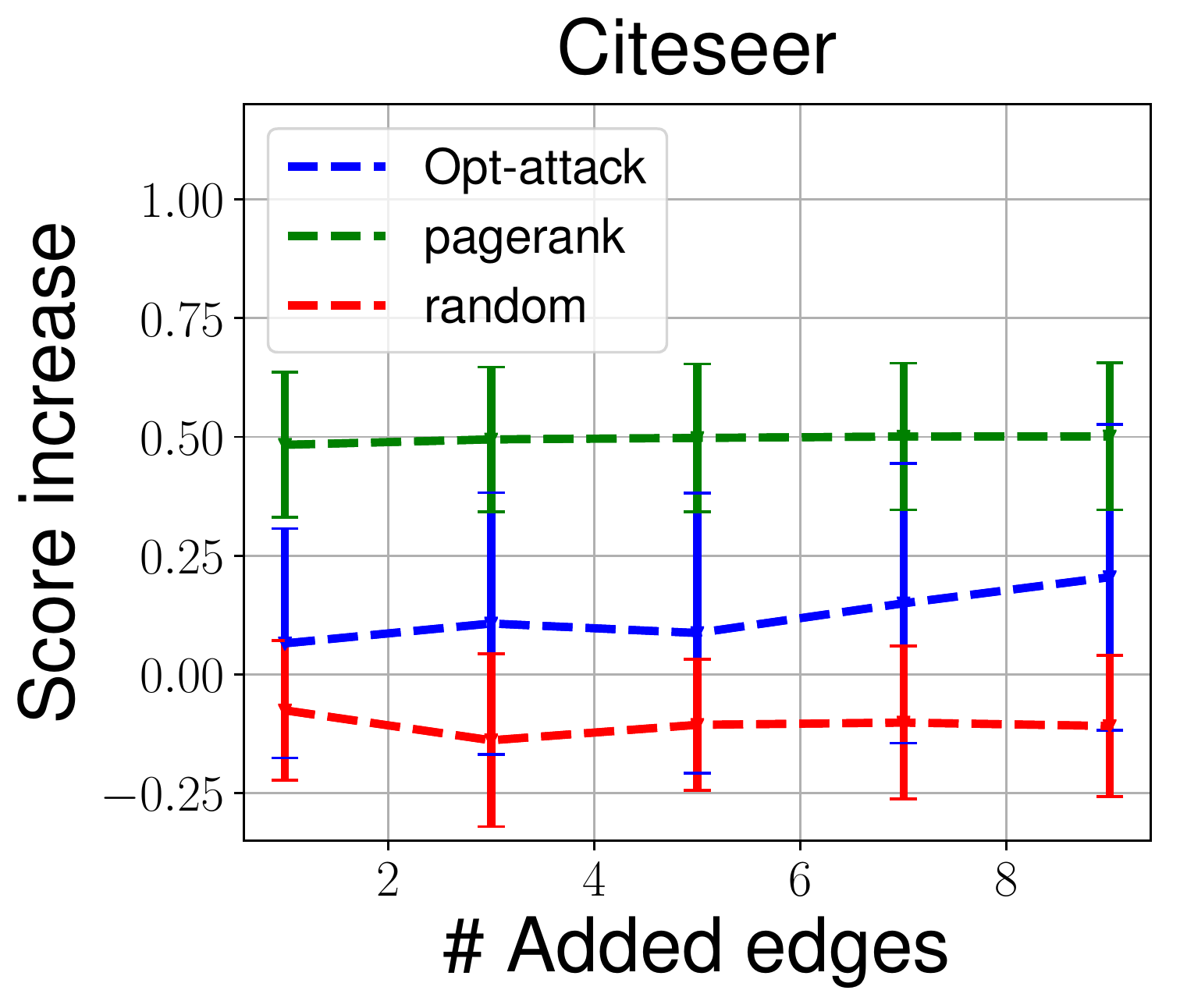}
 \caption{DeepWalk-Add-Up}
  \label{citeseer-1}
 \end{subfigure}
\end{minipage}
\begin{minipage}{.24\textwidth}
 \begin{subfigure}{\textwidth}
 \centering
 \includegraphics[width=\textwidth]{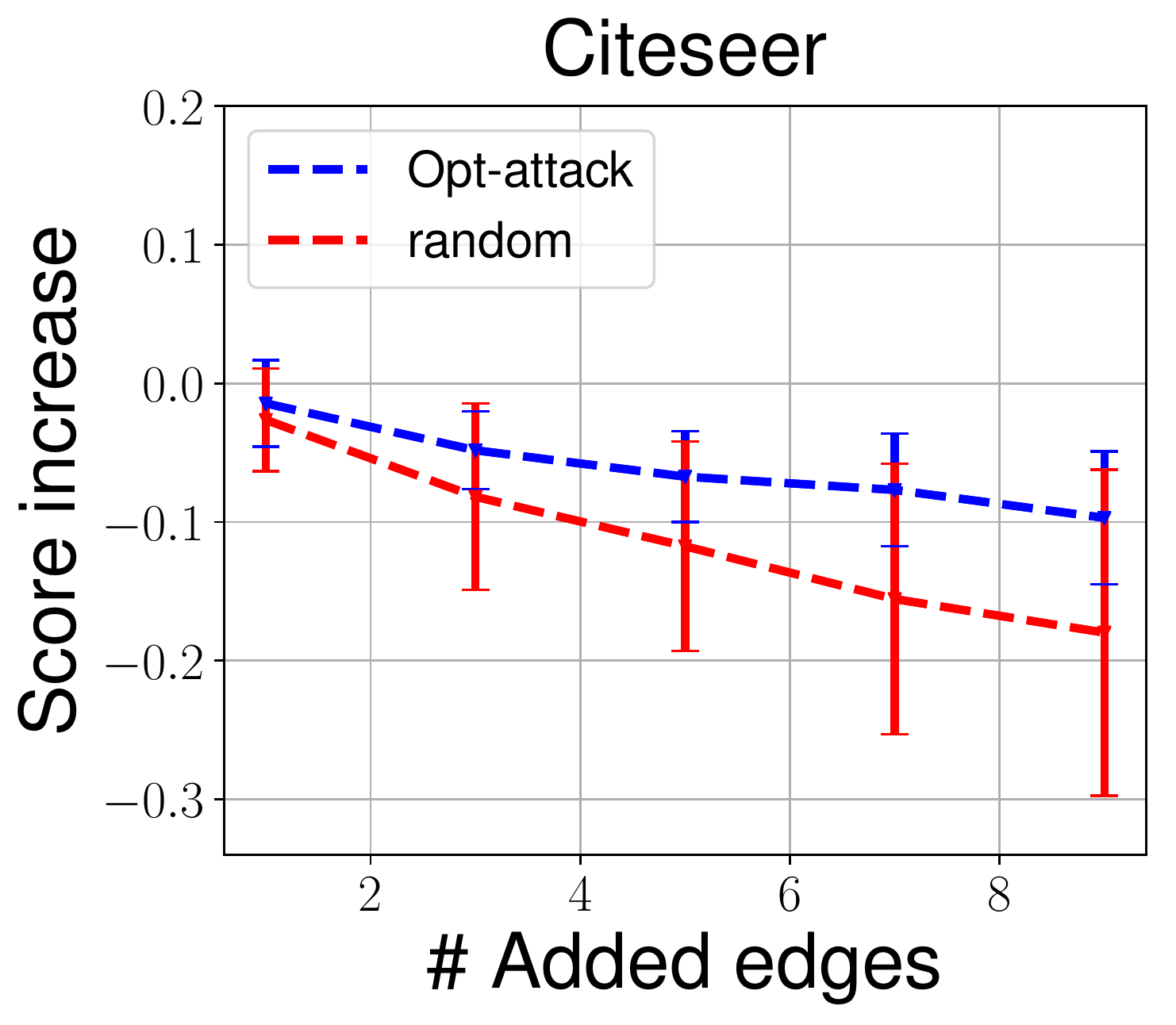}
 \caption{DeepWalk-Add-Down}
  \label{citeseer-2}
 \end{subfigure}
\end{minipage}
\begin{minipage}{.24\textwidth}
 \begin{subfigure}{\textwidth}
 \centering
 \includegraphics[width=\textwidth]{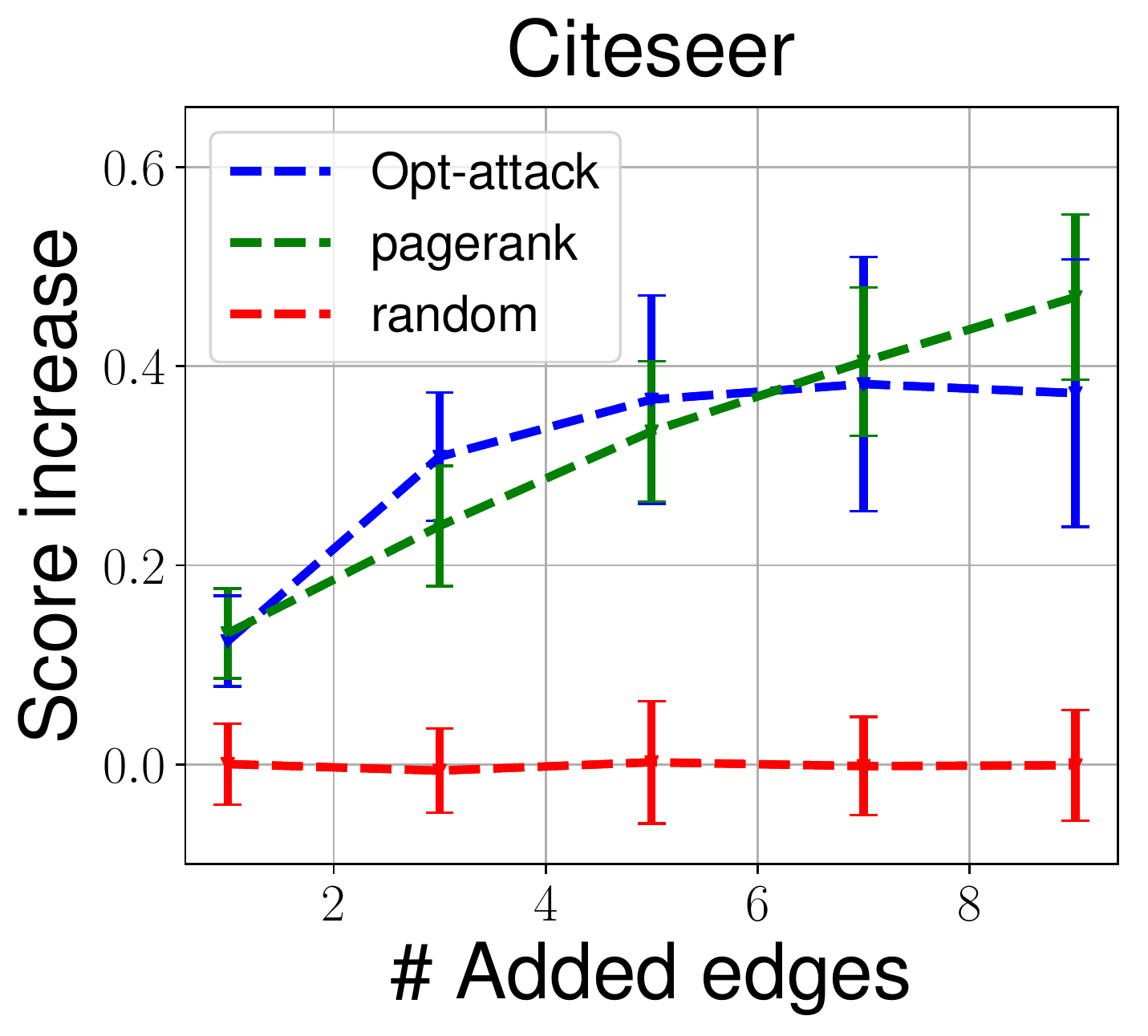}
 \caption{LINE-Add-Up}
 \label{citeseer-3}
 \end{subfigure}
\end{minipage}
\begin{minipage}{.24\textwidth}
 \begin{subfigure}{\textwidth}
 \centering
 \includegraphics[width=\textwidth]{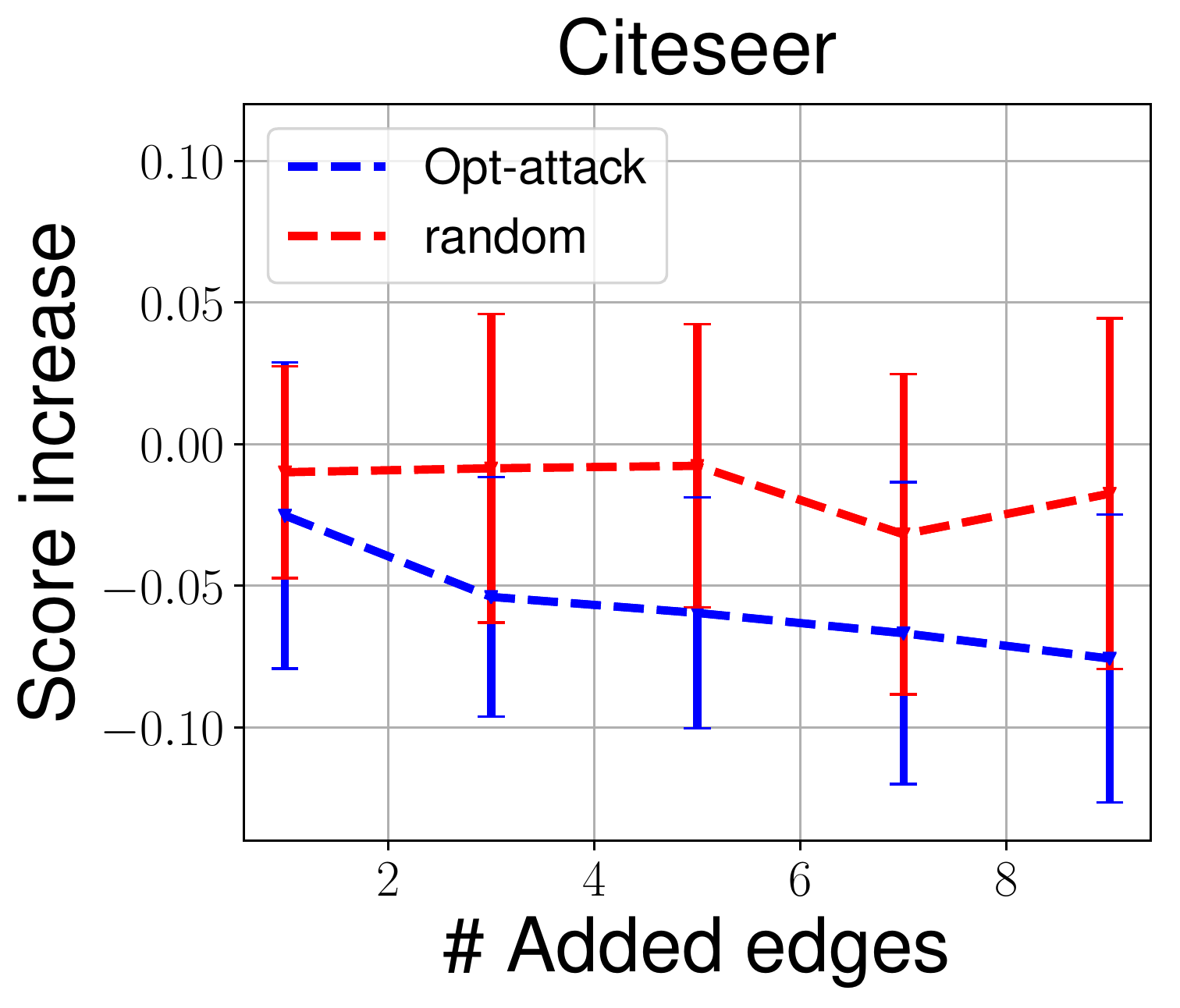}
 \caption{LINE-Add-Down}
  \label{citeseer-4}
 \end{subfigure}
\end{minipage}
\begin{minipage}{.24\textwidth}
 \begin{subfigure}{\textwidth}
 \centering
 \includegraphics[width=\textwidth]{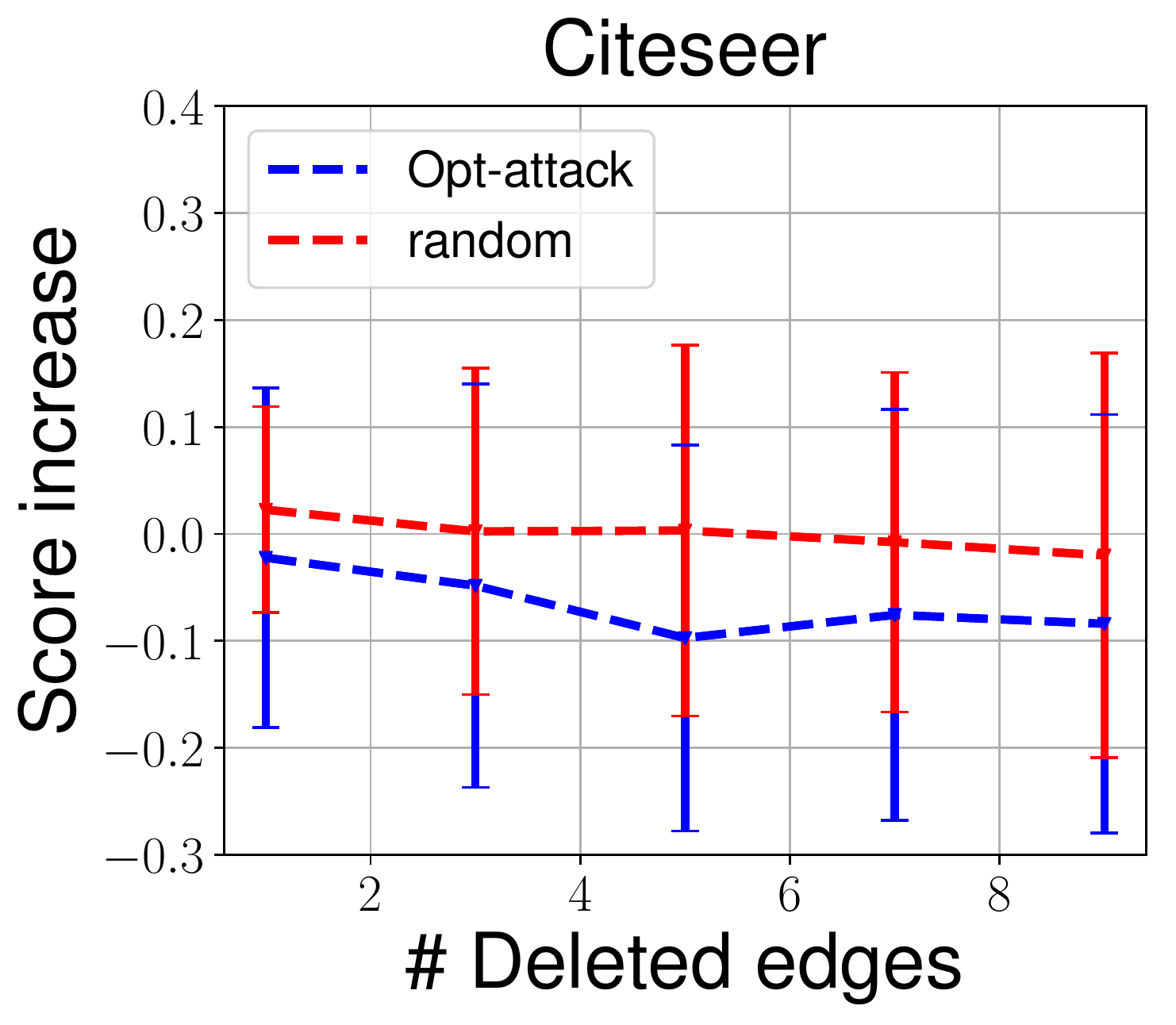}
 \caption{DeepWalk-Del-Up}
  \label{citeseer-5}
 \end{subfigure}
\end{minipage}
\begin{minipage}{.24\textwidth}
 \begin{subfigure}{\textwidth}
 \centering
 \includegraphics[width=\textwidth]{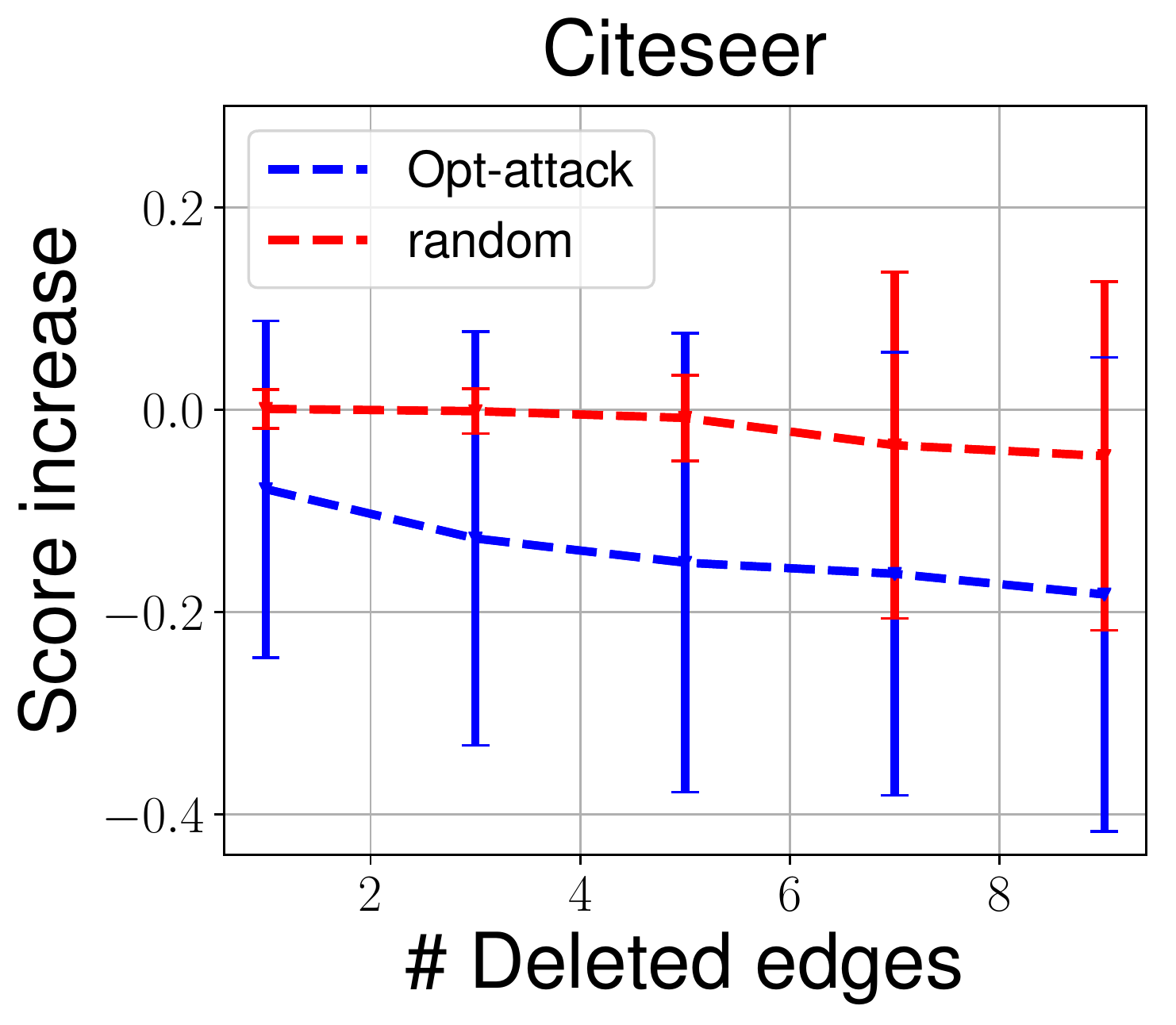}
 \caption{DeepWalk-Del-Down}
  \label{citeseer-6}
 \end{subfigure}
\end{minipage}
\begin{minipage}{.24\textwidth}
 \begin{subfigure}{\textwidth}
 \centering
 \includegraphics[width=\textwidth]{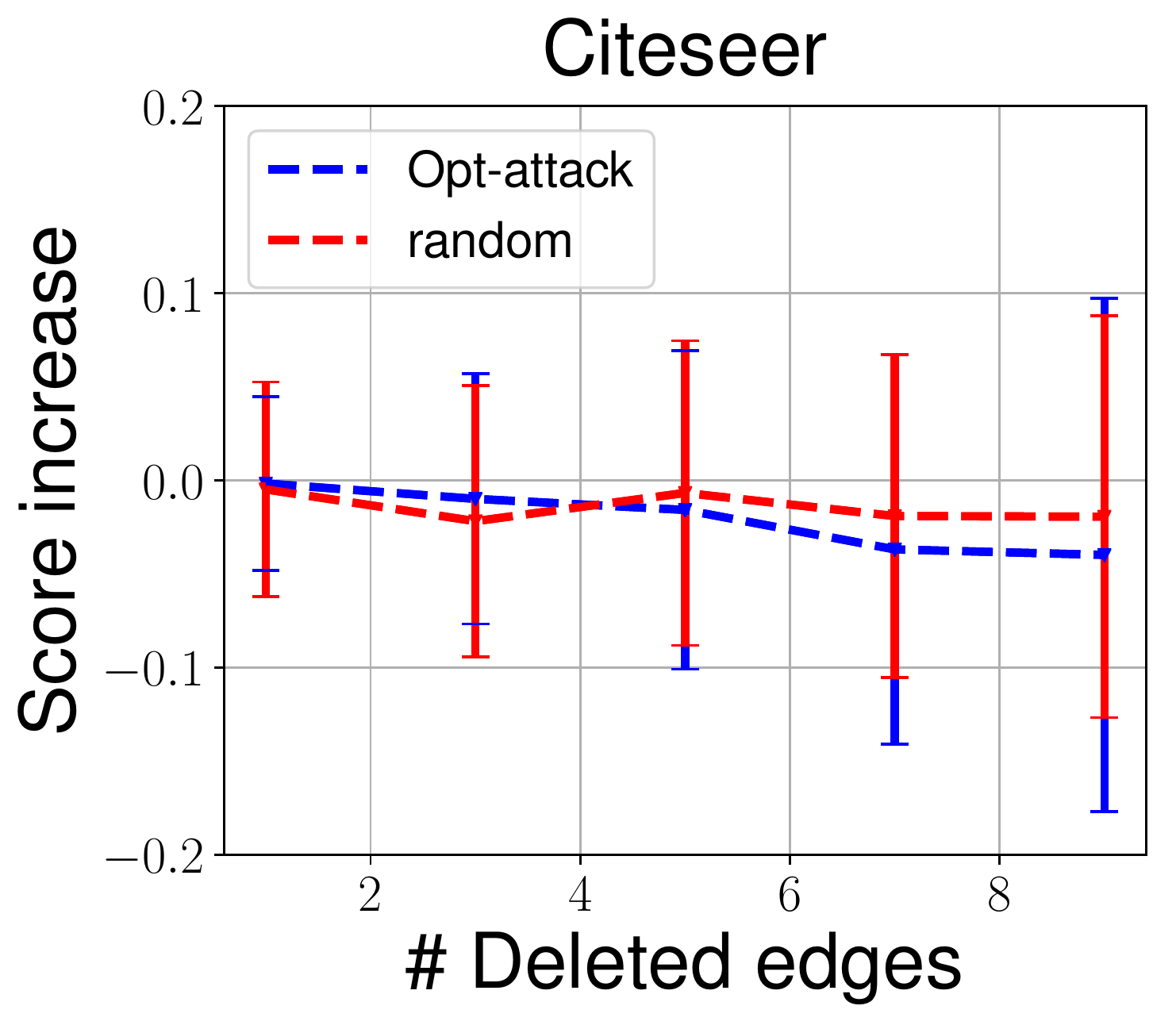}
 \caption{LINE-Del-Up}
  \label{citeseer-7}
 \end{subfigure}
\end{minipage}
\begin{minipage}{.24\textwidth}
 \begin{subfigure}{\textwidth}
 \centering
 \includegraphics[width=\textwidth]{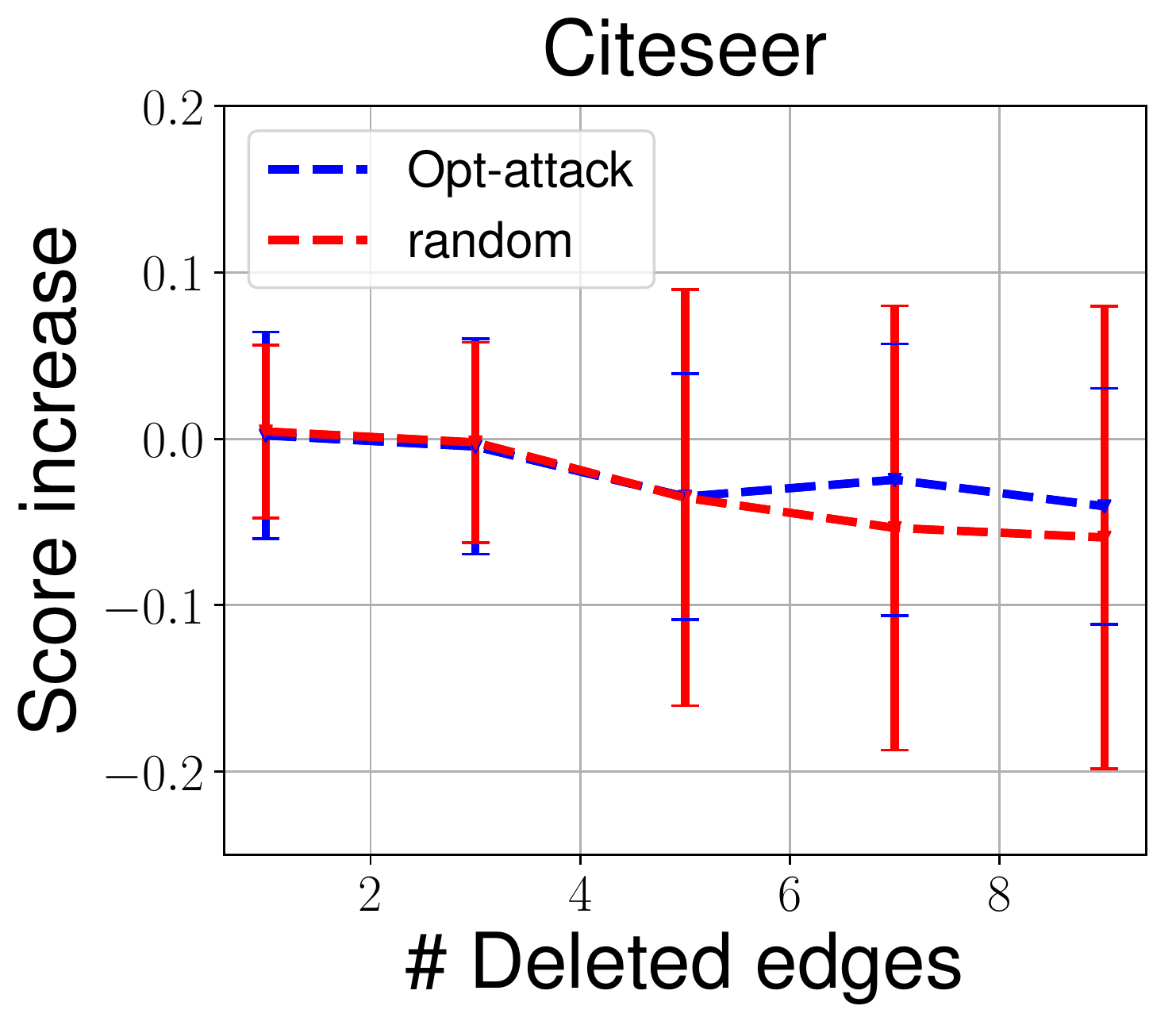}
 \caption{LINE-Del-Down}
  \label{citeseer-8}
 \end{subfigure}
\end{minipage}
\caption{Result for direct integrity attack against DeepWalk and LINE on Citeseer dataset.}
\label{citeseer}
\end{figure*}

\begin{figure*}[!htbp]
\centering
\begin{minipage}{.24\textwidth}
 \begin{subfigure}{\textwidth}
 \centering
 \includegraphics[width=\textwidth]{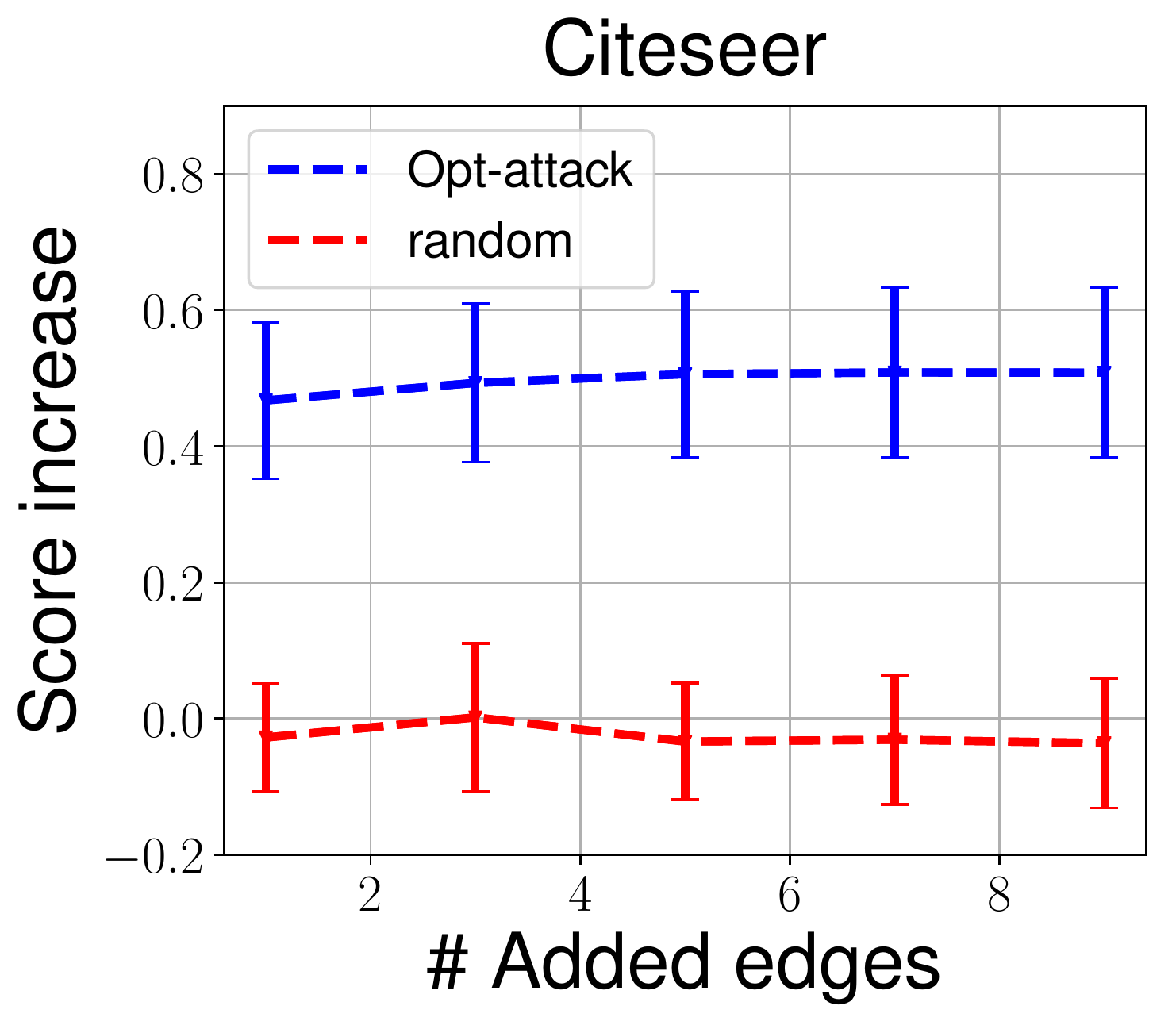}
 \caption{DeepWalk-Add-Up}
 \end{subfigure}
\end{minipage}
\begin{minipage}{.24\textwidth}
 \begin{subfigure}{\textwidth}
 \centering
 \includegraphics[width=\textwidth]{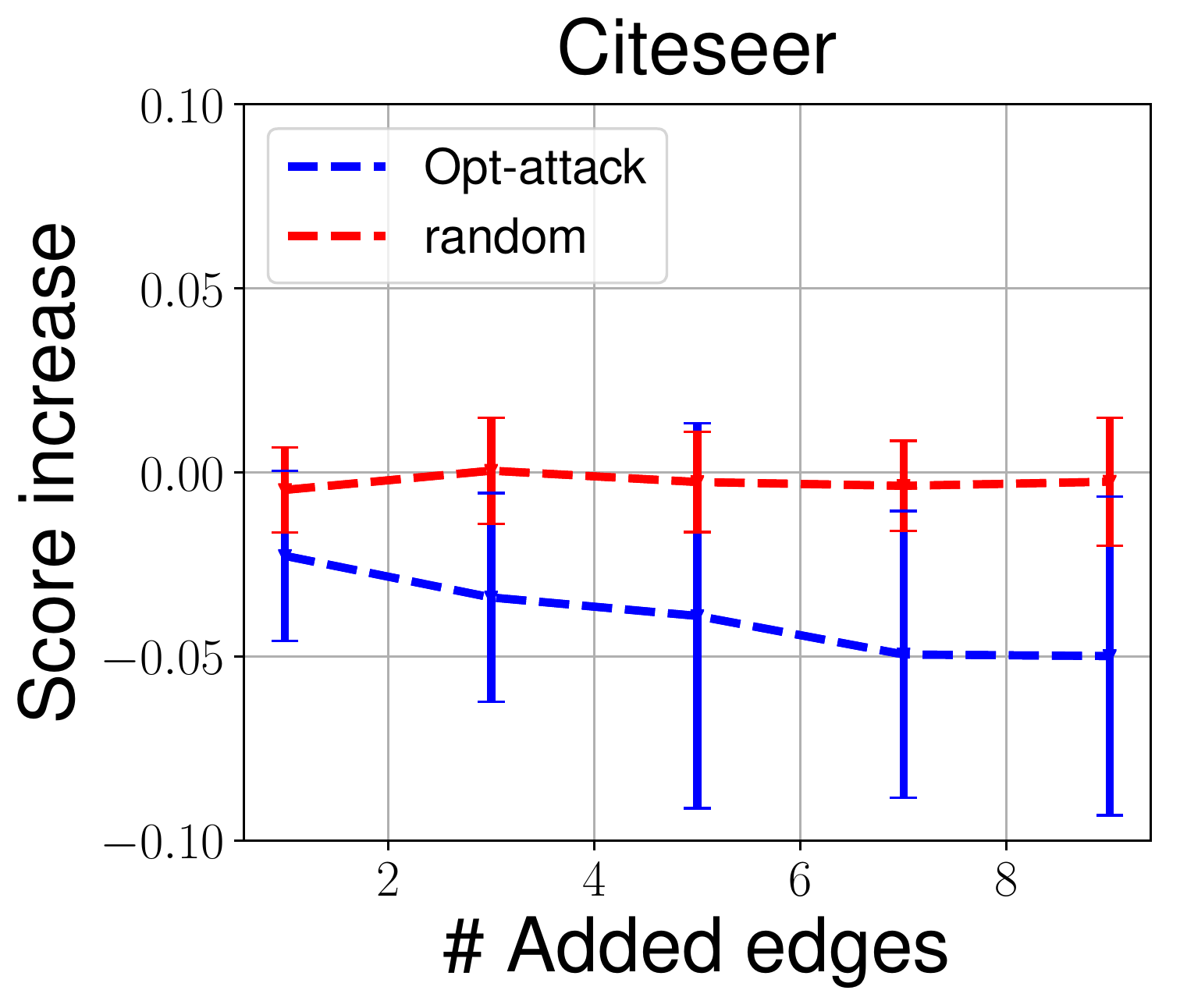}
 \caption{DeepWalk-Add-Down}
 \end{subfigure}
\end{minipage}
\begin{minipage}{.24\textwidth}
 \begin{subfigure}{\textwidth}
 \centering
 \includegraphics[width=\textwidth]{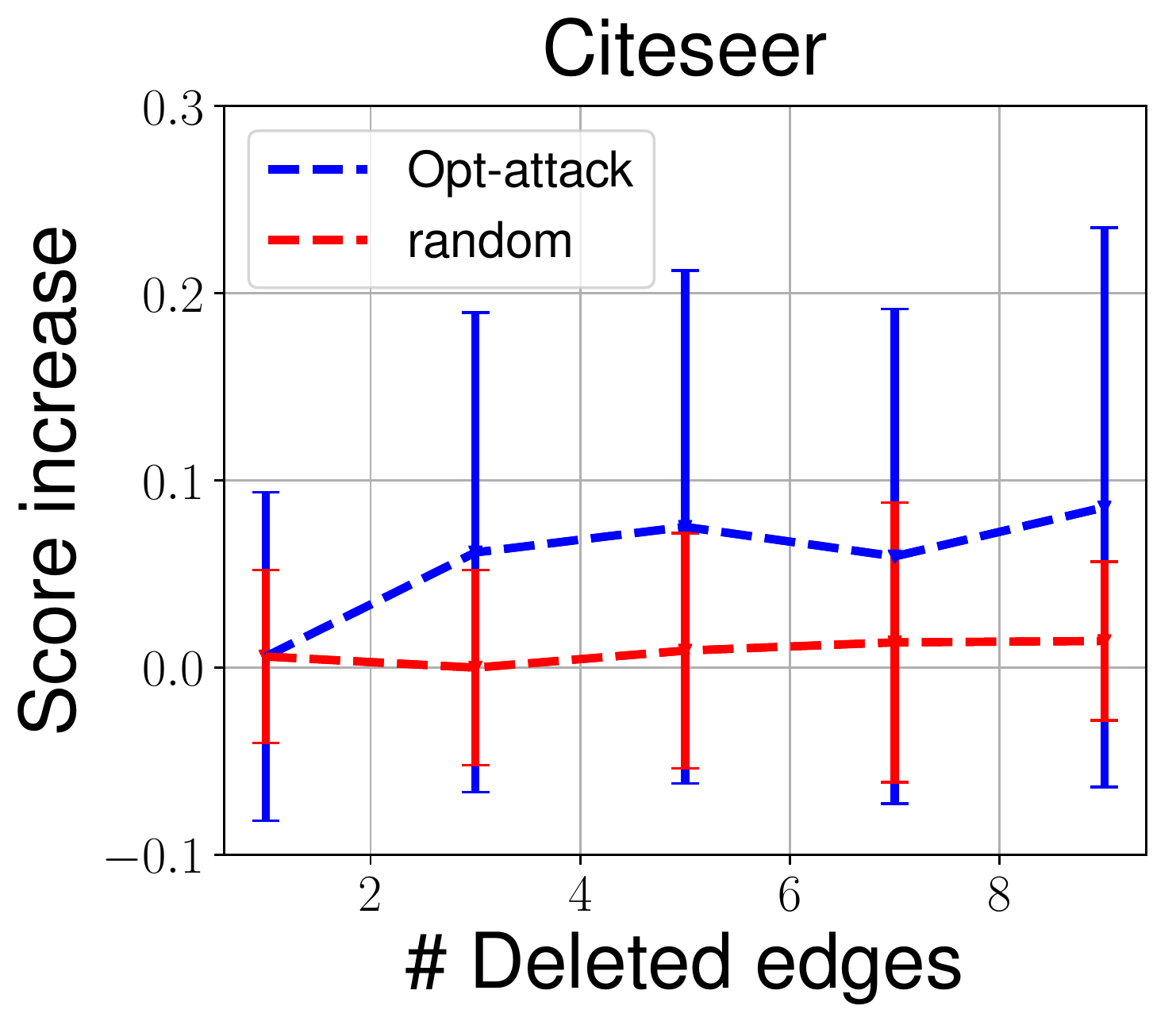}
 \caption{DeepWalk-Del-Up}
 \label{away-3-citeseer}
 \end{subfigure}
\end{minipage}
\begin{minipage}{.24\textwidth}
 \begin{subfigure}{\textwidth}
 \centering
 \includegraphics[width=\textwidth]{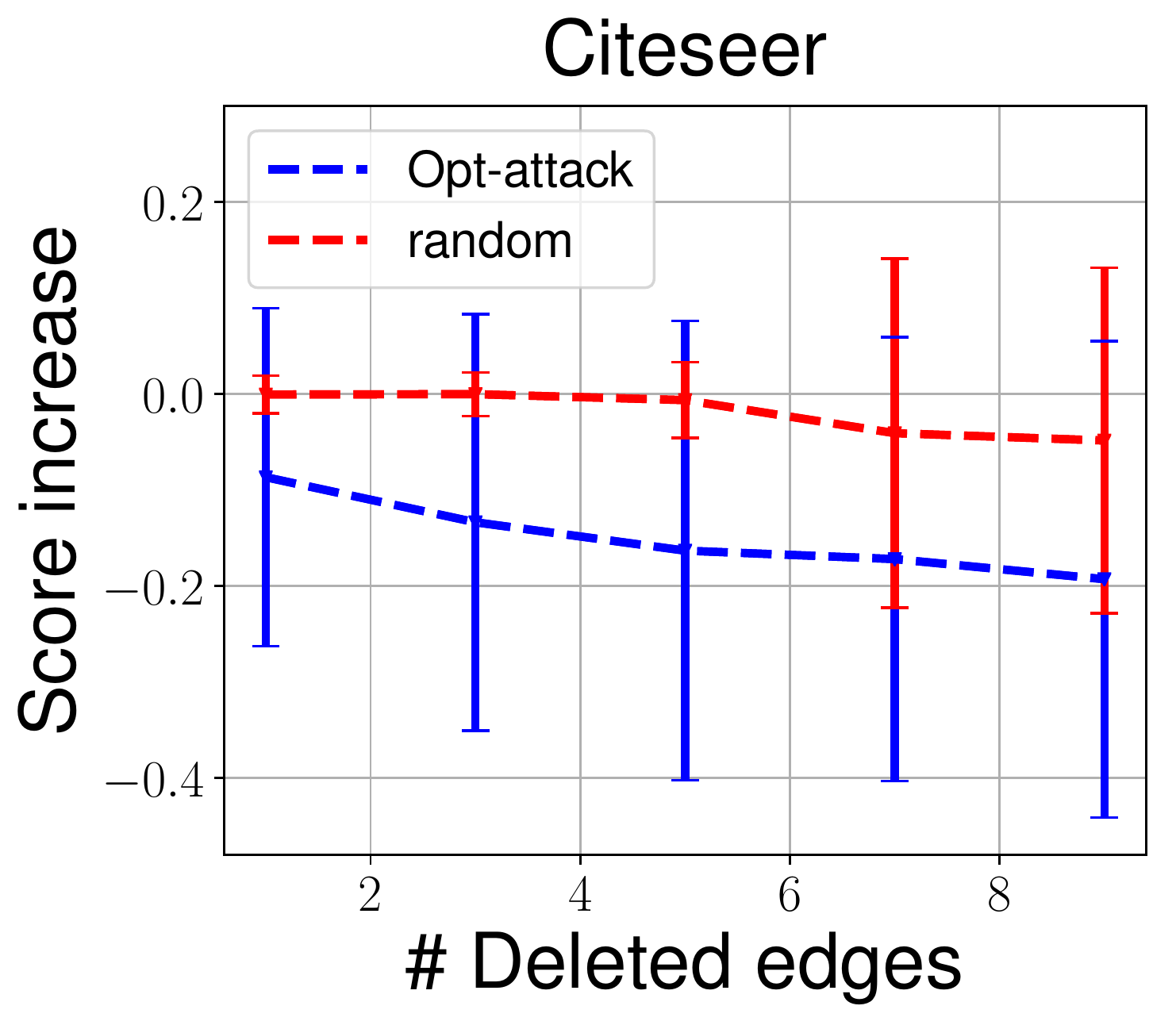}
 \caption{DeepWalk-Del-Down}
 \label{away-4-citeseer}
 \end{subfigure}
\end{minipage}
\caption{Result for indirect integrity attack against DeepWalk on Citeseer dataset.}
\label{citeseer-away}
\end{figure*}

Here we show additional experimental results. Figure~\ref{citeseer},~\ref{citeseer-away} summarize the results of integrity attack on Citeseer dataset. Figure~\ref{citeseer} shows our results for direct integrity attack and Figure~\ref{citeseer-away} shows our results for indirect integrity attack. 

\section{Availability attack}
\begin{figure*}[!htbp]
\centering
\begin{minipage}{.49\textwidth}
 \begin{subfigure}{\textwidth}
 \centering
 \captionsetup{singlelinecheck=off, margin={3.0cm, 0cm}, format=hang}
 \includegraphics[width=\textwidth]{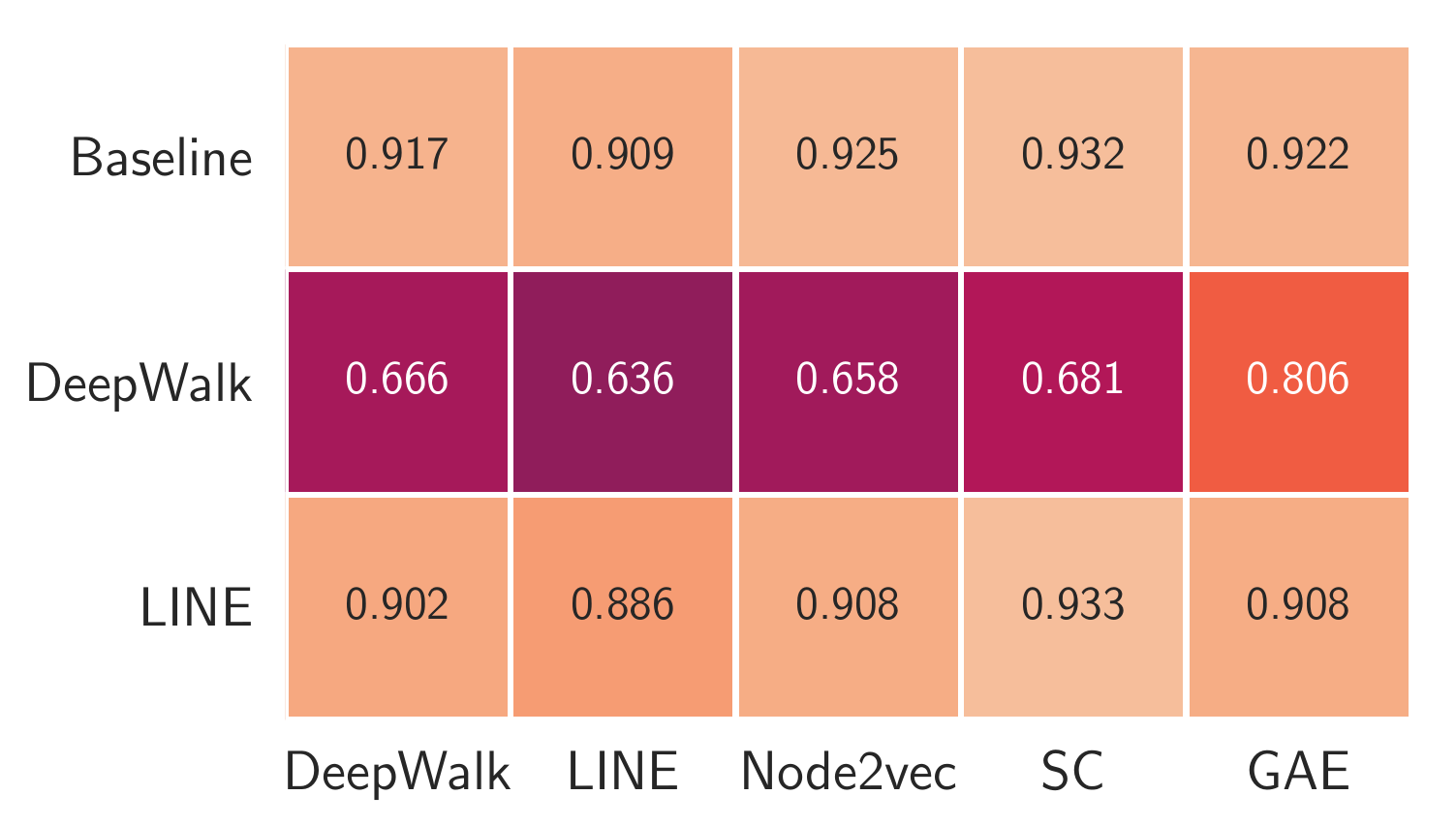}
 \caption{Cora-Add}
 \label{fig9:1}
 \end{subfigure}
\end{minipage}
\begin{minipage}{.49\textwidth}
 \begin{subfigure}{\textwidth}
 \centering
 \captionsetup{singlelinecheck=off, margin={3.0cm, 0cm}, format=hang}
 \includegraphics[width=\textwidth]{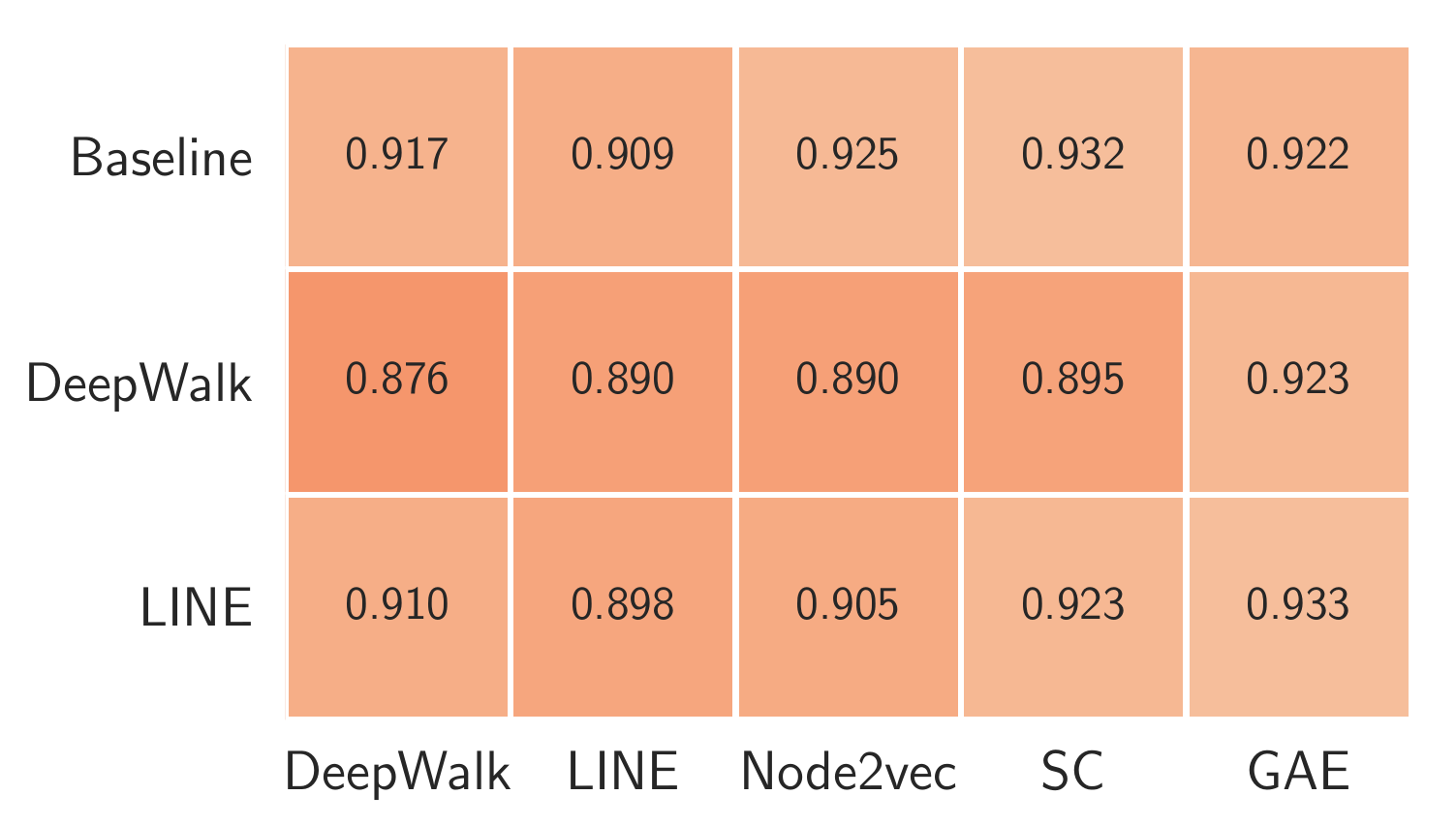}
 \caption{Cora-Del}
 \label{fig9:2}
 \end{subfigure}
\end{minipage}
\begin{minipage}{.49\textwidth}
 \begin{subfigure}{\textwidth}
 \centering
 \captionsetup{singlelinecheck=off, margin={2.8cm, 0cm}, format=hang}
 \includegraphics[width=\textwidth]{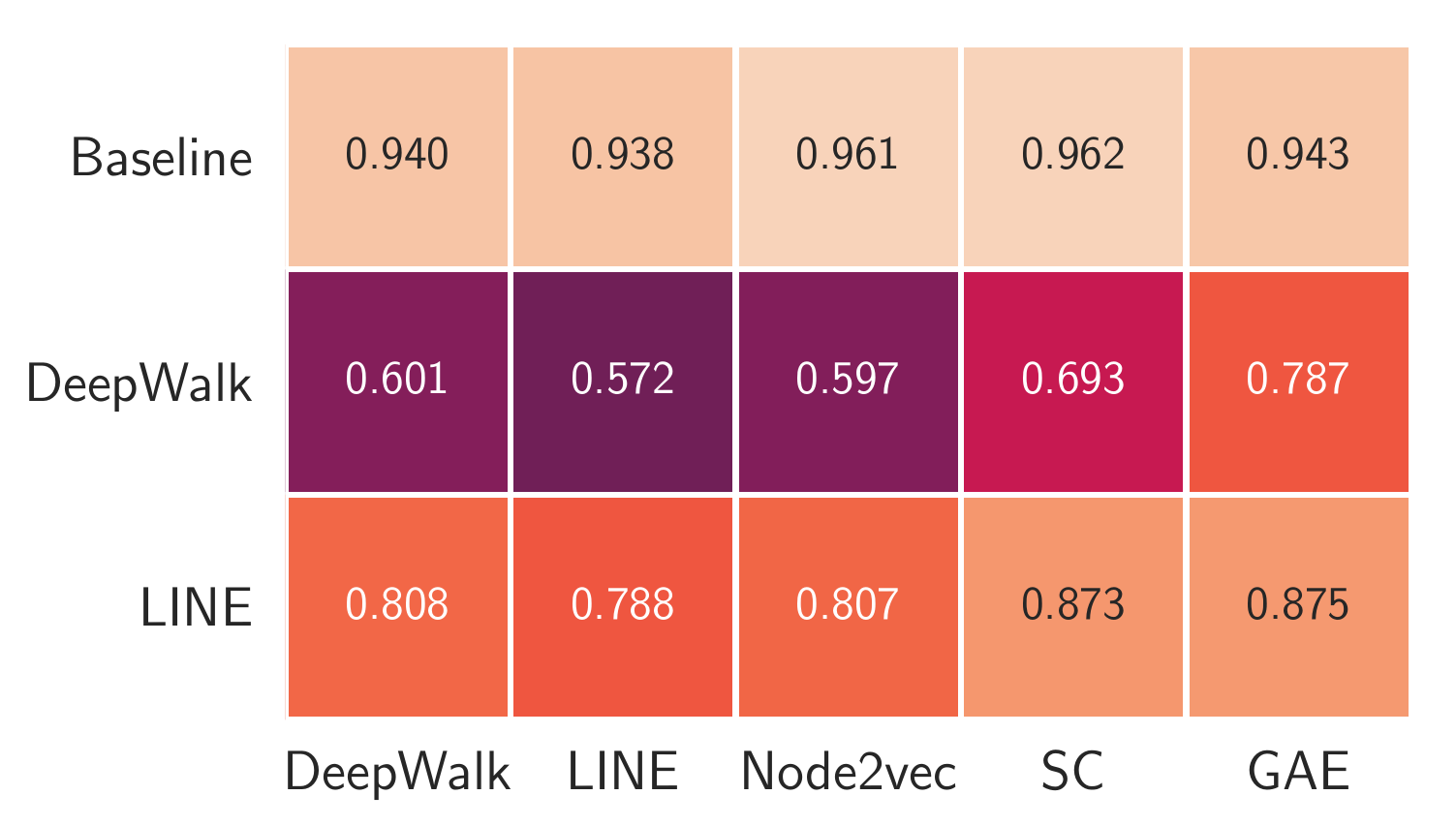}
 \caption{Citeseer-Add}
 \label{fig9:3}
 \end{subfigure}
\end{minipage}
\begin{minipage}{.49\textwidth}
 \begin{subfigure}{\textwidth}
 \centering
 \captionsetup{singlelinecheck=off, margin={2.8cm, 0cm}, format=hang}
 \includegraphics[width=\textwidth]{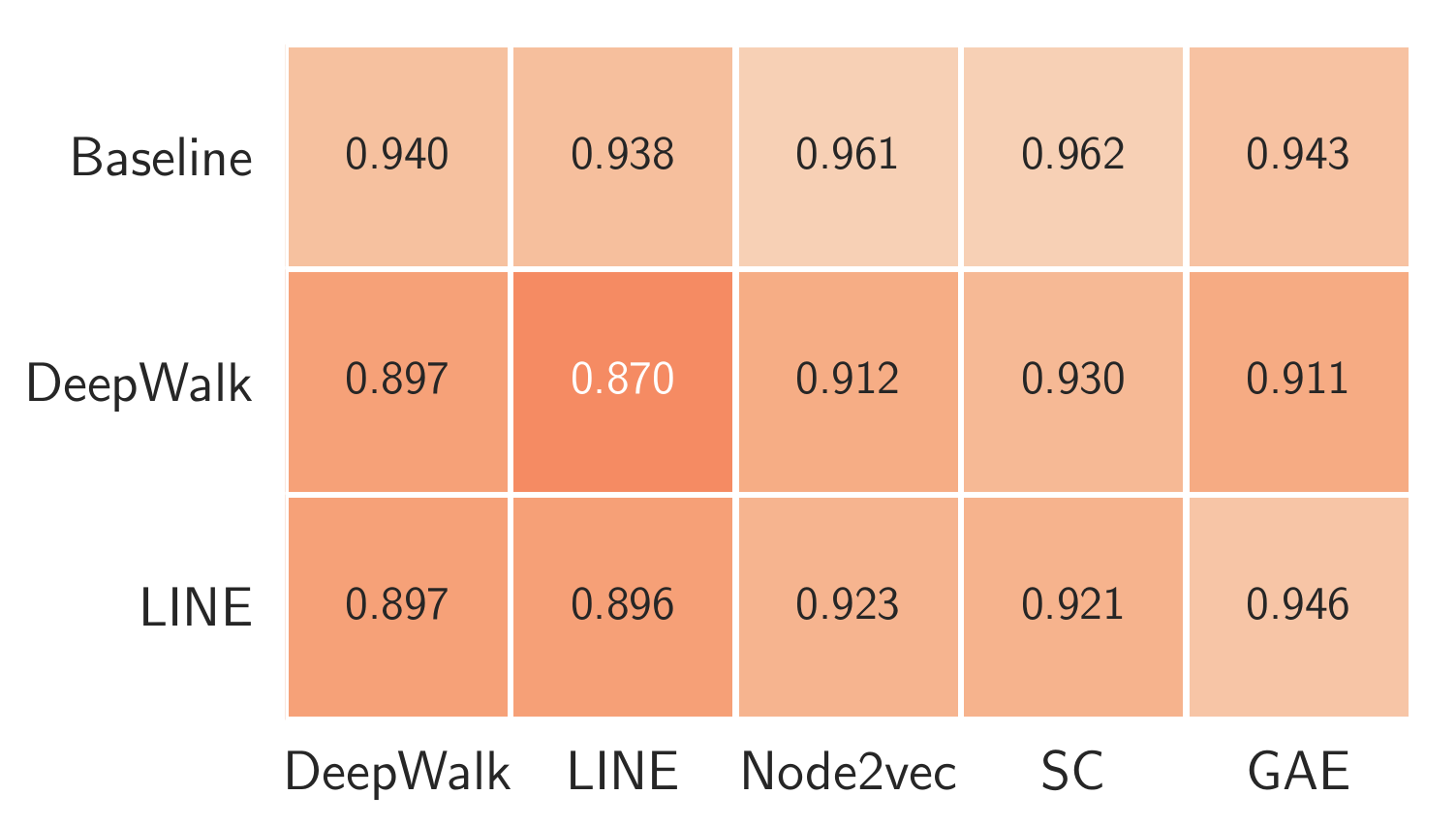}
 \caption{Citeseer-Add}
 \label{fig9:4}
 \end{subfigure}
\end{minipage}
\caption{Result for transferability analysis of our attack on two datasets, where the number of added/deleted edges is 100. X-axis indicates the method the attack is evaluated on. Y-axis includes the methods to generate the attack and also the baseline. The format ``Dataset | Type '' here is used to label the each sub-caption. ``Dataset'' refers to dataset while ``Type'' refers to attacker's action. }
\label{trans-2}
\vspace{-2.5ex}
\end{figure*}

\begin{figure*}[!htbp]
\centering
\begin{minipage}{.49\textwidth}
 \begin{subfigure}{\textwidth}
 \centering
 \captionsetup{singlelinecheck=off, margin={3.0cm, 0cm}, format=hang}
 \includegraphics[width=\textwidth]{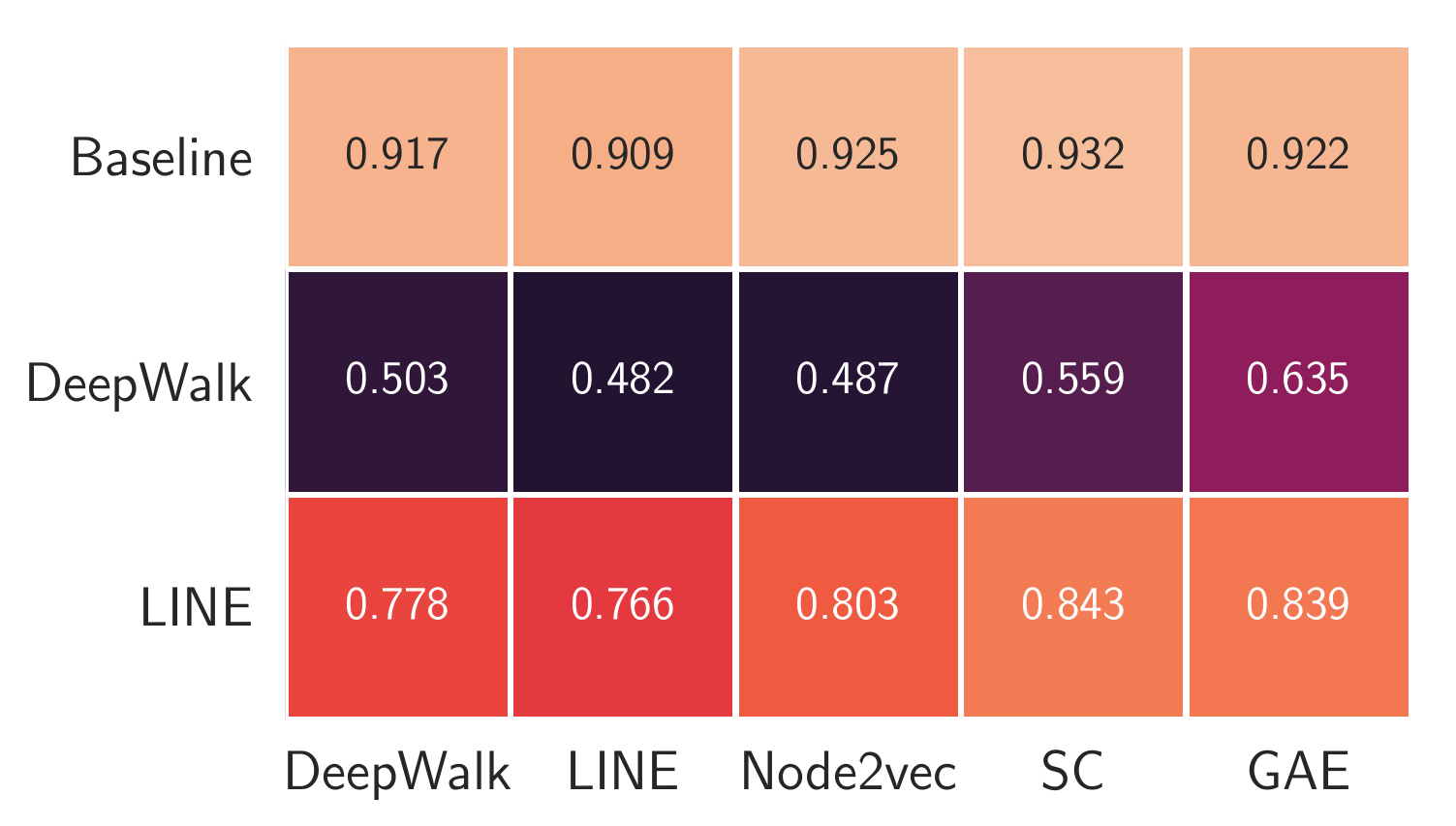}
 \caption{Cora-Add}
 \label{fig10:1}
 \end{subfigure}
\end{minipage}
\begin{minipage}{.49\textwidth}
 \begin{subfigure}{\textwidth}
 \centering
 \captionsetup{singlelinecheck=off, margin={3.0cm, 0cm}, format=hang}
 \includegraphics[width=\textwidth]{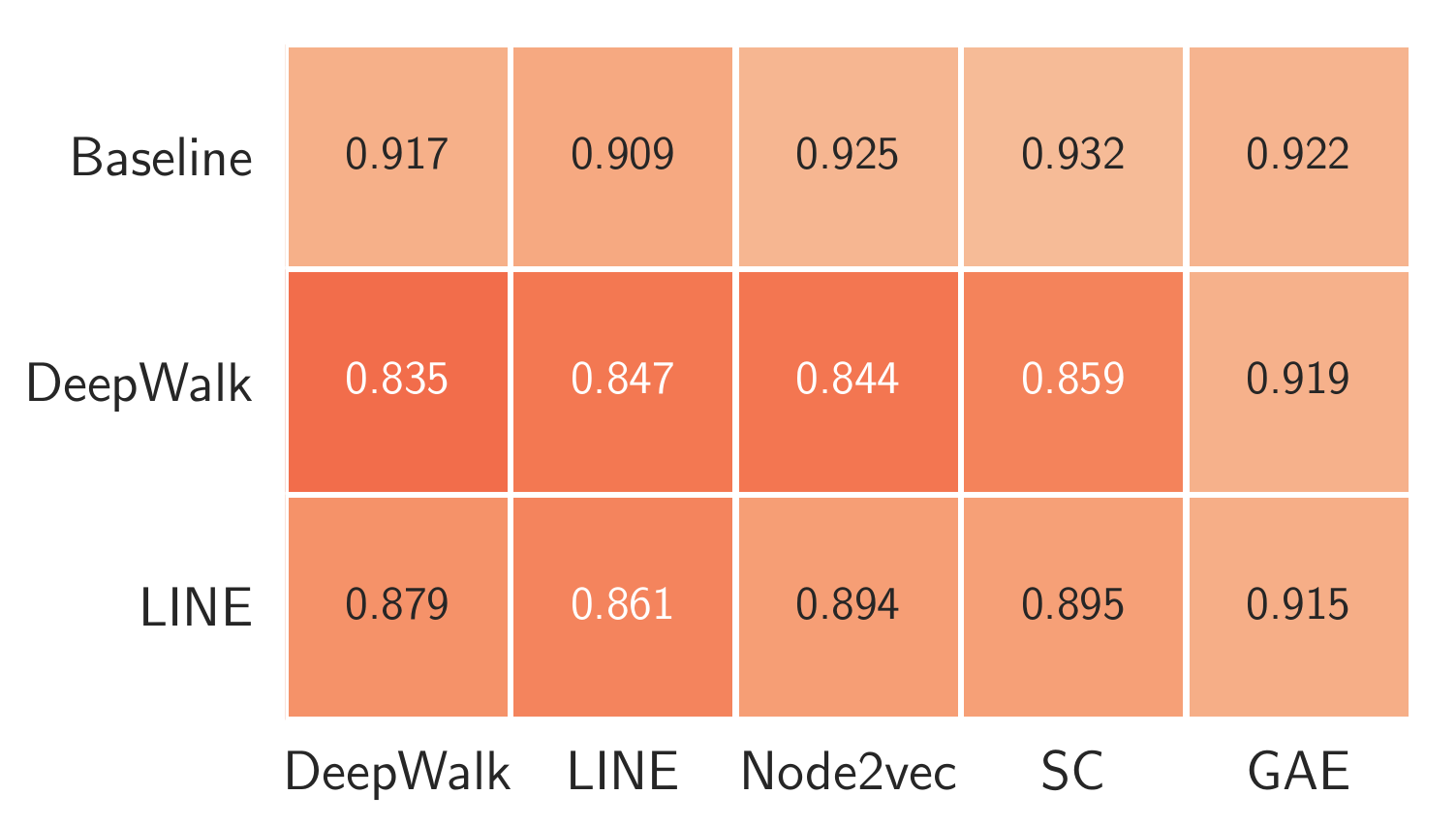}
 \caption{Cora-Del}
 \label{fig10:2}
 \end{subfigure}
\end{minipage}
\begin{minipage}{.49\textwidth}
 \begin{subfigure}{\textwidth}
 \centering
 \captionsetup{singlelinecheck=off, margin={2.8cm, 0cm}, format=hang}
 \includegraphics[width=\textwidth]{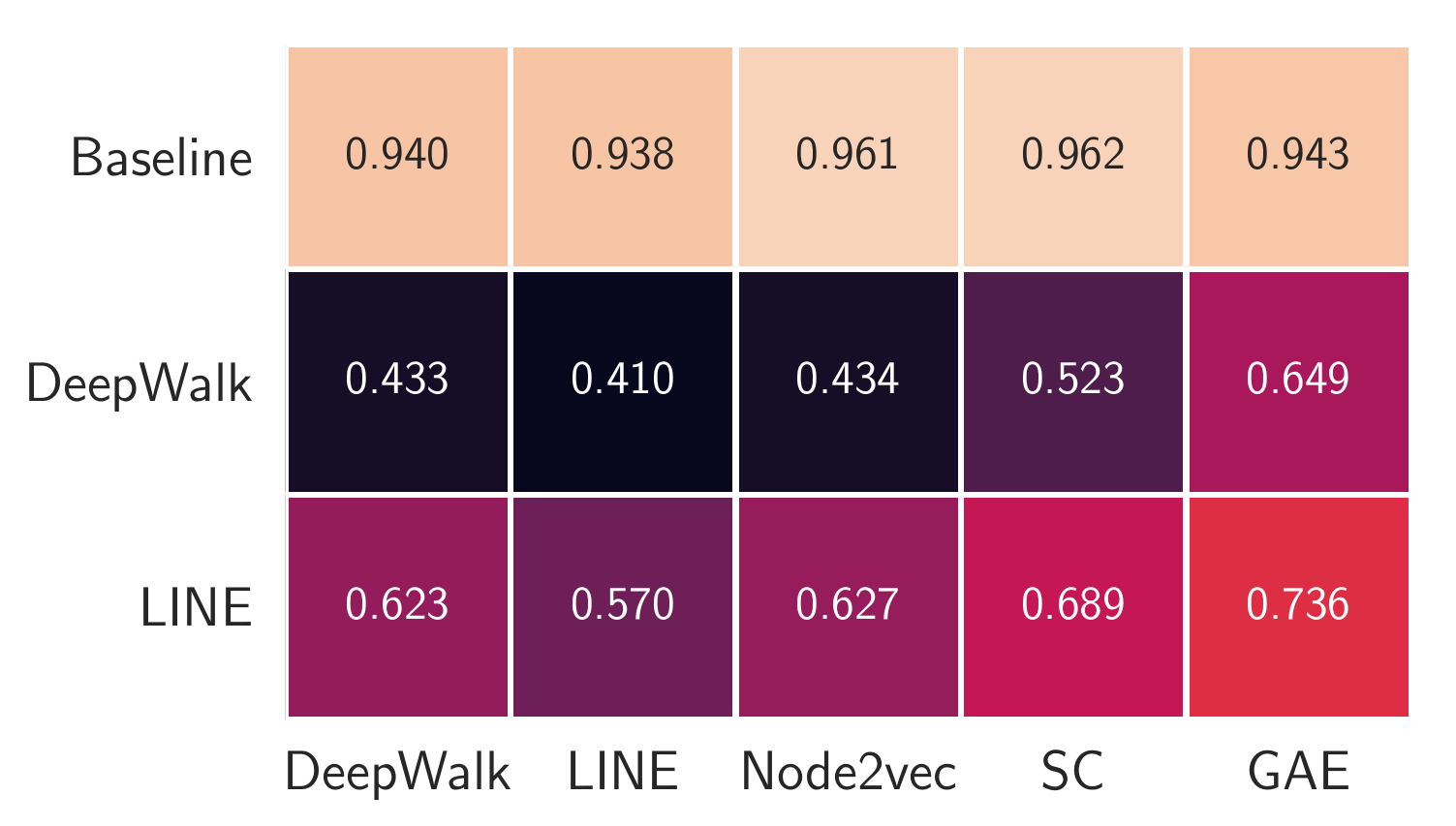}
 \caption{Citeseer-Add}
 \label{fig10:3}
 \end{subfigure}
\end{minipage}
\begin{minipage}{.49\textwidth}
 \begin{subfigure}{\textwidth}
 \centering
 \captionsetup{singlelinecheck=off, margin={2.8cm, 0cm}, format=hang}
 \includegraphics[width=\textwidth]{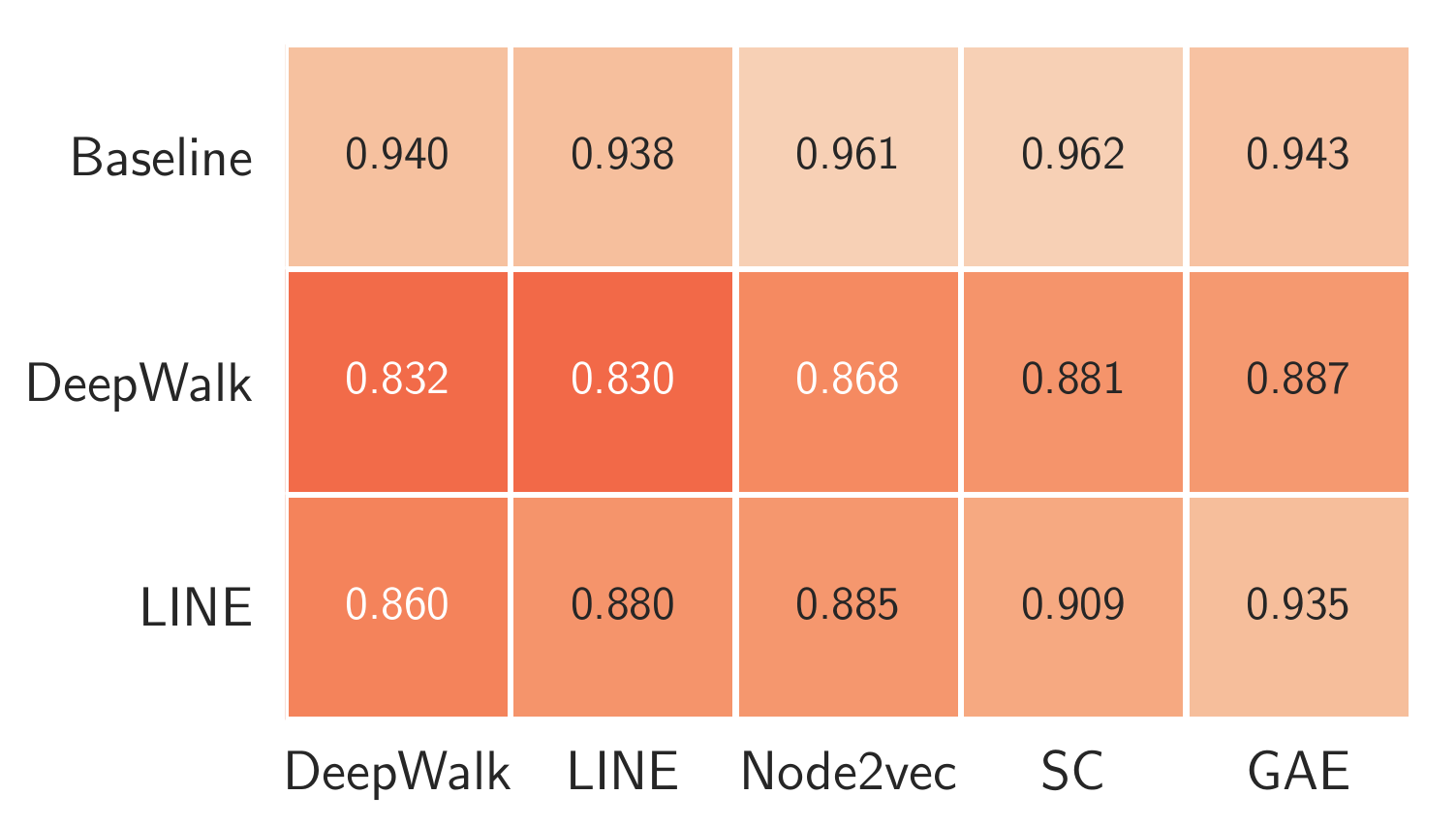}
 \caption{Citeseer-Add}
 \label{fig10:4}
 \end{subfigure}
\end{minipage}
\caption{Result for transferability analysis of our attack on two datasets, where the number of added/deleted edges is 300. X-axis indicates the method the attack is evaluated on. Y-axis includes the methods to generate the attack and also the baseline. The format ``Dataset | Type '' here is used to label the each sub-caption. ``Dataset'' refers to dataset while ``Type'' refers to attacker's action. }
\label{trans-3}
\vspace{-2.5ex}
\end{figure*}

Here we show additional results for availability attack. Table~\ref{citeseer-avail} shows the results of availability attack on Citeseer dataset. Figure~\ref{trans-2},~\ref{trans-3} show additional results of transferability analysis.
\renewcommand{\arraystretch}{1.2}
\begin{table}[!htbp]
\resizebox{1.0\textwidth}{!}{
\begin{tabular}{l|ccclllllll}
\hline
\multicolumn{1}{c|}{\multirow{2}{*}{Action}} & \multirow{2}{*}{Model}                        & \multirow{2}{*}{Baseline} & \multirow{2}{*}{Attack Method} & \multicolumn{7}{c}{\# added/deleted edges}                                                                                                                                        \\
\multicolumn{1}{c|}{}                        &                                               &                           &                                & \multicolumn{1}{c}{25} & \multicolumn{1}{c}{50} & \multicolumn{1}{c}{100} & \multicolumn{1}{c}{150} & \multicolumn{1}{c}{200} & \multicolumn{1}{c}{250} & \multicolumn{1}{c}{300} \\ \hline
\multicolumn{1}{c|}{\multirow{6}{*}{Add}}    & \multicolumn{1}{l}{\multirow{3}{*}{DeepWalk}} & \multirow{3}{*}{0.940}    & random                         & 0.940                  & 0.936                  & 0.935                   & 0.940                   & 0.937                   & 0.941                   & 0.942                   \\
\multicolumn{1}{c|}{}                        & \multicolumn{1}{l}{}                          &                           & degree sum                     & 0.919                  & 0.906                  & 0.886                   & 0.865                   & 0.864                   & 0.843                   & 0.828                   \\
\multicolumn{1}{c|}{}                        & \multicolumn{1}{l}{}                          &                           & Opt-attack                           & \textbf{0.810}                  & \textbf{0.706}                  & \textbf{0.601}                   & \textbf{0.536}                   & \textbf{0.493}                   & \textbf{0.461}                   & \textbf{0.433}                   \\ \cline{2-11} 
\multicolumn{1}{c|}{}                        & \multirow{3}{*}{LINE}                         & \multirow{3}{*}{0.938}    & random                         & 0.931                  & 0.934                  & 0.925                   & 0.925                   & 0.923                   & 0.926                   & 0.936                   \\
\multicolumn{1}{c|}{}                        &                                               &                           & degree sum                     & 0.929                  & 0.922                  & 0.925                   & 0.921                   & 0.920                   & 0.924                   & 0.924                   \\
\multicolumn{1}{c|}{}                        &                                               &                           & Opt-attack                           & \textbf{0.912}                  & \textbf{0.863}                  & \textbf{0.788}                   & \textbf{0.725}                   & \textbf{0.658}                   & \textbf{0.621}                  & \textbf{0.570}                   \\ \hline
\multirow{8}{*}{Delete}                      & \multirow{4}{*}{DeepWalk}                     & \multirow{4}{*}{0.940}    & random                         & 0.936                  & 0.927                  & 0.925                   & 0.926                   & 0.925                   & 0.919                   & 0.918                   \\
                                             &                                               &                           & degree sum           
                                & 0.932                  & 0.936                  & 0.927                   & 0.924                   & 0.921                   & 0.919                   & 0.922                   \\
                                             &                                               &                           & shortest path                  & 0.943                  & 0.925                  & 0.916                   & 0.907                   & 0.902                   & 0.900                   & 0.895                   \\
                                             &                                               &                           & Opt-attack                           & \textbf{0.923}                  & \textbf{0.914}                  & \textbf{0.897}                   & \textbf{0.884}                   & \textbf{0.867}                   & \textbf{0.860}                   & \textbf{0.832}                   \\ \cline{2-11} 
                                             & \multirow{4}{*}{LINE}                         & \multirow{4}{*}{0.938}    & random                         & 0.923                  & \textbf{0.919}                  & 0.915                   & 0.898                   & 0.900                   & 0.887                   & 0.893                   \\
                                             &                                               &                           & degree sum                     & 0.924                  & 0.926                  & 0.906                   & \textbf{0.889}                   & 0.902                   & 0.895                   & 0.888                   \\
                                             &                                               &                           & shortest path                  & \textbf{0.907}                  & \textbf{0.919}                  & 0.899                   & 0.915                   & 0.884                   & 0.884                   & 0.891                   \\
                                             &                                               &                           & Opt-attack                           & 0.925                  & 0.930                  & \textbf{0.896}                   & 0.897                   & \textbf{0.874}                   & \textbf{0.875}                   & \textbf{0.880}                   \\ \hline
\end{tabular}
}
\caption{Results for availability attack on Citeseer dataset. Here we report the AP score.
}
\label{citeseer-avail}
\end{table}

\end{appendices}
\end{document}